%% file: main.tex
\documentclass{article}

\usepackage[
  letterpaper,
  top=2cm,
  bottom=2cm,
  left=3cm,
  right=3cm,
  marginparwidth=1.75cm
]{geometry}
\usepackage[english]{babel}
\usepackage{csquotes}
\usepackage{placeins}
\usepackage{amsmath}
\usepackage{amssymb}
\usepackage{xcolor}

\usepackage{array}         
\usepackage{tabularx}      
\usepackage{booktabs}      
\usepackage{threeparttable}
\usepackage{float}         
\usepackage{multicol}      
\newcolumntype{Y}{>{\centering\arraybackslash}X} 

\usepackage{graphicx}
\usepackage{subcaption}
\usepackage[labelfont=bf]{caption}

\usepackage{algorithm}
\usepackage{algpseudocode}

\usepackage{enumitem}
\usepackage[title]{appendix}

\usepackage[
  backend=biber,
  style=apa,
  sortcites=true,
  maxbibnames=99,
  doi=true,
  url=true
]{biblatex}
\usepackage[colorlinks=true, allcolors=blue]{hyperref}

\title{Choosing the Right Regularizer for Applied ML: Simulation Benchmarks of Popular Scikit-learn Regularization Frameworks}
\author{Benjamin S. Knight and Ahsaas Bajaj}
\date{}

\begin{document}
\maketitle

\begin{abstract} 

\noindent This study surveys the historical development of regularization, tracing its evolution from stepwise regression in the 1960s to recent advancements in formal error control, structured penalties for non-independent features, Bayesian methods, and $\ell_0$-based regularization (among other techniques). We empirically evaluate the performance of four canonical frameworks—Ridge, Lasso, ElasticNet, and Post-Lasso OLS—across 134,400 simulations spanning a 7-dimensional manifold grounded in eight production-grade machine learning models. Our findings demonstrate that for prediction accuracy when the sample-to-feature ratio is sufficient ($n/p \geq 78$), Ridge, Lasso, and ElasticNet are nearly interchangeable. However, we find that Lasso recall is highly fragile under multicollinearity; at high condition numbers ($\kappa$) and low SNR, Lasso recall collapses to 0.18 while ElasticNet maintains 0.93. Consequently, we advise practitioners against using Lasso or Post-Lasso OLS at high $\kappa$ with small sample sizes. The analysis concludes with an objective-driven decision guide to assist machine learning engineers in selecting the optimal scikit-learn-supported framework based on observable feature space attributes.
\end{abstract}

\input{section_1_introduction}

\input{section_2_lit_review}

\input{section_3_regularization_in_sklearn}

\input{section_4_methodology}

\input{section_5_results}

\input{section_6_applications}

\input{section_7_conclusion}

\printbibliography

\newpage
\input{appendix}
\end{document}

%% file: section_1_introduction.tex
\section{Introduction}
Regularization is a key step in the prevention of overfitting while at the same time facilitating model parsimony. As machine learning has become more ubiquitous, a dizzying variety of regularization frameworks have emerged. In scikit-learn alone, we have the Least Absolute Shrinkage and Selection Operator (Lasso), LassoLars, Ridge, ElasticNet, and Orthogonal Matching Pursuit (OMP) among others. Parameterized regularization is also built into many common estimator classes including LogisticRegression, Support Vector Machines (SVMs), and neural networks (NNs).

We explore the three canonical regularization frameworks of Lasso, Ridge, and ElasticNet by simulating a space of feature spaces. Our goal is to emulate the data feature sets a Data Scientist or Machine Learning Engineer might prepare when creating a predictive model. While we look at feature recovery (the F1 score) as well as coefficient estimate accuracy (relative L2 error), our primary concern is predictive performance---we are in an applied setting where predictive power is paramount. In this context, we define success as minimizing the root mean squared error (RMSE), with run time and computational complexity being secondary concerns.

We start with a review of the regularization literature, highlighting how the initial question of which predictors to include in one's model has branched to encompass a variety of objectives, including predictor selection, shrinkage, the preservation of grouped predictors, and the formal enforcement of the false discovery rate (FDR---the proportion of
false discoveries among all coefficients included in the model) among other competing priorities. We then lay out our methodology for simulating a space of feature spaces suitable for empirically evaluating the canonical scikit-learn \parencite{pedregosa2011scikit} regularization frameworks of Lasso, Ridge, ElasticNet, as well as Post-Lasso Ordinary Least Squares (OLS). We explore these four frameworks' performances with respect to the three aforementioned criteria (feature recovery, coefficient accuracy, and predictive power). We conclude with recommendations intended for an applied setting.

\FloatBarrier

%% file: section_2_lit_review.tex
\section{Literature Review}
\subsection{Step-Wise Regression and \texorpdfstring{$\ell_2$}{l2} Norm-Based Regularization}

\indent The first solutions to the question of which features should be included in a model center on stepwise regression \parencite{efroymson1960regression}. Although different implementations of stepwise regression (for example, forward, backward) can improve predictive accuracy under ideal conditions, subsequent work demonstrates the tendency of the method to inflate the estimated $R^2$ \parencite{rencher1980inflation}, exaggerate the statistical significance of the $F$-test \parencite{wilkinson1981forward}, and generally lead to overfitting \parencite{copas1983shrinkage}.

More recent algorithms that improve upon stepwise regression typically fall along a spectrum, balancing two complementary objectives:

\begin{enumerate}[label=(\roman*), leftmargin=2cm, align=left, labelsep=1em]
    \item Reducing overfitting by minimizing the shrinkage of coefficient estimates when the model is applied to new data, and
    \item Reducing the risk of assigning nonzero coefficients to covariates that are unrelated to the true underlying model.
\end{enumerate}

Juggling these competing criteria, regularization tools vary in how they approach coefficient penalization - with some frameworks shrinking coefficients all the way to zero to address the second objective. Ridge regression \parencite{hoerl1970ridgeapp, draper1998regression, miller2019subset} uses the $\ell_2$ norm to shrink coefficients toward zero, but not to zero. Ridge regression cannot be used for variable selection, but is well-suited to large datasets where there is multicollinearity (we demonstrate this empirically in Section 5). Other methods emphasize selecting the best subset of variables. Some selection methods are unstable, with small perturbations in the data causing some methods to choose different sets of predictors, prompting Breiman to recommend approaches such as bagging and non-negative garrote \parencite{breiman1995garrote, breiman1996heuristics}. In their work on chemometric regression tools, Frank and Friedman propose `bridge' regression as a common framework underlying both Ridge regression and subset sighting \parencite{frank1993statistical}, with both methods minimizing the residual sum of squares subject to the constraint $\sum | \beta_{j} |^\gamma < t$ (where $\gamma$ = 2 for Ridge regression and $\gamma$ = 0 for subset selection) \parencite{fu1998bridge}.

\subsection{\texorpdfstring{$\ell_1$}{l1} Norm-Based Regularization}

Building on Breiman's work, \textcite{tibshirani1996lasso} proposes Lasso and its use of the $\ell_1$ norm as a robust solution to the variable selection problem. Aggarwal et al. observe that as the dimensionality of a feature space increases, norms with smaller parameter values tend to offer more contrast relative to higher order norms (e.g., $\ell_{k-1}$ tends to dominate $\ell_k$) and that the Manhattan distance ($\ell_1$) metric provides better contrast than the Euclidean distance ($\ell_2$) \parencite{aggarwal2001distance}.

Subsequent the advent of Lasso, \textcite{efron2004lar} identify commonalities between Lasso and forward stagewise regression in the form of Least Angle Regression (LARS) \parencite{hesterberg2008review}. LARS is a stepwise algorithm with analytic solutions for the steps, making the implementation of Lasso-esque methods computationally cheap and readily accessible in popular software packages like scikit-learn.

As successful as Lasso has been, the technique is not without its limitations. In their work on smoothly clipped absolute deviation (SCAD) penalization, Fan and Li note that $\ell_k$ penalty functions do not simultaneously satisfy the mathematical conditions for unbiasedness, sparsity, and continuity \parencite{fan2001nonconcave}. Zou flags Lasso's tendency to reduce the size of the fitted estimates, often by an excessive amount \parencite{zou2006adaptive}; this is particularly relevant for our primary objective of prediction accuracy. Leng et al. observe that Lasso and its family of regularization procedures (LARS, forward stagewise regression) are not consistent in their selection of variables when optimizing for model predictive accuracy, i.e., the subset of variables preserved do not reflect the true model \parencite{leng2006lasso}. Others note that Lasso can fail to include all the relevant variables while simultaneously pruning 100\% of the irrelevant features \parencite{fan2010sis, su2017lassopath}. Theoretical work shows that under certain circumstances, Lasso can fail to correctly return the correct level of support even when information theory implies that the correct level of support is retrievable \parencite{wainwright2009sparsity, wang2010limits, gamarnik2017binary}.

ElasticNet \parencite{zou2005elasticnetaddendum} is one of the more popular approaches seeking to improve on Lasso. This technique combines Ridge and Lasso, using a linear combination of $\ell_1$ and $\ell_2$. The end result of the ElasticNet penalty is a function that promotes the averaging of highly correlated features while at the same time encouraging a sparse solution (albeit one that is less sparse than Lasso) \parencite{hastie2009elements}. However, the combination also inherits the weaknesses of both approaches. Bertsimas, King, and Mazumder argue that like Lasso, ElasticNet does a poor job of recovering the pattern of sparsity in instances where the total number of variables greatly exceeds the number of relevant variables \parencite{bertsimas2016subset}. Lasso biases the regression regressors as a consequence of the $\ell_1$ norm, uniformly penalizing both large and small coefficients \parencite{bertsimas2020highdim}. Even worse for prediction accuracy, the $\ell_2$ component strongly biases the coefficients for important features \parencite{hesterberg2008review}.

\begin{figure}[H]
    \centering
    \includegraphics[width=0.9\textwidth]{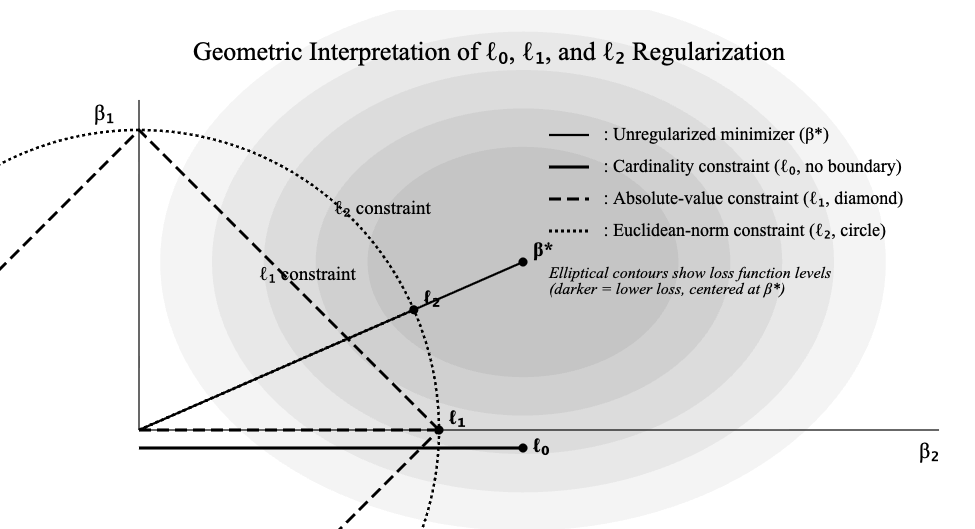}
    \caption{The three norms as applied to regularization. Regularization using the $\ell_0$ norm (solid line) performs discrete variable selection, keeping only the larger coefficient while setting the smaller one to zero. Lasso regularization (thick dashed line) via the $\ell_1$ norm nudges $\hat{\beta}$ towards the axes, thus tending towards partial sparsity with some shrinkage of both coefficients. Ridge regression (dotted circular line) uses the $\ell_2$ norm, shrinking both coefficients proportionally toward the origin while maintaining their relative magnitudes.}
    \label{fig:Visualization of the constraints enforced by L0, L1, and L2 norms.}
\end{figure}

\subsection{\texorpdfstring{$\ell_0$}{l0} Norm-Based Regularization and Bayesian Methods}

Going beyond ElasticNet penalization, more recent efforts to improve upon Lasso include the Minimax Concave Penalty (MCP) algorithm proposed by Cun-Hui Zhang. MCP uses a minimax concave penalty in conjunction with a penalized linear unbiased selection algorithm to improve upon Lasso's biasedness with respect to variable selection \parencite{zhang2010mcp}. Building on the findings of Pilanci et al. \parencite{pilanci2015boolean}, Bertsimas, Pauphilet and Van Parys argue that $\ell_0$ regularization (cardinality-based penalization) is now viable thanks to improvements in hardware and mixed-integer optimization methods. Specifically, the authors demonstrate the potential of applying boolean relaxation to cardinality penalized estimators (i.e., explicitly constraining the number of features) \parencite{bertsimas2020sparse, bertsimas2020rejoinder, bertsimas2021classification}.

While $\ell_0$-based regularization represents a promising development, more work remains to be done and it is not evident that $\ell_0$-based regularization outperforms traditional $\ell_1$-based regularization in all instances. In the discussion paper by \textcite{chen2020robustness}, the authors discuss how $\ell_1$-based regularization continues to be more computationally tractable than $\ell_0$-based regularization, potentially allowing more extensive exploration of a given feature space. In addition, the authors note how Lasso's tendency to aggressively shrink its estimates may be advantageous in feature spaces where the signal to noise ratio is relatively poor. Lastly, the $\ell_1$ norm enforces robustness, enabling Lasso to withstand large perturbations in the data, a major strength that Bertsimas et al. acknowledge in their rejoinder \parencite{bertsimas2020rejoinder}.

A potentially more computationally tractable way of achieving ideal, $\ell_0$-based feature selection is the incorporation of Bayesian methods. \textcite{park2008bayesianlasso} show how frequentist Lasso estimation corresponds to the posterior mode under a Laplace (double exponential) prior. This approach holds the promise of improving upon Lasso. Because Lasso performs variable selection and shrinkage simultaneously, deriving valid standard errors or confidence intervals for the selected coefficients is problematic. By generating full posterior distributions, Park and Casella exploit this equivalence---yielding Bayesian credible intervals that can guide variable selection and rigorously quantify uncertainty. 

Unfortunately, use of the double exponential prior is not without its downside. \textcite{carvalho2010horseshoe} note that models using the double exponential prior must incorporate a global scale parameter ($\tau$ for Carvalho et al., $\lambda$ for Park and Casella). Regardless of how $\tau$ is derived, this global scale parameter shrinks noise across the distribution---successfully near the origin, but also in the tails where the benefit of such shrinkage is more dubious. In their own words, ``estimation of $\tau$ under the double-exponential model must balance two competing forces: risk due to undershrinking noise, and risk due to overshrinking large signals. This compromise is forced by the structure of the prior, and will be required under any model without tails sufficiently heavy to ensure a redescending score function'' \parencite[p.~472]{carvalho2010horseshoe}.

To address this limitation, Carvalho et al. propose the following multivariate-normal scale mixture `Horseshoe' estimator.

\begin{equation}
  \label{eq:horseshoe_estimator}
\theta_i | \lambda_i \sim N(0, \lambda_i^2) \mbox{  and  } \lambda_i | \tau \sim C^+(0, \tau)
\end{equation}

The Horseshoe works by breaking the prior distribution down into two interconnected levels. The first (left) level encodes the local signal. Let $\theta_i$ represent the $i$-th parameter estimate (e.g., an individual regression coefficient of interest). $\lambda_i$ is the local shrinkage parameter (every feature in the dataset gets its own independent variance parameter). In this framework, every coefficient is drawn from a Normal distribution centered at zero. Note how the standard deviation of the Normal distribution from which the parameter is drawn is equal to the shrinkage parameter. If the model determines that $\lambda_i$ is tiny, the Normal distribution becomes tightly bounded around zero (effectively identifying it as noise). If $\lambda_i$ is large, the Normal distribution is wide, allowing the coefficient $\theta_i$ to remain large and unshrunk. 

The second (right) level dictates exactly how the local $\lambda_i$ parameters are generated. The local shrinkage parameters are drawn from a Half-Cauchy distribution (the positive reals) with a location parameter of 0 and a scale parameter equal to a \textit{global} shrinkage parameter, $\tau$. Through this hierarchical arrangement of global and local shrinkage logics, the Horseshoe estimator is able to recognize when a signal is large and leave it completely unshrunk and unbiased, while at the same time allowing it to act far more aggressively in shrinking irrelevant variables toward zero vis-\`a-vis the Bayesian Lasso.

\subsection{\texorpdfstring{$\ell_1$}{l1} Minimization and the Dantzig Selector}

While modern hardware has made $\ell_0$-based regularization more feasible, the problem is still fundamentally combinatorial in nature, with complexity scaling exponentially as a function of the number of features, $p$. A more computationally tractable approach specifically intended for under-determined feature spaces ($p>n$) is available in the form of the Dantzig Selector (shown below). 

\begin{equation}
  \label{eq:dantzig_selector}
\min_{\tilde{\beta} \in \mathbf{R}^p} \|\tilde{\beta}\|_{\ell_1} \quad \text{subject to} \quad \|X^* r\|_{\ell_\infty} \le (1 + t^{-1})\sqrt{2 \log p} \cdot \sigma,
\end{equation}

In their seminal 2007 paper, \textcite{candes2007dantzig} take a novel approach to feature selection by seeking to minimize the $\ell_1$ norm directly (the $\|\tilde{\beta}\|_{\ell_1}$ term, i.e., the sum of the absolute values of the coefficients). This minimization is subject to the constraint that $\|X^* r\|_{\ell_\infty}$ does not exceed a given threshold, where $X^* r$ is the inner product (correlation) between every single predictor variable in the design matrix $X$ and the residual vector, and $\|\cdot\|_{\ell_\infty}$ is the infinity norm, i.e., the maximum absolute value taken among said correlations. By virtue of this constraint, the Dantzig Selector enforces the requirement that any valid model produce residuals that are the equivalent of random noise.

Parameterizing the constraint ($(1 + t^{-1})\sqrt{2 \log p} \cdot \sigma$) is tricky. While $t$ is a simple scaling parameter set by the user, $\sigma$ quantifies the standard deviation of the underlying Gaussian noise that subsumes the observations. This parameter is difficult to estimate to the point of being unestimable at higher values of $p$. For this reason, the Dantzig Selector thresholding simplifies to $\|X^* r\|_{\ell_\infty} \le \lambda$ in an applied settings.

The strength of the Dantzig Selector is that it acts as a computationally tractable, convex relaxation of best subset selection ($\ell_0$ norm) as explored by Bertsimas et al. However, as much as the Dantzig Selector has revolutionized how we understand sparse signal recovery when $p > n$, direct applications of the methodology do not typically yield results that markedly improve upon Lasso. \textcite{meinshausen2007discussion} show that ``L2Boosting and Lasso are in general no worse and sometimes better than Dantzig" while noting that the Dantzig solution path is jittery relative to the solution paths for Lasso or L2Boosting---especially for highly correlated predictor variables. Bickel, Ritov, and Tsybakov show that under the same sparsity scenarios, Lasso and the Dantzig Selector exhibit virtually identical behavior \parencite{bickel2009simultaneous}. Lastly, the Dantzig Selector still relies on standard Linear Programming (LP), placing it at a computational disadvantage relative to LARS Lasso. In this way, the Dantzig Selector's primary contribution is in regards to the mathematical foundations of regularization as opposed to performant tooling.

\subsection{Accommodating Non-Independent Features} 

The approaches outlined above all share the assumption that the features in a dataset are independent---i.e., an `unordered bag of features.' However, there will be times when we are more interested in groups of features (e.g., the phenomenon of interest is encoded in the interaction of one or more variables) or there is ordinal information in the features (e.g., time series data). If our goal is change-point detection, then we need a regularization framework that not only imposes sparsity on the coefficients themselves, but also on their successive differences. To this end, \textcite{tibshirani2005fusedlasso} propose use of the Fused Lasso.

\begin{equation}
  \label{eq:fused_lasso}
\sum_{i=1}^N (y_i - \mathbf{x}_i^{\mathrm{T}} \beta)^2 + \lambda_N^{(1)} \sum_{j=1}^p |\beta_j| + \lambda_N^{(2)} \sum_{j=2}^p |\beta_j - \beta_{j-1}|
\end{equation}

\noindent The penalized least squares criterion laid out above is comprised of three terms. The left-most term is the standard residual sum of squares---measuring how well the linear model fits the observed data, where $y_i$ is the actual outcome and $\mathbf{x}_i^{\mathrm{T}} \beta$ is the predicted outcome for the $i$-th observation. The middle term is the standard Lasso penalty where $p$ denotes the total number of features in the dataset, $N$ is the sample size, and $j$ is the coefficient index. This term serves to promote model parsimony. The third and final term is the defining innovation of the Fused Lasso. It applies a $\ell_1$ penalty to the successive differences between adjacent coefficients---where `adjacency' is defined by the researcher (e.g., mass-over-charge ratio of a set of proteins).

While Tibshirani et al. extend Lasso to handle ordinal data, \textcite{yuan2006grouped} engage the problem of preserving groups of interdependent features through their work on Group Lasso. The Group Lasso loss function is defined as follows:

\begin{equation}
  \label{eq:group_lasso}
  \frac{1}{2} \left\| Y - \sum_{j=1}^J X_j \beta_j \right\|^2 + \lambda \sum_{j=1}^J \|\beta_j\|_{K_j}
\end{equation}

\noindent The left side term is the standard residual sum of squares adapted for grouped data, where $Y$ is the vector of the response variable and the predictors are divided into $J$ distinct, non-overlapping groups. $X_j$ represents the design matrix specifically for the $j$-th group of predictors. $\beta_j$ is the vector of regression coefficients. The right-hand side is the group penalty term where $\lambda$ is the tuning parameter that controls the overall strength of the regularization and $\|\beta_j\|_{K_j}$ is a bespoke norm applied to the coefficient vector of the $j$-th group (Yuan and Lin advise $K_j = p_j I_{p_j}$ where $p_j$ is the number of variables in the $j$-th group and $I_{p_j}$ is the standard identity matrix of size $p_j \times p_j$).

By combining the geometric properties of both the $\ell_1$ and $\ell_2$ norms, Group Lasso is able to act like the standard Lasso ($\ell_1$ penalty) at the group ($J$) level, while use of the $\ell_2$ norm applies Ridge-like shrinkage to individual coefficients within preserved groups. \textcite{huang2012groupselection} note that while Group Lasso displays strong performance in terms of prediction and $\ell_2$ estimation errors, Group Lasso still inherits some of Lasso's weaknesses - specifically a tendency to recruit unimportant variables into the model to compensate for over-shrinking of large coefficients, and consequently, a relatively high false positive rate when selecting features. Group Lasso also has a limitation: as a consequence of using the $\ell_2$ norm within groups, it cannot enforce within group sparsity. 

The incorporation of problem-specific assumptions can be a boon when modeling high-dimensional supervised learning problems. To enable within-group sparsity while preserving groups of features \textcite{simon2013sparsegroup} propose the Sparse-Group Lasso --- an extension of the Group Lasso. 

\begin{equation}
  \label{eq:sparse_group_lasso}
\min_{\beta} \frac{1}{2n} \left\| y - \sum_{l=1}^m X^{(l)} \beta^{(l)} \right\|_2^2 + (1 - \alpha)\lambda \sum_{l=1}^m \sqrt{p_l} \|\beta^{(l)}\|_2 + \alpha\lambda \|\beta\|_1
\end{equation}

The left-most term is the standard mean squared error loss function, evaluating how well the linear model fits the observed data $y$. The middle term is the penalty term from the Group Lasso. It applies an un-squared Euclidean norm ($\ell_2$ norm) to the coefficient vector of each group $l$, and scales it by the square root of the group's size ($\sqrt{p_l}$) to ensure large groups aren't unfairly penalized. The third and final term is the traditional $\ell_1$ penalty. Taken together, the loss function for the Sparse-Group Lasso adopts the same broad strategy of parameterized linear combination that ElasticNet takes when trading off Ridge and Lasso---with one important twist. Unlike ElasticNet, the Sparse-Group Lasso uses the un-squared $\ell_2$ norm due to it being non-differentiable exactly at zero. In this way, it acts just like a Lasso penalty at the macro-level, allowing it to completely zero-out entire groups of variables. 

\subsection{Advancements in Error Control}

One of the more significant developments in the field of regularization is the advent of stability selection \parencite{meinshausen2010stability}. While bagging (see \textcite{breiman1996heuristics}) can encourage estimate stability, it does not provide formal error control over the expected number of incorrect coefficients incorporated into the resulting model (i.e., the Per-Family Error Rate, or PFER). In their 2010 paper, Nicolai Meinshausen and Peter Bühlmann detail the Stability Selection procedure---a meta-algorithm designed to enhance existing variable selection methods.

The Stability Selection procedure is as follows: Our goal is to extract a set of stable features, denoted by $\hat{S}_{\mathrm{stable}}$. Let $\Lambda$ represent a predefined grid of candidates for the regularization parameter $\lambda$, and let $C_k^\lambda$ represent selection counters initialized to zero. For each iteration within an outer loop of length $M$, we draw a sub-sample of size $\lfloor n/2 \rfloor$ without replacement. Then, within an inner loop, for each value of $\lambda \in \Lambda$, we fit our model using the regularization framework of our choice (in this case, Lasso) and extract the resulting active set of non-zero coefficients, incrementing the counters for the selected features. We then calculate the empirical selection probability $\hat{\Pi}_k^\lambda$ for each feature $k$ at each penalty level. Features whose maximum selection probability across the entire grid $\Lambda$ exceeds a predefined threshold ($\max_{\lambda \in \Lambda} \hat{\Pi}_k^\lambda \ge \pi_{thr}$) are preserved.

Traditionally, researchers have struggled with defining and implementing the `correct' level of regularization. Through this method, a researcher is able to impose strict, finite-sample mathematical control over the expected number of false positives, allowing said researcher to set this tolerance level at the onset. A formal summary of the algorithm is laid out below.

\begin{algorithm}[H]
\caption{Stability Selection Wrapper with Lasso}
\begin{algorithmic}[1]
\Require Dataset with $n$ observations, regularization region $\Lambda$, threshold $\pi_{thr} \in (0, 1)$, iterations $M$
\Ensure Set of stable features $\hat{S}_{\mathrm{stable}}$
\State Initialize selection counters $C_k^\lambda = 0$ for all variables $k$ and $\lambda \in \Lambda$
\For{$m = 1, \dots, M$}
    \State Draw a random subsample $I_m \subset \{1, \dots, n\}$ of size $\lfloor n/2 \rfloor$ without replacement
    \For{each $\lambda \in \Lambda$}
        \State \textbf{Fit Model:} Minimize the Lasso objective function on $I_m$:
        \Statex \quad $\sum_{i=1}^n (y_i - \mathbf{x}_i^{\mathrm{T}} \beta)^2 + \lambda \sum_{j=1}^p |\beta_j|$
        \State \textbf{Extract:} Identify active set $\hat{S}^\lambda(I_m)$ of non-zero coefficients
        \For{each $k \in \hat{S}^\lambda(I_m)$}
            \State $C_k^\lambda \gets C_k^\lambda + 1$
        \EndFor
    \EndFor
\EndFor
\State \textbf{Calculate Probabilities:} $\hat{\Pi}_k^\lambda = \frac{C_k^\lambda}{M}$
\State \textbf{Thresholding:} $\hat{S}_{\mathrm{stable}} = \left\{k : \max_{\lambda \in \Lambda} \hat{\Pi}_k^\lambda \geq \pi_{thr}\right\}$
\State \Return $\hat{S}_{\mathrm{stable}}$
\end{algorithmic}
\end{algorithm}

While the Stability Selection procedure emphasizes control over the PFER (i.e., the expected absolute number of falsely selected variables in the final model), the Knockoff Filter pioneered by \textcite{candes2018modelx} is specifically designed to control the FDR.

To implement the Knockoff procedure, we start by specifying the target FDR. Our goal is to derive a set of active features (denoted by $\hat{S}$) that satisfies this aforementioned FDR threshold. Our next step is to construct the knockoffs. The key requirement when constructing the knockoff data is that it perfectly mimics the distribution of $X$ while being conditionally independent of $Y$ given $X$ ($\tilde{X} \perp\!\!\perp Y \mid X$). In other words, the two sets of data must exhibit the pairwise exchangeability property.

After combining the original and synthetic datasets, we extract the estimated coefficients for both the original features and their knockoff counterparts. We then calculate the feature statistic, $W_j$, which is typically the difference in the absolute magnitudes of the coefficient estimates for feature $j$ and its knockoff. Because the original variable and its knockoff are perfectly exchangeable under the null hypothesis, a true noise variable is equally likely to have a larger absolute coefficient or a smaller absolute coefficient compared to its knockoff. Thus, the sign of $W_j$ acts as a coin flip for noise variables. In contrast, we expect values of $W_j$ to tend positive when there is a genuine relationship with $Y$.

Finally, the algorithm evaluates candidate thresholds $t > 0$ to estimate the proportion of false discoveries. We define the final data-dependent threshold $\tau$ as the minimum value of $t$ that keeps the estimated False Discovery Proportion at or below our target FDR. Any feature $j$ with a difference statistic $W_j \ge \tau$ is permanently included in the final selected set $\hat{S}$.

\begin{algorithm}[H]
\caption{The Knockoff Filter Procedure with Lasso}
\begin{algorithmic}[1]
\Require Data matrix $X \in \mathbb{R}^{n \times p}$, response vector $y \in \mathbb{R}^n$, target FDR level $q \in (0, 1)$
\Ensure Selected set of active variables $\hat{S}$
\State \textbf{Construct Knockoffs:} Generate synthetic feature matrix $\tilde{X} \in \mathbb{R}^{n \times p}$ such that $(X, \tilde{X})$ respects pairwise exchangeability:
\Statex \quad $(X, \tilde{X})_{\text{swap}(S)} \overset{d}{=} (X, \tilde{X})$ for any subset $S \subset \{1, \dots, p\}$
\State \textbf{Fit Model:} Minimize the Lasso objective function on the augmented design matrix $[X, \tilde{X}]$:
\Statex \quad $\min_{b} \frac{1}{2n} \left\| y - [X, \tilde{X}] b \right\|_2^2 + \lambda \|b\|_1$
\State \textbf{Compute Feature Statistics:}
\For{$j = 1, \dots, p$}
    \State Extract coefficient for original variable: $Z_j = |\hat{b}_j|$
    \State Extract coefficient for knockoff variable: $\tilde{Z}_j = |\hat{b}_{j+p}|$
    \State Calculate the Lasso Coefficient Difference (LCD): $W_j = Z_j - \tilde{Z}_j$
\EndFor
\State \textbf{Data-Dependent Thresholding:} Calculate the strict threshold $\tau > 0$ to control the FDR:
\Statex \quad $\tau = \min \left\{ t > 0 : \frac{1 + \#\{j : W_j \le -t\}}{\#\{j : W_j \ge t\}} \le q \right\}$
\State \textbf{Final Selection:} $\hat{S} = \{j : W_j \ge \tau\}$
\State \Return $\hat{S}$
\end{algorithmic}
\end{algorithm}

\subsection{High-Dimensional Inference}                                                          Outside of machine learning (ML), one of the more common research objectives is estimating coefficients---values that ideally map to real-world phenomena such as the efficacy of a drug, change in revenue, etc.). The problem is that standard inference techniques (e.g., t-tests) fail catastrophically when applied to variables selected by regularization frameworks like Lasso. These procedures are data-dependent and tend to choose features that have the strongest correlation with the response variable, leading to biased coefficient estimates and highly inflated Type I error rates \parencite{fei2019spares, breiman1992bootstrap, berk2013postselection, lee2016postselection}. One of the more popular solutions to these problems---post-Lasso OLS---we summarize in the next section.

Other approaches involve debiasing the Lasso estimator directly. To this end, both \textcite{zhang2014confidence} and \textcite{vandegeer2014hdtests} propose bias correction by ``projecting" the residuals back onto the design matrix using an approximate inverse of the Gram matrix---with Zhang \& Zhang specifically focusing on individual regression coefficients and linear combinations, while van de Geer et al. extend their approach to encompass generalized linear models. While both approaches enable the construction of confidence intervals and p-values in high-dimensional ($p > n$) settings, it bears noting that the de-biasing is only mathematically valid under strict sparsity conditions---i.e. the number of non-zero parameters ($s_0$) must be much smaller than the square root of the sample size divided by the log of the number of parameters ($s_0 = o(\sqrt{n}/\log p)$).

Building on the work of van de Geer et al., \textcite{javanmard2014confidence} propose a computationally efficient method for constructing a de-biased estimator from the standard regularized LASSO solution: $\hat{\theta}^{u} = \hat{\theta}^{n} + \frac{1}{n}MX^{T}(Y - X\hat{\theta}^{n})$. Here $\hat{\theta}^{n}$ is the original (biased) Lasso estimate, and $(Y - X\hat{\theta}^{n})$ are the residuals vis-\`a-vis the Lasso model predictions and the training labels. Most importantly, $M$ is a design matrix fit via convex optimization that is applied to `decorrelate' the features. Javanmard and Montanari show that the confidence intervals derived from the resulting estimator are of the optimal size, the associated p-values are uniformly distributed, and that their framework is robust to non-Gaussian noise.

\begin{center}
* \phantom{aa} * \phantom{aa} * \phantom{aa}
\end{center}

\indent There is a rich array of literature on regularization---a consequence of the varied (and persistent) problems researchers face including variable selection, recovery of groups of features, coefficient estimation, high-dimensional hypothesis testing, and response variable prediction. Prediction---the last of these objectives---is our primary interest in an applied ML setting. We are specifically interested in the underlying efficacy of regularization procedures as implemented in the popular Python package scikit-learn. 

It bears noting that scikit-learn's linear models module currently lacks native support for structured penalties such as the Fused Lasso, Group Lasso, or Sparse-Group Lasso. At the same time, scikit-learn typically does not provide p-values, confidence intervals, or Post-Selection Inference for Lasso or ElasticNet. 

On the positive side, Gradient Descent (GD) can act as an implicit form of regularization (\textcite{tang2025benign}, \textcite{luo2026penalization}). Specifically, GD implicitly biases solutions toward minimum $\ell_2$-norm interpolants, enabling the underlying model to generalize optimally from noisy ($p>n$) training data. This is good news for users of scikit-learn's \texttt{SGDRegressor},  \texttt{SGDClassifier}, and other GD-based APIs, or for that matter, users who opt for GD-based optimization under the hood for standard linear models (e.g. \texttt{LogisticRegression}).

In the next section, we lay out our empirical approach for evaluating the foundational methods that are readily available in scikit-learn---Lasso, Ridge, and ElasticNet in addition to the popular regularization technique of Post-Lasso OLS. 

\FloatBarrier

%% file: section_3_regularization_in_sklearn.tex
\section{Evaluating Regularization in scikit-learn}

To assess the strengths and limitations of common regularization frameworks, we evaluate three canonical scikit-learn procedures: \href{https://scikit-learn.org/stable/modules/generated/sklearn.linear_model.LassoCV.html}{LassoCV}, \href{https://scikit-learn.org/stable/modules/generated/sklearn.linear_model.RidgeCV.html}{RidgeCV}, and \href{https://scikit-learn.org/stable/modules/generated/sklearn.linear_model.ElasticNetCV.html}{ElasticNetCV} as well as post-Lasso OLS \parencite{belloni2013least}.

\subsection{Canonical Regularization Frameworks in Scikit-learn}

To assess performance under different conditions, we simulate a space of feature spaces that varies as a function of (i.) the number of features, (ii.) rank ratio, (iii.) eigenvalue dispersion, (iv.) $\beta$ coefficient distribution, (v.) sparsity level, (vi.) signal-to-noise ratio, and (vii.) sample size. We will now walk through the respective loss functions of the different methods.

Scikit-learn's implementation of LassoCV attempts to minimize the following loss function:

\begin{equation}
  \label{eq:einstein}
  \min_{w} { \frac{1}{2n_{\text{samples}}} ||X w - y||_2 ^ 2 + \alpha ||w||_1}
\end{equation}

\noindent $X$ is the matrix containing the feature space, with the vector of coefficient estimates denoted by $w$. $y$ is the vector of training labels. The center term—$||X w - y||_2 ^ 2$—is the squared $\ell_2$ norm (i.e., the Euclidean norm) which we divide by 2 times the sample size (scaling by 2 yields some computational efficiency when computing the gradient). The rightmost term denotes the product of the $\ell_1$ norm of coefficient estimates and the regularization parameter $\alpha$, which can be rewritten as:
\begin{equation}
   \alpha ||w||_1 = \alpha \sum_{j=1}^{p} |w_j|
\end{equation}

\noindent Here $p$ is the number of features in our model while $j$ is an iterator, summing the absolute values of the coefficient estimates.

Compare scikit-learn's implementation of LassoCV with RidgeCV:
\begin{equation}
    \min_{w} || X w - y||_2^2 + \alpha ||w||_2^2
\end{equation}

There are two key differences between these frameworks. Most importantly, RidgeCV scales the regularization parameter - $\alpha$ - with the $\ell_2$ norm of the vector of coefficient estimates ($w$) instead of the $\ell_1$ norm. This is what gives rise to the curved space of constraints for Ridge as opposed to angular space of constraints for Lasso as shown in Figure 1. There is an additional difference in that RidgeCV does not normalize for sample size as LassoCV does. From an applied perspective, given that we empirically evaluate suitable values for $\alpha$, the absence of this normalization is arguably inconsequential \parencite{lemaitre2023ridge}.

Lastly, the objective function used by ElasticNetCV is:
\begin{equation}
\min_{w} { \frac{1}{2n_{\text{samples}}} ||X w - y||_2 ^ 2 + \alpha \rho ||w||_1 +
\frac{\alpha(1-\rho)}{2} ||w||_2 ^ 2}
\end{equation}

\noindent The first two terms are essentially the objective function used by LassoCV but with one difference - the introduction of the parameter $\rho$ that governs the relative influence of the $\ell_1$ and $\ell_2$ norms.

We use the parameterizations described in Table 1 when evaluating these three regularization procedures. For the L1 / $\rho$ parameter we use the values recommended in the \href{https://scikit-learn.org/stable/modules/generated/sklearn.linear_model.ElasticNetCV.html#sklearn.linear_model.ElasticNetCV}{documentation}.

\begin{table}[H]
\centering
{\large \textbf{Scikit-learn Regularization Frameworks Evaluated}}
\vspace{0.5em}
\begin{tabular}{
>{\centering\arraybackslash}m{3.0cm}|
>{\centering\arraybackslash}m{1.9cm}|
>{\centering\arraybackslash}m{2.1cm}|
>{\centering\arraybackslash}m{2.7cm}|
>{\centering\arraybackslash}m{2.3cm}
}
\rule{0pt}{1.8em}
&
\textbf{LassoCV} &
\textbf{Post-Lasso OLS} &
\textbf{ElasticNetCV} &
\textbf{RidgeCV} \\
\hline
\textbf{Candidate Regularization Parameter Values} &
\multicolumn{4}{c}{\parbox[c]{8.2cm}{\centering \footnotesize
\textbf{Allowed Values for $\alpha$:}\\[0.2em]
$[10^5, 10^4, 10^3, 10^2, 10^1, 10^0, 10^{-1}, 10^{-2}, 10^{-3}]$}} \\
\hline
\textbf{Cross-Validation (CV)} &
\multicolumn{4}{c}{5-fold ($n{=}100$); 4-fold ($n{=}1,000$); 3-fold ($n{=}10,000$); 2-fold ($n{=}100,000$)} \\
\hline
\textbf{Test Set Holdout} &
\multicolumn{4}{c}{20\%} \\
\hline
\textbf{L1 Ratio Values} &
n/a & n/a & {\footnotesize [0.0, 0.1, 0.5, 0.7, 0.9, 0.95, 0.99, 1.0]} & n/a \\
\hline
\textbf{CV Fits per Configuration} &
18--45 & 18--45 & 144--360 & 18--45 \\
\end{tabular}
\caption{Regularization parameters for the four methods evaluated. All methods use identical alpha grid with no scaling—this respects scikit-learn's design where Lasso has built-in $(1/2n)$ normalization but Ridge does not. ElasticNet L1 ratios based on scikit-learn recommended defaults plus pure Ridge (0.0). All methods use identical train/test splits (80\%/20\%) and adaptive cross-validation.}
\label{tab:model_comparison}
\end{table}

\subsection{Inclusion of Post-Lasso OLS}

Post-Lasso OLS (also known as the `Gauss-Lasso' selector) is a popular regularization framework that, while not available under a dedicated scikit-learn API, is straightforward enough to implement in Python that we include it among our analysis. A key drawback of Lasso is shrinkage bias (\textcite{fan2001nonconcave}, \textcite{freijeiro_gonzalez_critical_lasso_2022}). Lasso’s $\ell_1$ penalty systematically shrinks relevant coefficients toward zero. We can avoid this bias by using Lasso purely for feature selection, and then refitting with OLS. While practiced heuristically for years, the first rigorous description of Post-Lasso OLS's performance is by \textcite{belloni2013least} who demonstrate that post-Lasso OLS performs at least as well as standard Lasso in terms of convergence, while successfully reducing regularization bias. Post-Lasso OLS is also noteworthy in that it successfully set the stage for the `Double Lasso' (\textcite{belloni2013least}, \textcite{fitzgerald_sice_practical_insights_double_lasso_2023}) --- an expansion of multi-stage feature selection / regularization regimes into the domain of causal inference methods (methods that are out of scope for our analysis).

\FloatBarrier

%% file: section_4_methodology.tex
\section{Methods: Constructing a Space of Feature Spaces}
Our goal is to construct a framework for generating synthetic data that is diverse enough to capture the strengths and weaknesses of different regularization frameworks, while still being grounded in real-world data a Machine Learning (ML) Engineer or Data Scientist is likely to encounter. We draw our sample of eight productionized ML training sets (see the appendix) from customer-facing models as well as from logistics / fulfillment models, encompassing domains as varied as search ranking, auto-complete inference, and fraud detection. Our hypothetical model training sets are governed by the 7 hyper-parameters laid out in Table 2.

\begin{table}[htbp]
\centering
\small
\begin{tabular}{lcc}
\toprule
\textbf{Hyperparameter} & \textbf{Levels} & \textbf{Values} \\
\midrule
Features ($p$) & 2 & 64, 128 \\
Rank Ratio ($r/p$) & 2 & 0.9, 1.0 \\
Eigenvalue Dispersion ($\kappa$) & 2 & $\sim10^{1}$ (low), $\sim10^{5}$ (high) \\
$\beta$ Distribution & 5 & Gamma(0.04), Gamma(0.2), Gamma(1.0), Gamma(5.0), Uniform \\
Sparsity Level & 2 & 0\%, 15\% \\
Signal-to-Noise Ratio (SNR) & 3 & 0.04, 0.2, 1.0 \\
Sample Size ($n$) & 4 & 100, 1k, 10k, 100k \\
\midrule
\textbf{Total Configurations} & \multicolumn{2}{c}{$2 \times 2 \times 2 \times 5 \times 2 \times 3 \times 4 = \textbf{960}$} \\
\bottomrule
\end{tabular}
\caption{We derive our feature spaces from a space of 7 hyper-parameters.}
\label{tab:hyperparameters}
\end{table}

For those hyperparameters which can be estimated or measured (e.g. \textit{number of feature (p)}, \textit{rank-to-features ratio ($r/p$)}) we grounded the hyperparameter value in the aforementioned sample of 8 ML models in used by Instacart. Other hyper-parameters (e.g. \textit{sparsity}) have no clear `truth set' and so we have selected values that fall in the range an applied researcher is likely to encounter. For other hyper-parameters (e.g. \textit{sample size}), computational limitations were the primary determinant. The ratios of sample size (\textit{n}) to number of features \textit{p} fall within the lower end of the productionized models sampled (typically 50k-500k). While this excludes extremely large-scale scenarios (n/p $>$ 10$^5$), it captures the multicollinearity and ill-conditioning challenges central to regularization evaluation in typical applied ML settings.

\subsection{Parameterization of Spectral Space}
Our first step is to generate suitable eigenvalues which will form the basis for the feature covariance matrices. We use three hyperparameters to structure the creation of these eigenvalues. The first is the number of features used by our hypothetical model training set - the \textit{features}. This hyperparameter can take values of 64, or 128 (these values span the range observed from our empirical sample). The second hyperparameter, the \textit{rank ratio}, is the ratio between the number of non-zero eigenvalues and the number of features. This hyperparameter can take the values of 100\% (full-rank) or 90\% (slight rank deficiency). In the case of a target training dataset of 128 features, we create arrays of eigenvalues of lengths 128 and 115, and then zero-pad the remaining values to match the cardinality of the training dataset.

The third hyper-parameter that governs the creation of eigenvalues is \textit{dispersion}. We use 2 distribution families to create 2 distinct condition number regimes that span realistic machine learning scenarios - low and high eigenvalue dispersion. For low dispersion ($\kappa \approx 10$) we draw Eigenvalues from a Pareto($\alpha=2.0$) distribution, creating a well-conditioned design matrix with mild feature correlation - an ideal scenario with near-orthogonal features. For high dispersion ($\kappa \approx 10^{6}$) we draw Eigenvalues from a Log-Normal($\mu=-2.0$, $\sigma=2.5$) distribution, deliberately creating severe multicollinearity of the kind that can occur with grouped features or polynomial terms. After drawing eigenvalues, we normalize them so their sum equals the target rank $r$. To ensure numerical stability and prevent pathological condition numbers, we apply explicit capping: if the raw condition number $\kappa = \lambda_{\max}/\lambda_{\min} > 10^{6}$ (either before or after normalization), we adjust the smallest eigenvalue to $\lambda_{\min} = \lambda_{\max}/10^{6}$, thereby enforcing $\kappa \leq 10^{6}$. Finally, we replace any numerically negligible eigenvalues ($<10^{-8}$) with exact zeros to avoid marginal eigenvalues that create numerical instability. Note that the actual rank may be marginally less than the target rank after this thresholding.

\begin{figure}[H]
\centering
\includegraphics[width=\textwidth]{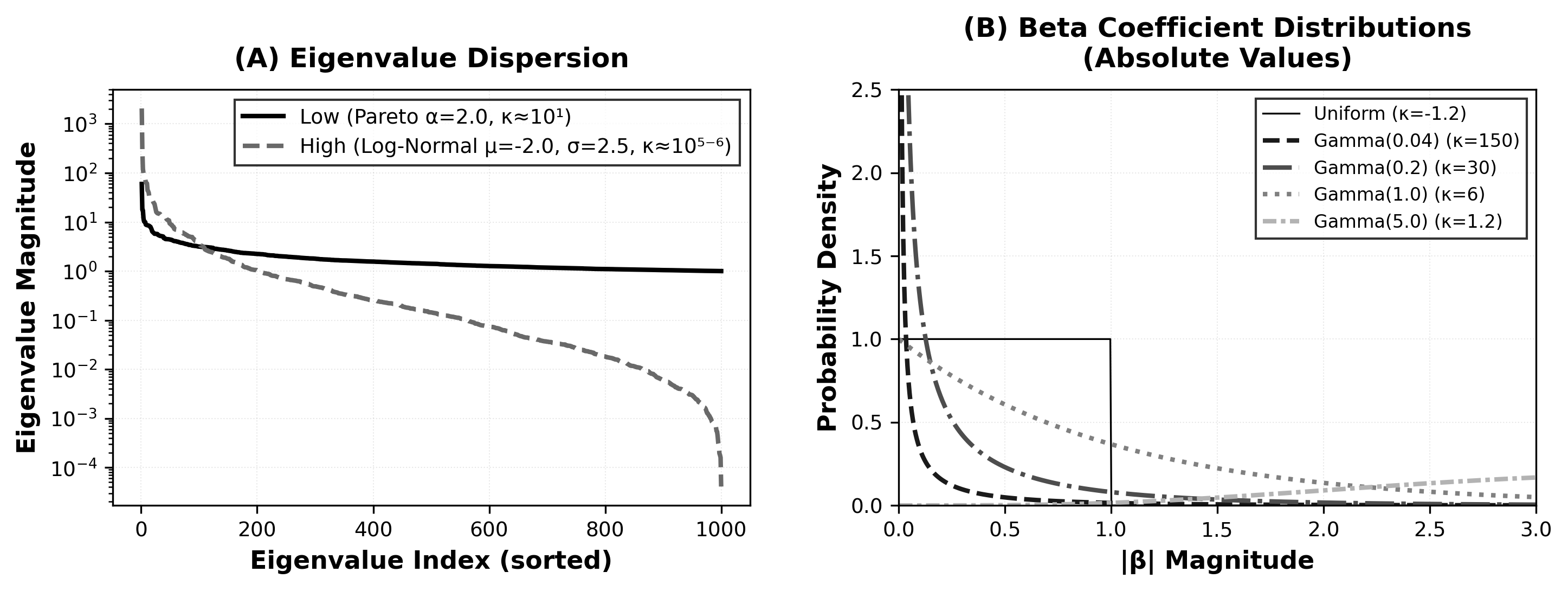}

\vspace{0.5em}

\begin{minipage}[t]{0.48\textwidth}
\subcaption{Eigenvalue dispersion distributions. We draw eigenvalues from one of two distributions: low dispersion (Pareto $\alpha$=2.0, $\kappa\approx10$) versus high dispersion (Log-Normal $\mu$=-2.0, $\sigma$=2.5, $\kappa\approx10^{5-6}$).}
\label{fig:eigenvalue_dispersion}
\end{minipage}%
\hfill
\begin{minipage}[t]{0.48\textwidth}
\raggedright
\subcaption{$\beta$ values are drawn from either a Gamma or a Uniform distribution for a total of 5 distribution / parameter combinations.}
\label{fig:beta_distributions}
\end{minipage}%

\caption{Distribution parameters for Eigenvalues and $\beta$ coefficients.}
\label{fig:dispersion_hyperparameters}
\end{figure}

\subsection{Determination of the Effect Sizes}
Effect sizes ($\beta$) are determined by our 4th hyper-parameter - $\beta$ distribution. We assume that within a typical productionized model training set, most features are weak, a few matter, and the magnitude of feature importance decays smoothly. This is our richest hyper-parameter, comprising of 5 distribution / parameter configurations (see Figure 1B).

A common goal of regularization is the flagging and removal of features where the true effect size is zero. For this reason, we add \textit{sparsity} --- the proportion of model features where $\beta$ is equal to zero - as an additional hyperparameter.

This leads to the delicate question of how best to implement sparsity. The first decision is whether we should implement sparsity at the level of the eigenbasis, or downstream within the feature space. We are interested in the efficacy of regularization in an applied space, so having firm control over the number of model features that are sparse is a requirement. As a consequence of implementing sparsity within the feature space, we lose the ability to isolate effects of the eigenvalue spectrum on regularization performance, and accept this as a limitation. To implement sparsity at the feature level, we randomly set exactly 0\%, or 15\% of regression coefficients to zero in the observed feature space, independent of their magnitude. The full $\beta$ generation process is laid out below.

  \begin{center}
  \begin{algorithm}[H]
  \caption{Beta Coefficient Generation}
  \begin{algorithmic}
  \If{distribution = Uniform}
      \State $\boldsymbol{\beta}_{\text{raw}} \gets \text{random\_choice}(\{-1, +1\}, \text{size}=p)$
  \ElsIf{distribution = Gamma($\alpha$)}
      \State $\mathbf{m} \gets \text{Gamma}(\alpha, 1, \text{size}=p)$ \Comment{Draw magnitudes}
      \State $\mathbf{s} \gets \text{random\_choice}(\{-1, +1\}, \text{size}=p)$ \Comment{Draw signs}
      \State $\boldsymbol{\beta}_{\text{raw}} \gets \mathbf{s} \odot \mathbf{m}$ \Comment{Element-wise product}
  \EndIf
  \If{sparsity $> 0$}
      \State $k \gets \lfloor \text{sparsity} \times p \rfloor$ \Comment{Number of coefficients to zero}
      \State Randomly select $k$ indices $\mathcal{I} \subset \{1, \ldots, p\}$
      \State $\boldsymbol{\beta}_{\text{raw}}[i] \gets 0$ for all $i \in \mathcal{I}$
  \EndIf
  \State Normalize: $\boldsymbol{\beta} \gets \boldsymbol{\beta}_{\text{raw}} \times
  \frac{\sqrt{p}}{\|\boldsymbol{\beta}_{\text{raw}}\|_2}$
  \end{algorithmic}
  \end{algorithm}
  \end{center}

\subsection{Varying the Signal-to-Noise Ratio}
Different regularization frameworks will vary in their efficacy as a function of the underlying \textit{signal-to-noise} (SNR) ratio - our sixth hyper-parameter. We simulate 3 different values of the SNR: 1.0, 0.2, and 0.04. We start by taking the variance of $X\beta$. We then normalize by the SNR to yield the variance of the subsequent i.i.d. Gaussian noise ($\epsilon$) which is then added to the value of the response variable, \textit{y}.

 \begin{center}
  \begin{algorithm}[H]
  \caption{Response Variable Generation}
  \begin{algorithmic}
  \State Generate $\mathbf{X} \sim \mathcal{N}(\boldsymbol{\mu}, \boldsymbol{\Sigma})$
  \State Compute signal: $\mathbf{s} \gets \mathbf{X}\boldsymbol{\beta}$
  \State Compute signal variance: $\sigma^2_{\text{signal}} \gets \text{Var}(\mathbf{s})$
  \State Compute noise variance: $\sigma^2_{\text{noise}} \gets \sigma^2_{\text{signal}} / \text{SNR}$
  \State Generate noise: $\boldsymbol{\epsilon} \sim \mathcal{N}(0, \sigma^2_{\text{noise}} \mathbf{I}_n)$
  \State Combine: $\mathbf{y} \gets \mathbf{s} + \boldsymbol{\epsilon}$
  \end{algorithmic}
  \end{algorithm}
  \end{center}

\subsection{Simulation Implementation}

\renewcommand{\arraystretch}{1.6}
Our final hyper-parameter is \textit{sample size}. This hyper-parameter can take one of 4 values: 100, 1k, 10k, and 100k. Each of the parameter configurations in Table 1 are repeated 35 times with different seeds for a total of 33,600 simulations per regularization framework (LassoCV, Post-Lasso OLS, ElasticNetCV, RidgeCV) for a grand total of 134,400 simulations.

The regularization step requires the selection of 2-3 additional parameters: the possible values for the regularization parameter ($alpha$) and the degrees of cross-validation, and the L1 ratio parameter as used by ElasticNet. We are wary of idiosyncrasies in the implementation of these different scikit-learn classes. For example, RidgeCV does not normalize for sample size as LassoCV does (although from a purely applied perspective, the absence of this normalization is arguably inconsequential \parencite{lemaitre2023ridge}). To ensure that we are evaluating all frameworks on an equal footing, we constrain $alpha$ to one of 9 values across all regularization frameworks. To help keep computation run times at reasonable levels, the we scale the number of CV folds by the sample size, with larger $n$ being given more rigorous cross-validation and vice versa.

\FloatBarrier

%% file: section_5_results.tex
\section{Findings: Coefficient Retrieval and Estimation}
In this section we share our simulation results, prioritizing the $F_{1}$ score to assess the precision-recall trade-off inherent in recovering the true $\beta$ coefficients. Given that parameter consistency is often a primary research objective, we then evaluate estimator accuracy via the relative $L_{2}$ norm of $\hat{\beta}$. While RMSE is a standard benchmark in predictive modeling, we defer its analysis to the Applications section. There, we discuss regularization performance and provide deployment recommendations for practitioners.

\subsection{Validating the Regularization Parameter}
To validate our simulations, we first analyze the $\alpha$ values selected by each regularization framework. When the distribution of $\alpha$ is centered within the permitted simulation range, we can be confident that no framework is inadvertently disadvantaged. As shown in Figure 3, the values are generally well-constrained, with the exception of a small proportion of LassoCV estimates that saturate at the maximum boundary. For context, LassoCV defaults to a sequence of 100 $\alpha$ values, geometrically spaced. The upper bound is defined as:

\begin{equation}\alpha_{\max} = \frac{1}{n} \left| X^T y \right|_{\infty}\end{equation}\noindent

and the lower bound is determined by the $eps$ parameter (defaulting to $10^{-3}$):

\begin{equation}\alpha_{\min} = \alpha_{\max} \cdot \text{eps}\end{equation}

Applying this logic to our simulated data yields a range of $\alpha \in [1.54, 33.0]$. Although our study explores a significantly broader range ($10^{-3}$ to $10^{5}$)—including values over 3,000 times larger than the scikit-learn defaults—we still observe a subset of values at the upper limit. These instances occur almost exclusively in under-determined regions: specifically, 1.56 observations per feature where SNR $\ge 0.2$, or 15.6 observations per feature where SNR $< 0.2$ (see Appendix Figure Figure A.2).This boundary saturation is less a limitation of the research design and more a reflection of the inherent difficulty of the under-determined feature space. Recovering the optimal $\alpha$ becomes computationally expensive and statistically unstable as the feature space becomes increasingly sparse or noise-heavy.

\begin{figure}[!htbp]
\centering
\includegraphics[width=1.0\textwidth]{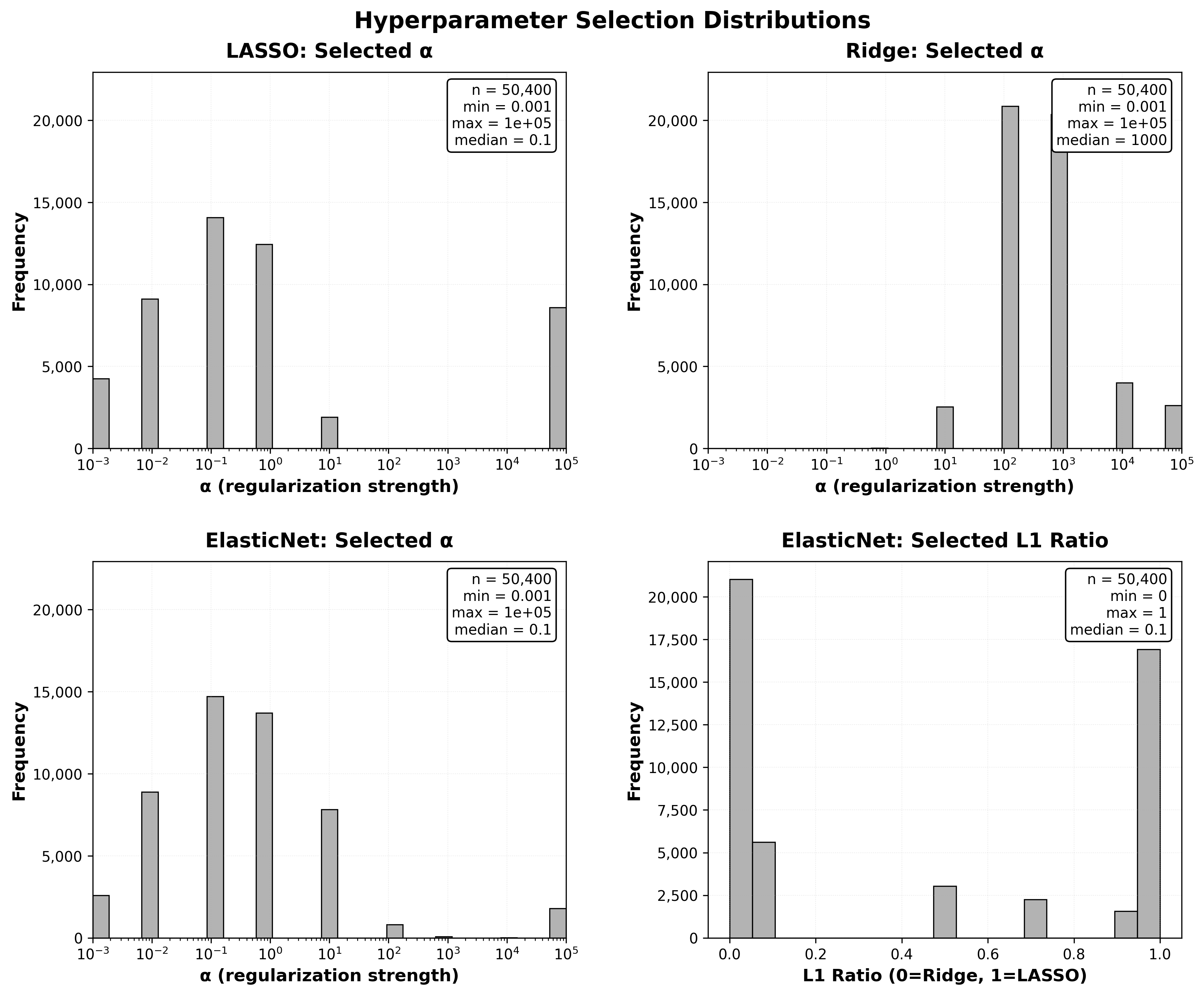}
\caption{We selected a wide range of potential $\alpha$ values to accommodate both RidgeCV and LassoCV. Note how the optimal L1 parameter tends to fall at the extremes (i.e. 0.0 or 1.0).}
\label{fig:distributions_of_hyper_parameters_elected}
\end{figure}

\subsection{Coefficient Recovery (Precision / Recall)}
When the researcher knows \textit{a priori} that all features have a non-zero value for $\beta$, Ridge will yield a recall of 100\% by design. However, a more common scenario is the classic precision/recall trade-off where our goal is to capture as many relevant model features as practical while minimizing the inclusion of random noise.

While Lasso is often favored for its parsimonious feature selection, Figure 4 reveals a significant recall fragility in the presence of noise. While Lasso's precision remains robust (averaging $\approx 0.85$ across all SNR tiers), its recall is highly sensitive to the signal-to-noise ratio. At $SNR=0.04$, Lasso's recall collapses to 0.18–0.20, suggesting that in low-signal regimes, the $\ell_1$  penalty becomes overly aggressive, discarding roughly 80\% of relevant predictors. We see further evidence of this in the extreme values of $\alpha$ recorded during the training process (see Appendix Figure B.1).

\begin{figure}[H]
\centering
\includegraphics[width=0.9\textwidth]{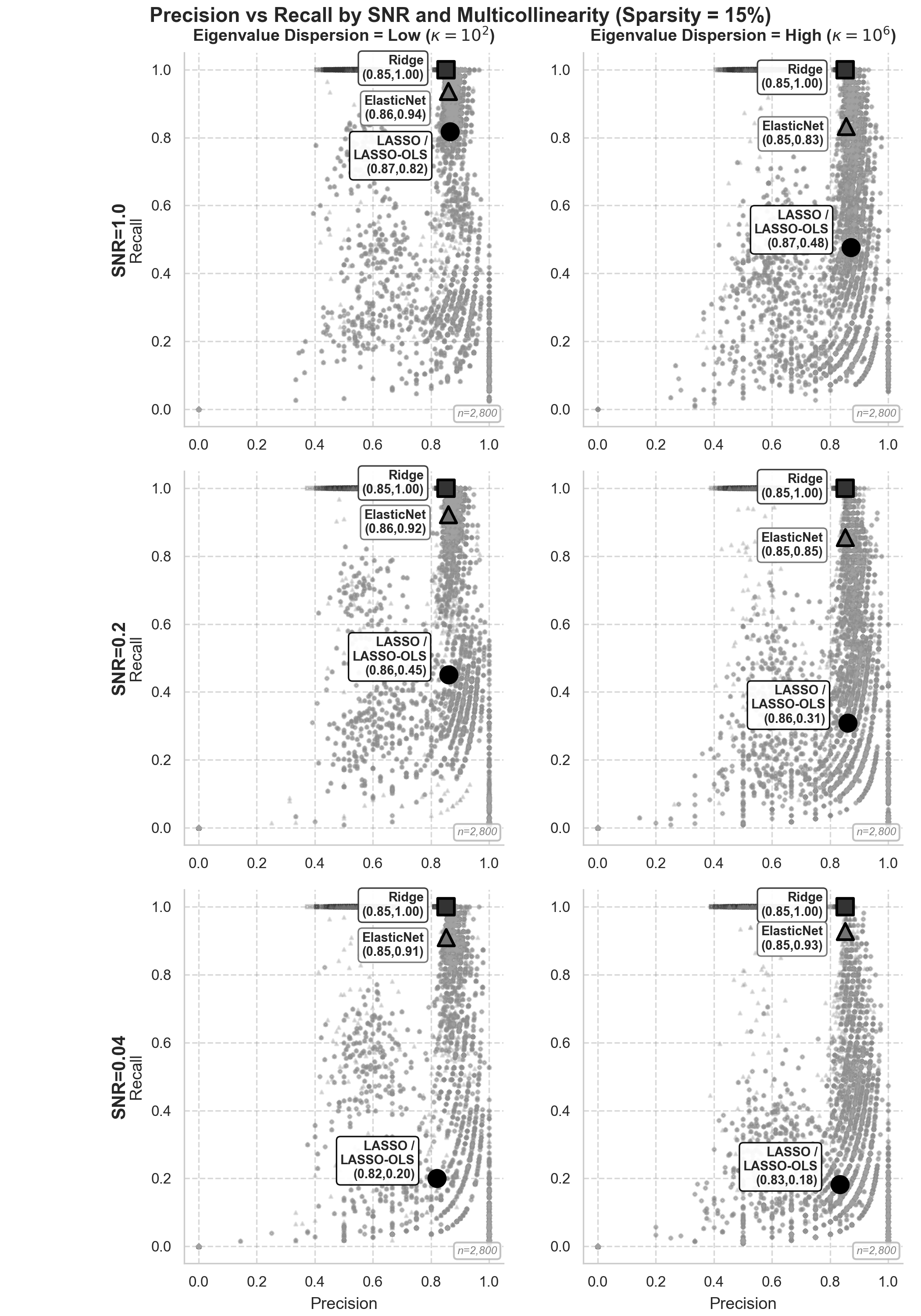}
\caption{The precision with which Lasso is able recover the true, non-zero $\beta$ coefficients is a function of the signal-to-noise ratio within the feature set.}
\label{fig:precision_recall_curves_for_coef_retrieval}
\end{figure}

Conversely, ElasticNet acts as a vital compromise. It preserves the `grouping effect' necessary to maintain high recall ($>0.83$ even in high-dispersion, low-SNR settings) without the total loss of discernment seen in Ridge. Furthermore, the Eigenvalue Dispersion (right column) acts as a performance anchor for Lasso; as multicollinearity increases, Lasso's ability to recover the true support of $\beta$ diminishes by nearly 50\% in high-signal environments.

To assess the relative influence of the 7 hyper-parameters on the efficacy of each regularization framework, we take the differences in the F-1 score and calculate the omega-squared.

\begin{equation}
\omega^2 =
\frac{SS_{\mathrm{effect}} - df_{\mathrm{effect}} \, MS_{\mathrm{error}}}
     {SS_{\mathrm{total}} + MS_{\mathrm{error}}}
\end{equation}

Walking through equation 12, $SS_{\mbox{\footnotesize effect}}$ is the between-group variability, i.e. $\sum n_i(\bar{Y}_i - \bar{Y})^2$ where $\bar{Y}_i$ is the group mean and $\bar{Y}$ is the grand mean (where \textit{i} is the group index). The second term in the numerator can be re-written as follows:

\begin{equation}
(df_{\mathrm{effect}})(MS_{\mathrm{error}}) = (k - 1)\Big(\sum (Y_{ij} - \bar{Yi})^2\Big/(N - k))
\end{equation}

Here \textit{k} is the number of comparisons / factors (the hyper-parameter \textit{sample size} would have \textit{k}=4 (100, 1k, 10k, 100k)). The numerator of the second term is just the sum of squared errors of individual observations vis-\`{a}-vis their group mean, with the denominator normalizing this within-group variability by the number of observations --- \textit{N} --- minus the number of comparisons. The final term --- $SS_{\mbox{\footnotesize total}}$ --- found in the denominator, is the total dataset variability defined as $\sum (Y_{ij} - \bar{Y})^2$. Through this method, the omega-squared statistics reports the proportion of variance associated with one or more main effects while applying a correction to account for the within-group variance. To give a sense of the most relevant interactions across hyper-parameters, we show the f-statistic from two-way Analysis of Variance (ANOVA). The results are shown in table 3 (we use effect size approximations from \cite{cohen1988statistical}).

\begin{table}[htbp]
\centering
\caption{Parameter Importance for F1 Score Differences:
    Main Effects ($\omega^2$) and Interactions (F-statistics)}
\label{tab:f1_parameter_importance}
\small
\begin{tabular*}{\textwidth}{@{\extracolsep{\fill}}lcccc@{}}
\toprule
\textbf{Parameter}
    & \textbf{L--R} & \textbf{L--EN} & \textbf{EN--R}
    & \textbf{Avg} \\
\midrule
Sample Size $(n)$
    & \textbf{0.549} & \textbf{0.284} & \textbf{0.091}
    & \textbf{0.308} \\
SNR
    & \textbf{0.117} & \textbf{0.058} & 0.021
    & \textbf{0.065} \\
$\beta$ Distribution
    & 0.013 & 0.040 & \textbf{0.112}
    & 0.055 \\
Dispersion $(\kappa)$
    & 0.022 & 0.016 & $<$0.001
    & 0.013 \\
Features $(p)$
    & 0.012 & 0.006 & 0.002
    & 0.007 \\
Sparsity
    & 0.006 & 0.002 & 0.001
    & 0.003 \\
Rank Ratio $(r/p)$
    & $<$0.001 & $<$0.001 & $<$0.001
    & $<$0.001 \\
\midrule
\multicolumn{5}{@{}l}{\textit{Two-Way Interactions (F-statistics):}} \\
$n \times$ SNR
    & 573.2*** & 843.8*** & 290.7***
    & 569.2*** \\
$n \times \beta$ Dist.
    & 240.2*** & 35.6*** & 96.1***
    & 123.9*** \\
$n \times \kappa$
    & 231.0*** & 51.7*** & 59.3***
    & 114.0*** \\
\bottomrule
\end{tabular*}
\\[0.5em]
\footnotesize
\textit{Notes:} Column headers: L = LASSO, R = Ridge, EN = ElasticNet.
$\omega^2$ measures proportion of variance in F1 Score \textit{differences}
explained by each parameter (one-way ANOVA); positive differences favour the first method (higher F1 = better).
Effect size thresholds: negligible ($\omega^2 < 0.01$), small ($0.01 \leq \omega^2 < 0.06$),
medium ($0.06 \leq \omega^2 < 0.14$), large ($\omega^2 \geq 0.14$).
Bold values indicate the two largest effect sizes in each column.
Interaction effects shown as F-statistics from two-way ANOVA;
top 3 interactions selected by average $\eta^2$ across comparisons (from all 21 tested).
We exclude LASSO-OLS since its feature selection (and therefore F1) is identical to LASSO. 
Significance: *** $p < 0.001$, ** $p < 0.01$, * $p < 0.05$.
\end{table}

The results from Table 3 summarize how the interplay of hyperparameters  dictate the different regularization frameworks' abilities to differentiate and recover the true $\beta$ coefficients. Sample size ($n$) is by far the most influential factor, explaining the largest proportion of variance in performance differences across all comparisons (Average $\omega^2 = 0.308$). The signal-to-noise ratio represents the second most important parameter for differences between Lasso and Ridge/ElasticNet, though its effect size is significantly smaller than sample size ($\omega^2 = 0.065$). The third most important factor is the $\beta$ distribution. This hyper-parameter has a medium effect size specifically when comparing ElasticNet to Ridge ($\omega^2 = 0.112$), indicating that the underlying distribution of coefficients impacts how these two methods differ.

\subsection{Coefficient Estimate Relative L2 Error}
We can evaluate a regularization procedure's ability to recover the true $\beta$ coefficients via the relative L2 error, i.e. the size of the Euclidean distance between $\beta$ and $\hat{\beta}$ normalized by $\beta$ (see Equation 14). 

\begin{equation}                                            
\text{Relative L2 Error} = \frac{\|\hat{\beta} - \beta\|_2}{\|\beta\|_2}                           = \frac{\sqrt{\sum_{j=1}^{p} \left(\hat{\beta}_j - \beta_j\right)^2}}{\sqrt{\sum_{j=1}^{p} \beta_j^2}}         
\end{equation}

\noindent The full results are laid out in the Appendix (Figures D.1 through D.6), with the omega-squared statistics and f-statistics from the 3 largest two-way iterations laid on Table 4 below.

\begin{table}[htbp]
\centering
\caption{Parameter Importance for Coefficient L2 Error Differences:
    Main Effects ($\omega^2$) and Interactions (F-statistics)}
\label{tab:l2_error_parameter_importance}
\small
\begin{tabular*}{\textwidth}{@{\extracolsep{\fill}}lccccccc@{}}
\toprule
\textbf{Parameter}
    & \textbf{R--L} & \textbf{EN--L} & \textbf{PL--L}
    & \textbf{R--EN} & \textbf{PL--EN} & \textbf{R--PL}
    & \textbf{Avg} \\
\midrule
$\beta$ Distribution
    & \textbf{0.2155} & \textbf{0.0773} & 0.0024
    & \textbf{0.1428} & 0.0024 & 0.0024
    & \textbf{0.0738} \\
Sample Size $(n)$
    & \textbf{0.0133} & 0.0031 & 0.0015
    & \textbf{0.0248} & 0.0015 & 0.0015
    & \textbf{0.0076} \\
Dispersion $(\kappa)$
    & 0.0020 & \textbf{0.0099} & \textbf{0.0026}
    & 0.0007 & \textbf{0.0026} & \textbf{0.0026}
    & 0.0034 \\
Rank Ratio $(r/p)$
    & 0.0001 & 0.0015 & \textbf{0.0027}
    & 0.0024 & \textbf{0.0027} & \textbf{0.0027}
    & 0.0020 \\
SNR
    & 0.0018 & 0.0010 & 0.0001
    & 0.0056 & 0.0001 & 0.0001
    & 0.0015 \\
Sparsity
    & 0.0007 & 0.0004 & $<$0.0001
    & 0.0002 & $<$0.0001 & $<$0.0001
    & 0.0002 \\
Features $(p)$
    & $<$0.0001 & $<$0.0001 & $<$0.0001
    & $<$0.0001 & $<$0.0001 & $<$0.0001
    & $<$0.0001 \\
\midrule
\multicolumn{8}{@{}l}{\textit{Two-Way Interactions (F-statistics):}} \\
$n \times \beta$ Dist.
    & 98.3*** & 43.5*** & 3.9***
    & 67.8*** & 3.9*** & 3.9***
    & 36.9*** \\
$n \times$ SNR
    & 65.6*** & 80.0*** & 7.9***
    & 52.3*** & 7.9*** & 7.9***
    & 36.9*** \\
$\kappa \times \beta$ Dist.
    & 103.3*** & 36.2*** & 19.7***
    & 56.3*** & 19.7*** & 19.7***
    & 42.5*** \\
\bottomrule
\end{tabular*}
\\[0.5em]
\footnotesize
\textit{Notes:} Column headers: L = LASSO, R = Ridge, EN = ElasticNet, PL = Post-Lasso OLS.
$\omega^2$ measures proportion of variance in Coefficient L2 Error \textit{differences}
explained by each parameter (one-way ANOVA).
Effect size thresholds: negligible ($\omega^2 < 0.01$), small ($0.01 \leq \omega^2 < 0.06$),
medium ($0.06 \leq \omega^2 < 0.14$), large ($\omega^2 \geq 0.14$).
Bold values indicate the two largest effect sizes in each column.
Interaction effects shown as F-statistics from two-way ANOVA;
top 3 interactions selected by average $\eta^2$ across comparisons (from all 21 tested).
Significance: *** $p < 0.001$, ** $p < 0.01$, * $p < 0.05$.
\end{table}

When it comes to regularization performance with respect to the L2 error rate, we see two primary dynamics at play (note how the largest omega-squared statistics vary as a function of whether post-Lasso OLS is an option). The first dynamic is the extent Lasso is able to successfully recover all relevant features. Lasso's correctness here has profound implications for the effectiveness of post-Lasso OLS. Whereas Ridge and ElasticNet distribute regularization across groups of coefficient estimates, post-Lasso OLS attempts to fit the selected subset without any shrinkage. As a result, post-Lasso OLS simultaneously unlocks the best possible L2 error rate, but fails catastrophically if the initial Lasso stage fails to recover the correct features due to the underlying feature geometry being `unfriendly' (Eigenvalues are highly concentrated and /or rank is low). Specifically, in high-$\kappa$ environments, the unregularized matrix inversion $({X_S}^T X_S)^{-1}$ in the OLS stage amplifies noise exponentially, leading to the massive L2 errors (see the ``darker" regions of the heatmaps in Figures D.1, D.2 and D.5). To summarize, the relative efficacy of post-Lasso OLS is primarily dictated by the geometry and stability of the feature space (versus the signal strength ($SNR$) or feature sparsity). Because post-LASSO OLS is fundamentally at the mercy of the first stage's false negative rate (FNR) and FDR, we consider post-Lasso OLS a high-risk / high-reward framework --- one that requires careful understanding of the feature eigenspace. 

The second dynamic is the interaction between our two most influential hyper-parameters as laid out in Table 4---the distribution of the $\beta$  coefficients, and the sample size of the training data. Not just the catastrophic failures of post-Lasso OLS described above, but the weakness of Ridge vis-\'a-vis ElasticNet and Ridge vis-\'a-vis Lasso follow a pattern whereby Ridge under-performs when the distribution of the $\beta$  coefficients is highly peaked, but emerges as the stronger framework when the $\beta$  coefficients are more uniformly dispersed. While this is expected, it is notable that the regularization frameworks' performances with respect to the L2 error become more distinguishable as the $n/p$ ratio \textit{increases} (this is in contrast to the test RMSE where higher $n/p$ ratios tend to dampen the difference in regularization performance across frameworks --- more in the next section).

This second dynamic is a manifestation of the competing pathways of regularization failure: (i.) collapse of coefficient retrieval via extreme values of the regularization parameter ($\alpha$), and (ii.) incorrect coefficient retrieval when the distribution of $\beta$ coefficients is approximately uniform. Figure B.1 (see the appendix) highlights how, when the feature space is under-determined, Lasso / has a tendency to assign extremely high values to $\alpha$ --- the consequence being that few, if any, coefficients are recovered. The silver lining of this type of failure is that the resulting model has so few features that there is a much smaller window for biased coefficient estimation.

\begin{figure}[!htbp]
\centering
\includegraphics[width=1.0\textwidth]{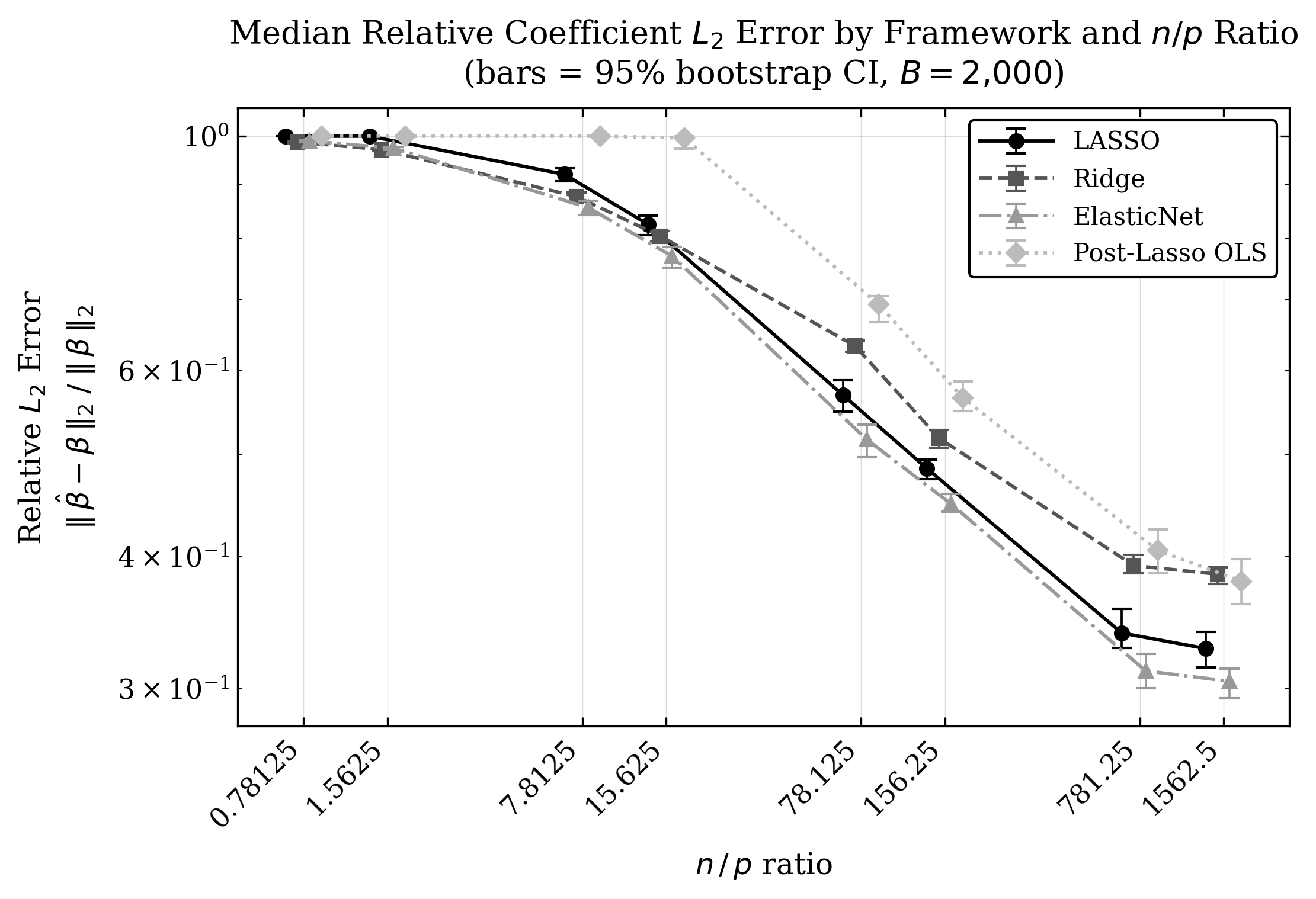}
\caption{Post-Lasso OLS has the largest L2 error across the 8 $n/p$ ratios simulated.}
\label{fig:95_ci_plot_of_l2_error_np_ratio}
\end{figure}

In summary, researchers interested in optimizing for the accuracy of coefficient estimates are advised to start by evaluating the eigenspace of the feature set. Post-lasso OLS is only likely to be the best regularization strategy when the feature space is exceptionally friendly --- full rank, evenly-distributed eigenvalues, and even then, the $\beta$s may be too uniformly distributed to make correct retrieval viable. Should researchers elect not to use post-Lasso OLS, ElasticNet will tend to yield the best results assuming $n/p$ ratios $>78$ (i.e., the model is not under-determined).   
\FloatBarrier

%% file: section_6_applications.tex
\section{Applications}

\label{sec:applications}

The preceding sections characterize how four regularization frameworks---RidgeCV, LassoCV, ElasticNetCV, and Post-Lasso OLS---perform across 134,400 simulations spanning a 7-dimensional manifold of feature space configurations. The findings are grounded in fully observable, controlled parameters: we know the true SNR, the true sparsity, and the true coefficient vector $\beta$ for every simulation. The applied practitioner enjoys no such luxury.

In this section, we bridge this gap. We start with an overview of how the four regularization frameworks perform with respect to predictive accuracy---a quality of paramount importance to MLEs. We then distill these results into a Practitioner's Guide---a set of decision rules that branch exclusively on quantities a Data Scientist / MLE can compute before fitting a single model. The goal is to minimize out-of-sample error and computational waste by routing the practitioner to the regularization framework best supported by the evidence for their observable data regime.

\subsection{Evaluating Implications for Predictive Power (RMSE)}

While predictive accuracy is the primary benchmark for machine learning frameworks, its sensitivity to model specification depends heavily on data availability. As illustrated in Figure 6, the performance gap between regularization strategies diminishes as sample size increases, suggesting an asymptotic convergence in estimator efficiency. Conversely, in data-constrained environments, the choice of regularization becomes a critical determinant of model generalization, as the penalty term must compensate for the increased risk of overfitting inherent in small datasets.

\begin{figure}[!htbp]
\centering
\includegraphics[width=1.0\textwidth]{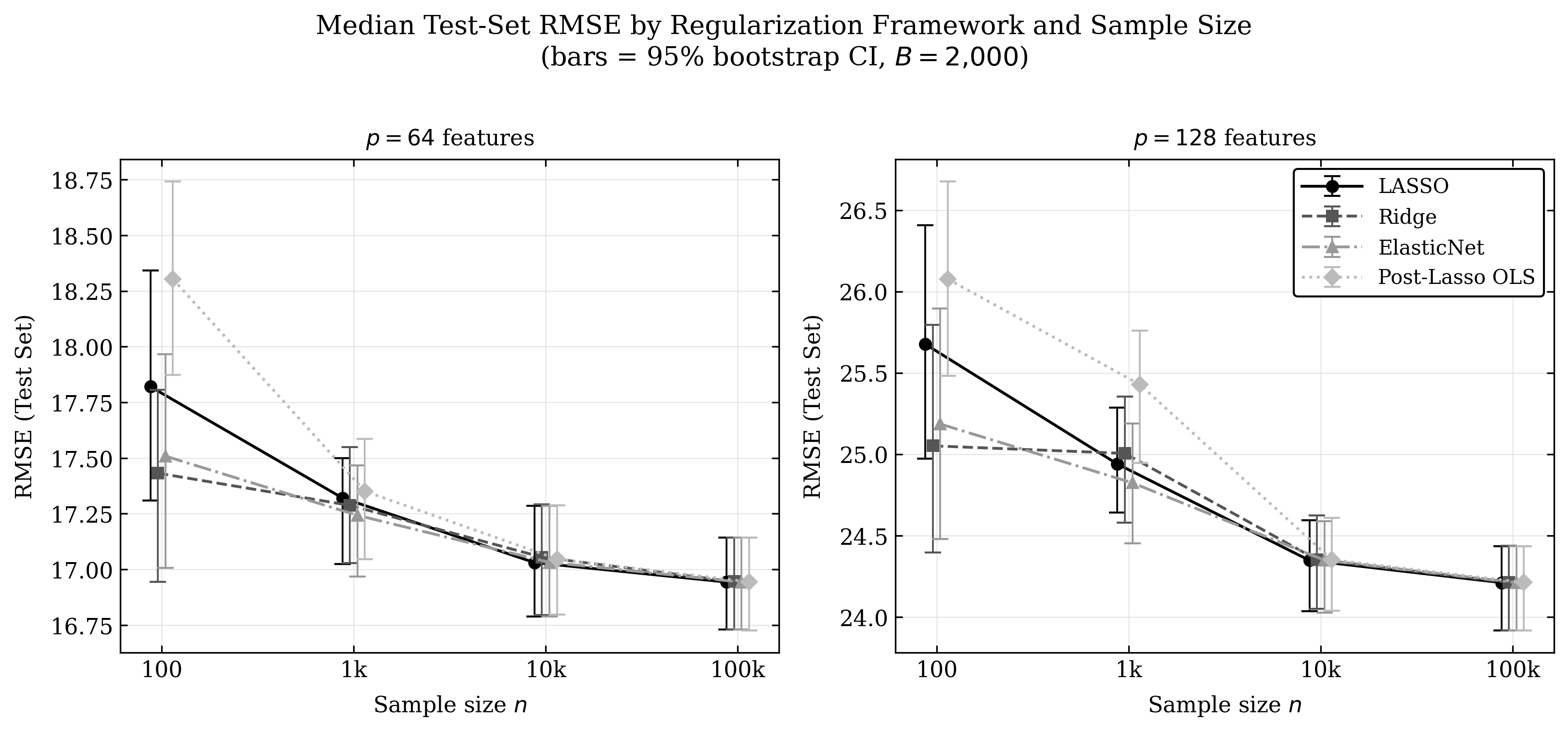}
\caption{ElasticNetCV was most likely to yield the smallest RMSE within the test set.}
\label{fig:95_ci_plot_of_test_rmse}
\end{figure}

As with the F1 and L2 error statistics from the previous section, the omega-squared statistics are presented below in Table 5 alongside the three largest f-statistics from the interactions between hyperparameters. We can see that, just as Figure 6 suggests, sample size is going to be the most salient predictor of which regularization framework is most performant, followed by the number of features. Looking into the results in more detail (see figures E1-E6 in the appendix), we can confirm that immediately following the $n/p$ ratio, the distribution of $\beta$ coefficients is the most influential hyperparameter. Specifically, when we are in under-determined space ($n/p <$ 7.8) this hyperparameter takes over, determining whether Lasso / Ridge / ElasicNet nets the lowest RMSE in the test set. We explore the implications of these results in more detail in the subsequent section.  
   
\begin{table}[H]
\centering
\caption{Parameter Importance for RMSE Differences:
    Main Effects ($\omega^2$) and Interactions (F-statistics)}
\label{tab:method_superiority_omega2}
\small
\begin{tabular*}{\textwidth}{@{\extracolsep{\fill}}lccccccc@{}}
\toprule
\textbf{Parameter}
    & \textbf{R--L} & \textbf{EN--L} & \textbf{PL--L}
    & \textbf{R--EN} & \textbf{PL--EN} & \textbf{R--PL}
    & \textbf{Avg} \\
\midrule
Sample Size $(n)$
    & \textbf{0.0077} & \textbf{0.0110} & \textbf{0.0893}
    & 0.0005 & \textbf{0.0930} & \textbf{0.0800}
    & \textbf{0.0469} \\
Features $(p)$
    & 0.0007 & 0.0007 & \textbf{0.0047}
    & $<$0.0001 & \textbf{0.0052} & \textbf{0.0047}
    & \textbf{0.0027} \\
$\beta$ Distribution
    & \textbf{0.0068} & 0.0014 & 0.0001
    & \textbf{0.0051} & $<$0.0001 & 0.0011
    & 0.0024 \\
SNR
    & 0.0020 & \textbf{0.0031} & 0.0032
    & \textbf{0.0014} & 0.0006 & 0.0014
    & 0.0020 \\
Dispersion $(\kappa)$
    & 0.0011 & 0.0003 & 0.0007
    & 0.0006 & 0.0002 & $<$0.0001
    & 0.0005 \\
Rank Ratio $(r/p)$
    & $<$0.0001 & $<$0.0001 & 0.0002
    & $<$0.0001 & $<$0.0001 & 0.0001
    & $<$0.0001 \\
Sparsity
    & $<$0.0001 & $<$0.0001 & $<$0.0001
    & $<$0.0001 & $<$0.0001 & $<$0.0001
    & $<$0.0001 \\
\midrule
\multicolumn{8}{@{}l}{\textit{Two-Way Interactions (F-statistics):}} \\
$n \times$ SNR
    & 83.0*** & 119.3*** & 48.6***
    & 28.5*** & 13.4*** & 9.9***
    & 50.4*** \\
$n \times p$
    & 14.5*** & 5.9*** & 108.3***
    & 15.4*** & 92.7*** & 105.5***
    & 57.1*** \\
$n \times \beta$ Dist.
    & 15.0*** & 3.9*** & 1.2
    & 9.9*** & 0.3 & 2.6**
    & 5.5*** \\
\bottomrule
\end{tabular*}
\\[0.5em]
\footnotesize
\textit{Notes:} Column headers: L = LASSO, R = Ridge, EN = ElasticNet, PL = Post-Lasso OLS.
$\omega^2$ measures proportion of variance in RMSE \textit{differences}
explained by each parameter (one-way ANOVA); negative differences favour the first method (lower RMSE = better).
Effect size thresholds: negligible ($\omega^2 < 0.01$), small ($0.01 \leq \omega^2 < 0.06$),
medium ($0.06 \leq \omega^2 < 0.14$), large ($\omega^2 \geq 0.14$).
Bold values indicate the two largest effect sizes in each column.
Interaction effects shown as F-statistics from two-way ANOVA;
top 3 interactions selected by average $\eta^2$ across comparisons (from all 21 tested).
Significance: *** $p < 0.001$, ** $p < 0.01$, * $p < 0.05$.
\end{table}

\subsection{The Applied Challenge: Observable vs.\ Latent Parameters}
\label{sec:observable_vs_latent}

A fundamental asymmetry separates simulation studies from real-world applications. In our simulation design, every generative parameter---the Signal-to-Noise Ratio (SNR), the true sparsity level, the condition number ($\kappa$), the $\beta$ distribution, and the rank ratio---is known by construction. In a production environment, the majority of these quantities are latent and unobservable. A MLE building a demand forecasting model or a fraud detection pipeline cannot query the data-generating process for the true SNR or the fraction of coefficients that are exactly zero.

Specifically, the parameters most critical for differentiating regularization performance fall into two categories: those that are directly computable and those that can be approximated via inexpensive diagnostics.

\paragraph{Directly observable parameters.}
\begin{itemize}[nosep,leftmargin=*]
  \item Sample size ($n$): Known exactly.
  \item Number of features ($p$): Known exactly.
  \item Condition number ($\kappa$) of the design matrix $X$: Computable via \texttt{numpy.linalg.cond(X)}, providing a concrete measure of eigenvalue dispersion and multicollinearity. Our simulations tested two regimes---low dispersion ($\kappa \approx 10^1$) and high dispersion ($\kappa \approx 10^5$--$10^6$)---that span the range observed in a sample of eight production ML models (Appendix~A, Figure~A.1). The thresholds used in this section ($\kappa > {\sim}10^4$ and $\kappa < {\sim}10^2$) are interpolated from these two tested extremes. Practitioners with intermediate condition numbers ($10^2 < \kappa < 10^4$) should treat the recommendations with additional caution, as this range was not directly evaluated.
\end{itemize}

\newpage
\paragraph{Latent parameters with diagnostic proxies.}
\begin{itemize}[nosep,leftmargin=*]
  \item Signal-to-Noise Ratio (SNR): Not directly observable, but the regularization strength $\alpha$ elected by LassoCV acts as a reliable proxy.\footnote{\textcite{lederer2021lasso} define the term $2\|X^T \varepsilon\|_\infty / n$---the maximum correlation between the design matrix and the noise vector---as the ``effective noise.'' In this framework, the regularization parameter $\alpha$ must be at least as large as the effective noise to ensure the Lasso recovers only true variables (\cite[Lemma~6.1]{buhlmann2011highdim}). Since the magnitude of this noise term scales with the noise standard deviation $\sigma$, we expect $\alpha$ to be inversely correlated with the SNR (defined in Section~4 as $\text{SNR} = \text{Var}(X\beta) / \sigma^2$).} When the CV procedure selects an extremely high $\alpha$---or saturates at the maximum boundary defined by $\alpha_{\max} = \frac{1}{n}\|X^{T}y\|_{\infty}$---it signals a low-SNR or under-determined regime. Figure~B.1 provides direct empirical confirmation: the highest elected $\alpha$ values (approaching $10^4$--$10^5$) concentrate almost exclusively in the small-$n$, low-SNR rows of our simulation grid.
  \item Sparsity: Not directly observable, but domain expertise provides strong priors (see Section~\ref{sec:diagnostics}).
\end{itemize}

\paragraph{Which parameters matter most?}
Not all seven simulation hyperparameters contribute equally to performance differentiation. Tables~\ref{tab:f1_parameter_importance} and~\ref{tab:method_superiority_omega2} report $\omega^2$ (omega-squared) effect sizes from one-way ANOVA on pairwise performance differences, providing a rigorous ranking:

\begin{itemize}[nosep,leftmargin=*]
  \item For variable selection (F1 score), sample size $n$ is overwhelmingly dominant (average $\omega^2 = 0.308$ across all pairwise comparisons; Table~\ref{tab:f1_parameter_importance}), constituting a large effect size under standard thresholds \parencite{cohen1988statistical}. The latent SNR is the second most important factor overall (average $\omega^2 = 0.065$). The $\beta$ distribution achieves a medium effect size for the ElasticNet--Ridge comparison specifically ($\omega^2 = 0.112$; Table~\ref{tab:f1_parameter_importance}), making it the third most important factor overall and indicating that the relative advantage of ElasticNet over Ridge for variable selection depends in part on the unobservable shape of the coefficient vector. The condition number $\kappa$ is the strongest \emph{observable} secondary factor (average $\omega^2 = 0.013$), followed by features $p$ (average $\omega^2 = 0.007$). Sparsity ($\omega^2 = 0.003$) and rank ratio ($\omega^2 = 0.000$) are negligible.
  
  \item For predictive accuracy (RMSE), the picture is strikingly different. Among the three core methods---Ridge, Lasso, and ElasticNet---nearly all pairwise $\omega^2$ values fall below 0.01, with the single exception being Sample Size for the ElasticNet--Lasso comparison ($\omega^2 = 0.011$), which remains well below the threshold for a medium effect size (Table~\ref{tab:method_superiority_omega2}: all values $< 0.02$ for R--L, EN--L, R--EN). This is itself a critical finding: it confirms that the three canonical methods are nearly interchangeable on prediction, and that the practitioner's decision should be driven by secondary objectives (variable selection, coefficient estimation) and computational cost. Sample size achieves meaningful effect sizes only in comparisons involving Post-Lasso OLS ($\omega^2 = 0.0893$ for LO--L, $0.0930$ for LO--EN, $0.0800$ for R--LO), reflecting that method's unique vulnerability at small~$n$.
\end{itemize}

\paragraph{Two-way interactions.}
Two-way interactions reinforce this hierarchy. The $n \times \text{SNR}$ interaction is the strongest across all comparisons (average $F = 569$, $p < 0.001$; Table~\ref{tab:f1_parameter_importance}), confirming that the impact of the underlying signal-to-noise ratio on method differentiation intensifies as sample size decreases. This interaction empirically justifies the structure of the guide that follows: the SNR diagnostic (Section~\ref{sec:diagnostics}) becomes most consequential precisely in the small-$n$ regime where it is deployed. The $n \times \beta$ distribution interaction (average $F = 124$, $p < 0.001$) and $n \times \kappa$ interaction (average $F = 114$, $p < 0.001$) are secondary (Table~\ref{tab:f1_parameter_importance}).



\paragraph{The large-$n/p$ simplification.}
The dominance of sample size yields a powerful practical corollary. While Tables~\ref{tab:f1_parameter_importance} and~\ref{tab:method_superiority_omega2} report $n$ and $p$ as separate main effects, the significant $n \times p$ interaction (average $F = 57.1$, $p < 0.001$; Table~\ref{tab:method_superiority_omega2}) and Figure~5 confirm that the operative quantity is the ratio $n/p$, not $n$ in isolation. Across all RMSE heatmaps (Figures~E.1--E.6), F1 heatmaps (Figures~C.1--C.3), and coefficient $L_2$ error heatmaps (Figures~D.1--D.6), the cells corresponding to $n/p \geq 78$ (i.e., $n/p \geq 78$ for the $p \leq 128$ tested here) are uniformly near zero relative difference---regardless of $\kappa$, $\beta$ distribution, sparsity, or rank ratio. At $n/p \geq 78$, the performance gaps between all methods effectively vanish across all three metrics.

In this large-sample regime, the decision should be driven by computational efficiency and functional requirements. Figure~F.1 reports fit times at $n = 100{,}000$: Ridge achieves mean fit times of 0.03 minutes ($p = 64$) and 0.15 minutes ($p = 128$). Lasso is comparable at the median but exhibits heavier tails (mean $= 0.68$ minutes at $p = 128$). ElasticNet, by contrast, incurs mean fit times of 6.57 minutes ($p = 64$) and 25.11 minutes ($p = 128$)---167$\times$ and 219$\times$ slower than Ridge by mean, or 5--23$\times$ by median (the mean is inflated by heavy-tailed CV runs)---due to its joint grid search over $\alpha$ and the L1 ratio $\rho$.

Important caveat: The 167--219$\times$ mean overhead is specific to the 8-value L1 ratio grid used in our simulations (Table~\ref{tab:model_comparison}). Because ElasticNet's additional cost relative to Lasso scales linearly with the number of L1 ratio candidates, a practitioner using a coarser 3-value grid (e.g., $\rho \in \{0.1, 0.5, 0.9\}$) would incur approximately $3\times$ overhead rather than $167$--$219\times$. Per-CV-fit compute times---rather than per-full-grid-search times---provide a more intrinsic comparison; on this basis ElasticNet's per-fit cost is comparable to Lasso's. For models that are retrained daily or hourly, total grid search overhead nonetheless translates directly into infrastructure cost and pipeline latency.

\paragraph{The large-$n/p$ rule.}
If $n/p \geq 78$, performance differences across all methods are negligible, and the decision should be guided by computational efficiency and the practitioner's functional requirements. For prediction and coefficient estimation, RidgeCV is the recommended default due to its superior runtime profile (Figure~F.1). For variable selection, ElasticNetCV or LassoCV remain appropriate if the practitioner requires an explicitly sparse model, since Ridge assigns nonzero coefficients to all features by construction. At this ratio, the choice among sparse methods (Lasso vs.\ ElasticNet) is immaterial.

The remainder of this section is therefore most consequential for regimes with $n/p < 78$, where the differences documented in Section~5 are most pronounced and the choice of regularization framework has material impact on all three objectives.

\paragraph{The under-determined regime ($p > n$).}
Our simulation grid includes configurations where $n = 100$ with $p \in \{64, 128\}$, yielding under-determined ($p > n$) or near-under-determined regimes. These configurations produce the most extreme performance differences: Lasso's $\alpha$ frequently saturates at the upper boundary of the search grid (see Appendix~B, Figure~B.1), and recall collapses to 0.18--0.20 (Figure~\ref{fig:precision_recall_curves_for_coef_retrieval}). When $p > n$, practitioners should exercise particular caution with Lasso and default to Ridge or ElasticNet across all objectives. The flowchart in Figure~\ref{fig:decision_flowchart} now includes an explicit $p > n$ check as the first decision node.

\subsection{An Objective-Driven Guide}
\label{sec:objective_driven_guide}

The recommendations below branch on two observable quantities that our simulations identify as the most empirically consequential: the sample-to-feature ratio ($n/p$) and the condition number ($\kappa$) of the design matrix, computable via \texttt{numpy.linalg.cond(X)}. A tertiary diagnostic---the CV-elected $\alpha$ from LassoCV---is used as a proxy for the latent SNR when finer discrimination is needed (see Section~\ref{sec:diagnostics}). Figure~\ref{fig:decision_flowchart} provides a consolidated decision flowchart.

We structure the guide around three canonical practitioner objectives.

\subsubsection*{Path A: Priority is Predictive Accuracy (Minimizing RMSE)}
\label{sec:path_a}

The overriding finding for prediction is that the three core methods---Ridge, Lasso, and ElasticNet---are nearly interchangeable. The median test RMSE across 33,600 simulations differs by at most 0.3\% between any pair (Figure~\ref{fig:95_ci_plot_of_test_rmse}: ElasticNet $= 25.67$, Ridge $= 25.68$, Lasso $= 25.74$). Table~\ref{tab:method_superiority_omega2} confirms this: no single hyperparameter achieves even a small effect size ($\omega^2 \geq 0.01$) for RMSE differences among these three methods. Every pairwise $\omega^2$ is negligible ($< 0.02$).

We emphasize that these aggregate statistics are computed across all 960 configurations, including many large-$n$ regimes where all methods trivially converge. At small sample sizes ($n \leq 1{,}000$), conditional RMSE differences can be substantially larger---on the order of 5--15\% in specific cells of the heatmaps (Figures~E.1, E.4, E.5). The recommendations below therefore include explicit small-$n$ caveats.

Given this near-equivalence in accuracy, the decision reduces to computational efficiency and robustness:

\begin{itemize}[nosep,leftmargin=*]
  \item Default: use RidgeCV. Ridge achieves the lowest or near-lowest mean fit time across all configurations (Figure~F.1: mean $= 0.03$\,min at $p = 64$, $0.15$\,min at $p = 128$ for $n = 100$k). Its closed-form solution for each candidate $\alpha$ makes runtime predictable and stable.

  \item Avoid ElasticNetCV unless secondary objectives demand it. ElasticNet's joint grid search over $\alpha$ and the L1 ratio $\rho$ imposes a computational penalty whose magnitude depends on the L1 ratio grid size. In our simulations (8-value grid), mean fit times were 6.57\,min at $p = 64$ and 25.11\,min at $p = 128$ (Figure~F.1)---167$\times$ and 219$\times$ slower than Ridge \emph{by mean}, respectively. By median, the overhead is substantially smaller (5--23$\times$), as ElasticNet's mean is inflated by heavy-tailed runs under extended CV grids. With a coarser 3-value grid, the overhead would be substantially lower than the 167--219$\times$ mean reported here (see discussion above). Regardless of grid size, this overhead yields a median RMSE improvement of just 0.04\% over Ridge, a margin that is negligible in virtually all applied contexts.
  
  \item Avoid Post-Lasso OLS. Post-Lasso OLS is the only method that separates meaningfully from the pack on RMSE, and it does so in the wrong direction: its median RMSE is 1.4\% higher than ElasticNet's (Figure~\ref{fig:95_ci_plot_of_test_rmse}). This degradation is concentrated at small $n$ (Figures~E.2, E.3, E.6), where the unpenalized OLS refit amplifies first-stage selection errors. Sample size is the dominant driver of this gap ($\omega^2 = 0.0893$ for LO--L, $0.0930$ for LO--EN; Table~\ref{tab:method_superiority_omega2}).
  
  \item The one scenario where method choice matters for RMSE: At $n = 100$ (the smallest sample size tested), the RMSE heatmaps (Figures~E.1, E.4, E.5) show visible but modest differences. Although $\kappa$ does not achieve a meaningful aggregate effect size for RMSE differences ($\omega^2 < 0.002$ across all comparisons; Table~\ref{tab:method_superiority_omega2}), inspection of the $n = 100$ cells reveals localized effects. Figure~E.4 (Ridge vs.\ Lasso) shows Ridge achieving lower RMSE in the majority of $n = 100$ cells, with the advantage most pronounced at low SNR. Figure~E.5 (Ridge vs.\ ElasticNet) shows ElasticNet achieving lower RMSE by 5--15\% at $n = 100$ with high SNR (SNR $\approx 1.0$); critically, this effect appears at \emph{both} $\kappa$ levels equally---confirming that SNR, not $\kappa$, is the operative driver. The same figure shows teal cells (Ridge marginally better than EN) at $n = 100$ with very low SNR ($\approx 0.04$), reinforcing Ridge as the default when signal is weak. At $n/p \geq 78$, all cells are near-zero across all six RMSE heatmaps. These patterns should be understood as cell-level observations rather than systematic trends.
\end{itemize}

\noindent Summary for prediction: Use Ridge. The RMSE differences among the core three methods are too small to justify computational overhead or added complexity. \emph{Exception:} at $n \approx 100$ with high SNR (SNR $\approx 1.0$), ElasticNet offers a detectable 5--15\% edge regardless of $\kappa$; at $n \approx 100$ with very low SNR ($\approx 0.04$), Ridge is marginally preferred. In either case, the margin is modest relative to the improvement available from increasing sample size.

\subsubsection*{Path B: Priority is Variable Selection (Maximizing F1 / Precision \& Recall)}
\label{sec:path_b}

Variable selection is where the choice of regularization framework matters most, and where our simulations provide the clearest differentiation. Sample size ($n$) explains the largest share of variance in F1 score differences (average $\omega^2 = 0.308$; Table~\ref{tab:f1_parameter_importance}), followed by the latent SNR ($\omega^2 = 0.065$) and $\kappa$ ($\omega^2 = 0.013$). The $n \times \text{SNR}$ interaction (average $F = 569$, $p < 0.001$; Table~\ref{tab:f1_parameter_importance}) confirms that SNR's influence on method differentiation intensifies as $n$ decreases. Among observable parameters, $n$ and $\kappa$ are the two strongest drivers.

We note that the $\beta$ distribution---while latent and therefore excluded from the branching logic---achieves a medium effect size ($\omega^2 = 0.112$) for the ElasticNet--Ridge comparison specifically (Table~\ref{tab:f1_parameter_importance}). This indicates that the relative advantage of ElasticNet over Ridge for variable selection depends in part on the unobservable shape of the coefficient vector, introducing a source of variation not captured by the flowchart.

\paragraph{Branch 1---Sample-to-feature ratio ($n/p$).}
\begin{itemize}[nosep,leftmargin=*]
  \item If $n/p \geq 78$: Figures~C.1, C.2, and C.3 show near-zero $\Delta$F1 across all $\kappa$ and $\beta$ distribution cells at $n = 10$k and $n = 100$k. All sparse methods achieve high F1. Use ElasticNet as the safe default for its stable recall, or Lasso if parsimony is strongly preferred---both perform well. Ridge is not recommended here despite its competitive F1 scores, because it achieves them with recall $= 1.0$ (all features retained) rather than through genuine variable selection.
  \item If $n/p < 78$: Differences become substantial. Proceed to Branch~2.
\end{itemize}

\paragraph{Branch 2---Condition number ($\kappa$).}
\begin{itemize}[nosep,leftmargin=*]
  \item If $\kappa$ is high (ill-conditioned, $\kappa > {\sim}10^4$): Use ElasticNetCV. The evidence for this recommendation is among the strongest in the entire study. Figure~\ref{fig:precision_recall_curves_for_coef_retrieval} (right column, high dispersion): Lasso's recall drops to 0.48 at $\text{SNR} = 1.0$ and to 0.18 at $\text{SNR} = 0.04$. ElasticNet maintains recall of 0.83--0.93 across all SNR tiers at high~$\kappa$ (0.83 at SNR$=1.0$, 0.85 at SNR$=0.2$, 0.93 at SNR$=0.04$). This represents a ${\sim}1.7$--$5\times$ recall advantage for ElasticNet over Lasso in the high-$\kappa$ regime. Figure~C.1 (Lasso vs.\ ElasticNet F1) confirms: the strongest positive $\Delta$F1 values (0.50--0.75) are concentrated in the high-$\kappa$ columns at $n = 100$ and $n = 1$k. Figure~C.3 (Ridge vs.\ Lasso F1) shows Ridge massively outperforming Lasso at high $\kappa$ with small $n$, because Ridge's perfect recall ($1.0$) dominates when Lasso's recall collapses. Do not use Lasso at high $\kappa$ with small $n$. This is one of the most robust findings in the study.

SNR caveat at high $\kappa$: Figure~C.2 (Ridge vs.\ ElasticNet F1) reveals one exception to the unconditional ElasticNet recommendation. At $n \approx 100$, very low SNR ($\approx 0.04$, diagnosed by a saturated CV-elected $\alpha$; see Section~\ref{sec:diagnostics}), and high $\kappa$, Ridge achieves the \emph{highest} F1 score---including an advantage over ElasticNet---because Ridge's guaranteed recall of $1.0$ dominates when both Lasso and ElasticNet recall collapse under the combined pressure of very low SNR and high multicollinearity. This is not genuine variable selection (see ``When to consider Ridge'' below); if explicit sparsification is required, ElasticNet remains the least-bad option ($\gg$ Lasso). At moderate to high SNR (SNR $\geq 0.20$) or $n \geq 1{,}000$, ElasticNet is clearly preferred.

  \item If $\kappa$ is low (well-conditioned, $\kappa < {\sim}10^2$): The picture is more nuanced. Figure~\ref{fig:precision_recall_curves_for_coef_retrieval} (left column, low dispersion) shows: at $\text{SNR} = 1.0$, Lasso recall $= 0.82$ and ElasticNet recall $= 0.94$ (both reasonable); at $\text{SNR} = 0.2$, Lasso recall $= 0.45$ and ElasticNet recall $= 0.92$; at $\text{SNR} = 0.04$, Lasso recall $= 0.20$ and ElasticNet recall $= 0.91$. The pattern across all three SNR tiers is consistent: ElasticNet's recall is stable and high regardless of SNR ($\geq 0.91$ at low $\kappa$), while Lasso's recall is highly sensitive to SNR even at low $\kappa$. Default to ElasticNetCV even at low $\kappa$, unless the CV diagnostic (Section~\ref{sec:diagnostics}) confirms a high-SNR regime (low elected $\alpha$) and domain expertise suggests high sparsity, in which case Lasso becomes a reasonable alternative.
\end{itemize}

\paragraph{When to consider Ridge for variable selection.}
Figures~C.2 and C.3 reveal a finding that may surprise practitioners: Ridge frequently achieves the highest F1 scores at small~$n$, despite never performing explicit variable selection. This occurs because Ridge's guaranteed recall of 1.0 overwhelms the precision advantage of sparse methods when those methods' recall collapses---particularly at low SNR and high~$\kappa$. It is essential to understand that Ridge's high F1 in this regime arises from stable coefficient estimation with all features retained, not from genuine feature selection. A practitioner who requires an explicitly sparse model should not use Ridge, regardless of its F1 score.

If the practitioner suspects a low-SNR regime (diagnosed via high CV-elected~$\alpha$; see Section~\ref{sec:diagnostics}) and has small~$n$, a natural extension---not evaluated in this study---would combine Ridge's stable estimates with post-hoc feature importance methods (e.g., permutation importance) to achieve explicit variable selection. We note this as a direction for future work rather than an empirically validated recommendation.

\begin{table}[H]
\centering
\footnotesize
\caption{Recommended Regularization Method by Observable Regime: Variable Selection (F1)}
\label{tab:var_selection_guide}
\begin{tabular*}{\textwidth}{@{\extracolsep{\fill}}llp{4cm}@{}}
\toprule
\textbf{Observable Regime} & \textbf{Recommendation} & \textbf{Evidence} \\
\midrule
$n/p \geq 78$, any $\kappa$ & ElasticNet (safe default) or Lasso & Figs.\ C.1--C.3: near-zero $\Delta$F1 at large~$n$ \\
$n/p < 78$, high $\kappa$, SNR $\geq 0.20$ or $n \geq 1$k & \textbf{ElasticNet} & Fig.\ 4: ${\sim}1.7$--$5\times$ recall advantage; Figs.\ C.1, C.3 \\
$n \approx 100$, high $\kappa$, $\alpha$ saturated (SNR $\approx 0.04$) & Ridge$^\dagger$ (EN if genuine selection needed) & Fig.\ C.2: Ridge $>$ EN via recall $= 1.0$ at very low SNR \\
$n/p < 78$, low $\kappa$, high $\alpha$ & Ridge (competitive F1)$^\dagger$ & Figs.\ C.2, C.3: Ridge wins at small $n$, low SNR \\
$n/p < 78$, low $\kappa$, \textit{all of}: low elected $\alpha$ AND sparse domain & Lasso is viable & Fig.\ 4 (low $\kappa$, SNR~$= 1.0$): Lasso recall~$= 0.82$; EN recall~$= 0.94$ \\
$n/p < 78$, low $\kappa$, uncertain SNR & ElasticNet (safe default) & EN recall $\geq 0.91$ across all SNR at low $\kappa$ \\
\bottomrule
\end{tabular*}
\\[0.3em]
\footnotesize $^\dagger$\,Ridge achieves high F1 through recall $= 1.0$ (all features retained), \emph{not} genuine variable selection. Post-hoc filtering via permutation importance is a natural extension but was not evaluated in this study.
\end{table}

\subsubsection*{Path C: Priority is Coefficient Estimation (Minimizing Coefficient $L_2$ Error)}
\label{sec:path_c}

For coefficient estimation, the parameter hierarchy differs from both the F1 and RMSE analyses. Table~\ref{tab:l2_error_parameter_importance} reports $\omega^2$ effect sizes for pairwise $L_2$ error differences: the dominant main effect is the $\beta$ distribution (average $\omega^2 = 0.074$), which reflects the structural advantage Ridge holds over Lasso when the true coefficient vector is dense, and vice versa when it is sparse. Sample size $n$ ($\omega^2 = 0.008$) and condition number $\kappa$ ($\omega^2 = 0.003$) are secondary main effects---both below the negligible threshold ($< 0.01$) under standard conventions. We nevertheless branch on $\kappa$ here for a specific practical reason: it is the only observable parameter that \emph{reverses the direction} of the comparative advantage between methods. At high $\kappa$, ElasticNet dominates regardless of sparsity level; at low $\kappa$, the optimal choice depends on the latent sparsity of $\beta$. Sample size, by contrast, scales the magnitude of differences without changing their direction. The $\kappa \times \beta$ distribution interaction (average $F = 43$, $p < 0.001$; Table~\ref{tab:l2_error_parameter_importance}) formally confirms this reversal.

\paragraph{Branch 1---Condition number ($\kappa$).}
\begin{itemize}[nosep,leftmargin=*]
  \item If $\kappa$ is \textbf{high} ($\kappa > {\sim}10^4$): Use ElasticNetCV. Figure~D.3 (Lasso vs.\ ElasticNet) shows ElasticNet achieving 20--40\% lower $L_2$ error than Lasso at high $\kappa$ with $n = 100$ and $n = 1$k. Figure~D.4 (Ridge vs.\ ElasticNet) confirms ElasticNet holds a consistent advantage over Ridge at high $\kappa$ regardless of sparsity: the advantage is largest for sparse models (20--40\% lower error; Figs.~D.3, D.4) and modest but consistent for dense models---Figure~D.4 shows light-brown cells (EN better) at high $\kappa$ with no teal cells indicating Ridge ever clearly wins. Figure~D.6 confirms Ridge outperforms Lasso at high $\kappa$, reinforcing that Lasso should be avoided in this regime. Use ElasticNet at high~$\kappa$; Ridge is an acceptable fallback given its simpler implementation, but ElasticNet is the more defensible default.
  
  \item If $\kappa$ is low ($\kappa < {\sim}10^2$): The tradeoff between Ridge and Lasso reverses. Figure~D.6 shows that at low $\kappa$ with Sparsity $= 15\%$, Lasso generally outperforms Ridge, correctly identifying and removing zero-coefficient features. At Sparsity $= 0\%$, Ridge holds the advantage. Figure~D.3 at low $\kappa$ shows Lasso and ElasticNet performing comparably. Practical rule at low $\kappa$: If domain expertise suggests sparsity, Lasso is a reasonable choice. If the model is expected to be dense, Ridge is preferred. ElasticNet is a safe middle ground.
\end{itemize}

\paragraph{Branch 2---Sample-to-feature ratio ($n/p$).}
\begin{itemize}[nosep,leftmargin=*]
  \item If $n/p \geq 78$: Coefficient estimation differences narrow across all methods and all $\kappa$ regimes. Figures~D.1--D.6 consistently show near-zero relative differences in the $n = 10$k and $n = 100$k rows. At sufficient sample size, all methods converge toward the true coefficients. Any method is acceptable; differences are negligible.
  \item If $n/p < 78$: The differences documented above become consequential. The combination of small $n$ and high $\kappa$ is the worst-case scenario for coefficient estimation and is where method choice matters most.
\end{itemize}

\paragraph{Methods to avoid for coefficient estimation.}
\begin{itemize}[nosep,leftmargin=*]
  \item Post-Lasso OLS. Figure~D.2 shows predominantly higher coefficient $L_2$ error than standard Lasso across virtually the entire heatmap. This gap intensifies at decreasing SNR and high~$\kappa$. When Lasso's first-stage selection is imperfect, the unpenalized OLS refit amplifies errors by inflating coefficients for incorrectly selected noise variables while failing to recover contributions from omitted signal variables. We note that this finding applies to our specific implementation (LassoCV for first-stage selection using the same training data for the OLS refit); alternative two-stage procedures such as Double Lasso \parencite{fitzgerald_sice_practical_insights_double_lasso_2023} may mitigate some of these issues but were not evaluated in this study.
  \item Additionally, as documented in Section~2.7, standard inference techniques fail on Lasso-selected variables \parencite{berk2013postselection, lee2016postselection}. $P$-values and confidence intervals derived from OLS regressions on Lasso-selected subsets are invalid due to selection bias.
\end{itemize}

\begin{table}[H]
\centering
\footnotesize
\caption{Recommended Regularization Method by Observable Regime: Coefficient Estimation ($L_2$ Error)}
\label{tab:coef_estimation_guide}
\begin{tabular*}{\textwidth}{@{\extracolsep{\fill}}llp{7cm}@{}}
\toprule
\textbf{Observable Regime} & \textbf{Recommendation} & \textbf{Evidence} \\
\midrule
$n/p \geq 78$, any $\kappa$ & Any method (diff.\ negligible) & Figs.\ D.1--D.6: near-zero cells at large~$n$ \\
$n/p < 78$, high $\kappa$ & \textbf{ElasticNet} & Figs.\ D.3, D.4: 20--40\% lower error (sparse); consistent edge over Ridge (dense) \\
$n/p < 78$, low $\kappa$, sparse & Lasso or ElasticNet & Fig.\ D.6: Lasso beats Ridge with sparsity \\
$n/p < 78$, low $\kappa$, dense & \textbf{Ridge} & Fig.\ D.6: Ridge beats Lasso without sparsity \\
Any $n$, any $\kappa$ & Avoid Post-Lasso OLS$^\ddagger$ & Fig.\ D.2: higher error across nearly all cells \\
\bottomrule
\end{tabular*}
\\[0.3em]
\footnotesize $^\ddagger$\,Applies to the specific implementation tested (LassoCV first-stage, same-data OLS refit). Alternative two-stage procedures were not evaluated.
\end{table}

\subsection{Data-Driven Diagnostics}
\label{sec:diagnostics}

The guide above branches on two observable quantities ($n/p$ and $\kappa$) and one domain-knowledge prior (sparsity). Two latent parameters---SNR and true sparsity---remain important for fine-grained decisions. We offer practical strategies for approximating each.

\paragraph{Estimating SNR via the CV-elected $\alpha$.}
The regularization strength $\alpha$ selected by LassoCV serves as the most accessible diagnostic proxy for the underlying signal-to-noise regime:

\begin{itemize}[nosep,leftmargin=*]
  \item Scikit-learn's LassoCV explores $\alpha$ values up to a maximum defined as $\alpha_{\max} = \frac{1}{n}\|X^{T}y\|_{\infty}$ (Equation~10 in Section~5). This upper bound is the smallest $\alpha$ at which all coefficients are driven to exactly zero.
  \item If the CV procedure selects an $\alpha$ that is extremely high or saturates at the maximum boundary of the search grid, the data resides in an under-determined or low-SNR regime. Figure~B.1 provides direct confirmation: the darkest cells ($\alpha$ approaching $10^4$--$10^5$) are concentrated exclusively in small-$n$, low-SNR rows.
  \item If the CV procedure selects a moderate or low $\alpha$ (e.g., $\alpha < 1.0$ for standardized features), this indicates sufficient signal for the $\ell_1$ penalty to operate discriminatively.
\end{itemize}

\begin{table}[H]
\centering
\footnotesize
\caption{Actionable Mapping: CV-Elected $\alpha$ as SNR Diagnostic. Threshold ranges are approximate and derived from the empirical distribution of elected $\alpha$ across our simulation grid (Figure~B.1), conditional on known SNR tiers (0.04, 0.2, 1.0). Practitioners should treat these as heuristic guidelines rather than sharp cutoffs; the appropriate thresholds will vary with feature scaling and dataset characteristics.}
\label{tab:alpha_diagnostic}
\begin{tabular*}{\textwidth}{@{\extracolsep{\fill}}llp{7cm}@{}}
\toprule
\textbf{CV-Elected $\alpha$} & \textbf{Likely Regime} & \textbf{Recommended Action} \\
\midrule
High / at boundary & Low SNR
    & Switch to \textbf{Ridge} for all objectives.\newline
      Lasso recall collapses to 0.18--0.20 in this regime
      (Fig.~\ref{fig:precision_recall_curves_for_coef_retrieval});\newline
      Ridge outperforms Lasso on F1 (Fig.~C.3) and
      RMSE (Fig.~E.4, low-SNR rows). \\
Moderate (e.g., 0.1--10) & Moderate SNR
    & \textbf{ElasticNet} provides the best risk-adjusted
      choice across all objectives. \\
Low (e.g., $< 0.1$) & High SNR
    & \textbf{Lasso} becomes viable if $\kappa$ is also low
      and sparsity is expected.\newline
      Ridge and ElasticNet remain safe defaults. \\
\bottomrule
\end{tabular*}
\end{table}

\noindent This diagnostic is computationally free---LassoCV reports \texttt{alpha\_} after fitting---and can be run as a preliminary screening step before committing to a final model.

\paragraph{Estimating sparsity via domain expertise.}
True sparsity is never directly observed, but domain knowledge provides reliable priors. Our simulations tested two sparsity levels (0\% and 15\%). While sparsity itself had a negligible direct effect on RMSE differences ($\omega^2 < 0.001$ across all comparisons; Table~\ref{tab:method_superiority_omega2}) and a small effect on F1 differences ($\omega^2 = 0.003$ average; Table~\ref{tab:f1_parameter_importance}), it interacts meaningfully with $\kappa$ for coefficient estimation at low~$\kappa$ (Figure~D.6):

\begin{itemize}[nosep,leftmargin=*]
  \item Assume a dense model (favoring Ridge or ElasticNet) when working with engineered feature sets typical of industry ML: churn prediction, conversion modeling, demand forecasting, recommendation systems. Feature engineering deliberately constructs predictors believed to carry signal, and the fraction of truly zero coefficients is likely small.
  \item Assume a sparse model (favoring Lasso or ElasticNet with high L1 ratio) when working in genomics, text classification (bag-of-words), sensor arrays, or high-dimensional screening where $p$ is large by construction and the vast majority of features are expected to be irrelevant---provided the CV diagnostic above does not signal a low-SNR regime. If it does, ElasticNet is the safer choice even in sparse domains, as it preserves the grouping effect and maintains high recall (Figure~\ref{fig:precision_recall_curves_for_coef_retrieval}, Figure~C.1). Note that these high-dimensional applications typically involve $p$ values far exceeding the range tested in our simulations ($p \in \{64, 128\}$) and sparsity levels well above 15\%; extrapolation to such settings requires additional validation.
\end{itemize}

\begin{figure}[!htbp]
\centering
\includegraphics[width=\textwidth]{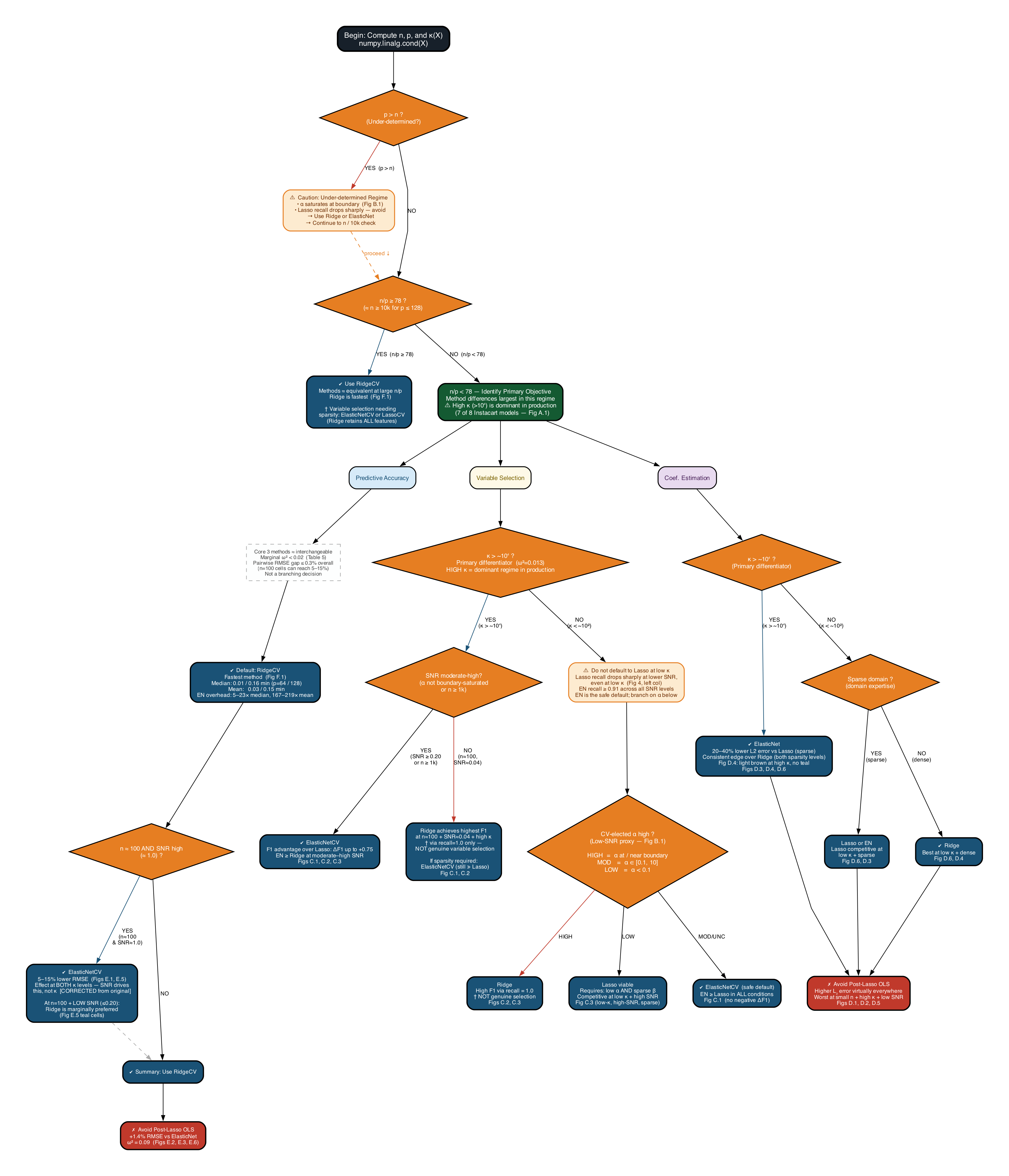}
\caption{Consolidated decision framework for regularization selection. Decision rules branch on observable quantities: the ratio $p/n$ (under-determined check), the sample-to-feature ratio $n/p$ (large-sample threshold $\geq 78$, empirically determined from Figure~5), condition number ($\kappa$), and CV-elected $\alpha$. Recommendations are validated within $p \in \{64, 128\}$, $n \in \{100\text{--}100\text{k}\}$, $\kappa \in \{{\sim}10^1, {\sim}10^5\text{--}10^6\}$, sparsity $\in \{0\%, 15\%\}$. Thresholds for $\kappa$ (${\sim}10^2$ and ${\sim}10^4$) are interpolated from two tested regimes; the intermediate range was not directly evaluated. Ridge achieves high F1 in certain regimes through recall $= 1.0$ (all features retained), not genuine variable selection ($\dagger$). The $\beta$ distribution ($\omega^2 = 0.112$ for EN--R; Table~\ref{tab:f1_parameter_importance}), $n \times \text{SNR}$ interaction ($F = 569$; Table~\ref{tab:f1_parameter_importance}), and Sparsity $\times$ $\kappa$ interaction (Figures~D.4, D.6; not formally quantified) are additional sources of variation not captured in the branching structure. ElasticNet compute overhead (167--219$\times$ by mean; 5--23$\times$ by median) is specific to the 8-value L1 grid used in this study and scales linearly with grid size. Ridge + post-hoc filtering is noted as future work, not an empirical finding of this study.}
\label{fig:decision_flowchart}
\end{figure}

\paragraph{Summary decision rule.}
Compute $\kappa$ via \texttt{numpy.linalg.cond(X)} and the ratio $n/p$. If $p > n$, default to Ridge or ElasticNet (Lasso recall collapses in under-determined regimes). If $n/p \geq 78$, use Ridge for prediction and coefficient estimation, or ElasticNet/Lasso if explicit variable selection is needed. If $n/p < 78$, branch on~$\kappa$: at high~$\kappa$, use ElasticNet; at low~$\kappa$, run a quick LassoCV and inspect the elected~$\alpha$. If $\alpha$ is high, use Ridge. If $\alpha$ is low and sparsity is expected, Lasso is viable. In all uncertain cases, ElasticNet is the safest general-purpose default, and Post-Lasso OLS should be avoided. Figure~\ref{fig:decision_flowchart} consolidates these rules into a single decision flowchart.

\FloatBarrier

%% file: section_7_conclusion.tex
\section{Concluding Remarks}

In this analysis we empirically examined the strengths and weaknesses of the canonical scikit-learn regularization frameworks of Lasso, Ridge, ElasticNet, and Post-Lasso OLS across 134,400 simulations spanning 960 configurations of a 7-dimensional manifold (\textit{number of features ($p$)}, \textit{rank ratio}, \textit{eigenvalue dispersion ($\kappa$)}, \textit{$\beta$ distribution}, \textit{sparsity}, \textit{signal-to-noise ratio (SNR)}, and \textit{sample size ($n$)}). While our analytic framework carries limitations---a relatively narrow range of $p$ and $n$, only two sparsity levels, and $\kappa$ tested only at two extremes---we have the advantage of grounding our hyperparameter ranges in eight real-world productionalized ML models (Appendix~A).

Several results are sharper than the regularization literature has previously characterized for applied ML settings. First, {Ridge, Lasso, and ElasticNet are nearly interchangeable for prediction: median test-set RMSE differs by at most 0.3\% across the three methods and no hyperparameter achieves even a small effect size ($\omega^2 \geq 0.01$) for pairwise RMSE differences among them (Table~\ref{tab:method_superiority_omega2}: all values $< 0.02$). This near-equivalence is itself the primary finding for practitioners whose sole objective is prediction accuracy---the choice of regularizer matters far less than sample size. Second, Lasso's recall is fragile under multicollinearity: at high $\kappa$ and low SNR, Lasso recall collapses to 0.18 while ElasticNet maintains 0.93, a ${\sim}5\times$ advantage (Figure~\ref{fig:precision_recall_curves_for_coef_retrieval}). Do not use Lasso at high $\kappa$ with small $n$; this is the most robust finding of the study. Third, method differences have a hard ceiling: at $n/p \geq 78$ (equivalently, $n \geq 10{,}000$ for $p \leq 128$ as tested), performance gaps among all methods vanish across prediction, variable selection, and coefficient estimation simultaneously (Figures~C.1--C.3, D.1--D.6, E.1--E.6). Below this threshold---particularly at $n < 1{,}000$---method choice is consequential.

For practitioners navigating these tradeoffs, Section~\ref{sec:objective_driven_guide} distills the simulation results into an objective-driven decision guide (Figure~\ref{fig:decision_flowchart}) that branches on three observable quantities: the sample-to-feature ratio ($n/p$), condition number ($\kappa$), and the CV-elected $\alpha$ as a proxy for the latent SNR. The guide separates recommendations by objective: prediction favours Ridge; variable selection at high $\kappa$ favours ElasticNet; coefficient estimation at high $\kappa$ also favours ElasticNet regardless of sparsity; and Post-Lasso OLS should be avoided across all regimes.

Four directions for future work follow directly from this study's findings and limitations.

\smallskip

\begin{enumerate}[nosep, leftmargin=*]
  \item Formalizing the $\alpha$--SNR proxy. Fitting a calibration model on the existing simulations to map $\alpha_{\text{elected}} / \alpha_{\max}$ to known SNR tiers would convert Table~\ref{tab:alpha_diagnostic}'s heuristic thresholds into a quantitative diagnostic.
  \item Filling the intermediate-$\kappa$ gap. The decision thresholds in Figure~\ref{fig:decision_flowchart} are interpolated from two tested extremes; adding 2--3 intermediate $\kappa$ levels would determine whether transitions are smooth or exhibit phase-change behaviour.
  \item Ridge + post-hoc variable selection. Combining RidgeCV estimates with permutation importance would provide an empirically validated alternative to ElasticNet in the low-SNR, small-$n$ regime where Ridge achieves high F1 through recall $= 1.0$ rather than genuine sparsification.
  \item Extending the benchmark to newer methods. This study evaluates methods formalized by 2013 (Ridge, 1970; Lasso, 1996; ElasticNet, 2005; Post-Lasso OLS, 2013), while the literature review covers a substantially richer landscape through 2026. Priority candidates for the same simulation harness include the Adaptive Lasso \parencite{zou2006adaptive}, MCP and SCAD \parencite{fan2001nonconcave, zhang2010mcp}---which may reduce the recall collapse documented here---the Knockoff Filter \parencite{candes2018modelx}, and gradient-descent-based implicit regularization \parencite{tang2025benign, luo2026penalization}. The simulation framework developed here is directly reusable as a harness for any such comparison.
    \item Focused assessment of regularizers with closed-form solutions. As $n$ increases, the predictive performance gaps among regularizers diminish, leaving computational efficiency as a primary deciding factor. Ridge regression is particularly efficient due to its closed-form analytic solution. The LARS-Lasso algorithm also provides analytic solutions for the algorithmic steps, making it a potential competitor to Ridge in terms of its utility in an applied ML setting.    
\end{enumerate}
\smallskip
Our hope is that the practitioner's roadmap laid out above will help better equip Machine Learning Engineers, Data Scientists, and Analysts to make principled, evidence-based regularization decisions---and that the simulation framework itself will serve as a reusable benchmark for the richer landscape of methods that lies ahead.

\FloatBarrier

%% file: appendix.tex
\begin{appendices}

\renewcommand{\thefigure}{\Alph{section}.\arabic{figure}}
\makeatletter
\@addtoreset{figure}{section}
\makeatother

\section{Sample Model Feature Distributions}

\begin{figure}[!htbp]
\centering
\includegraphics[width=1.0\textwidth]{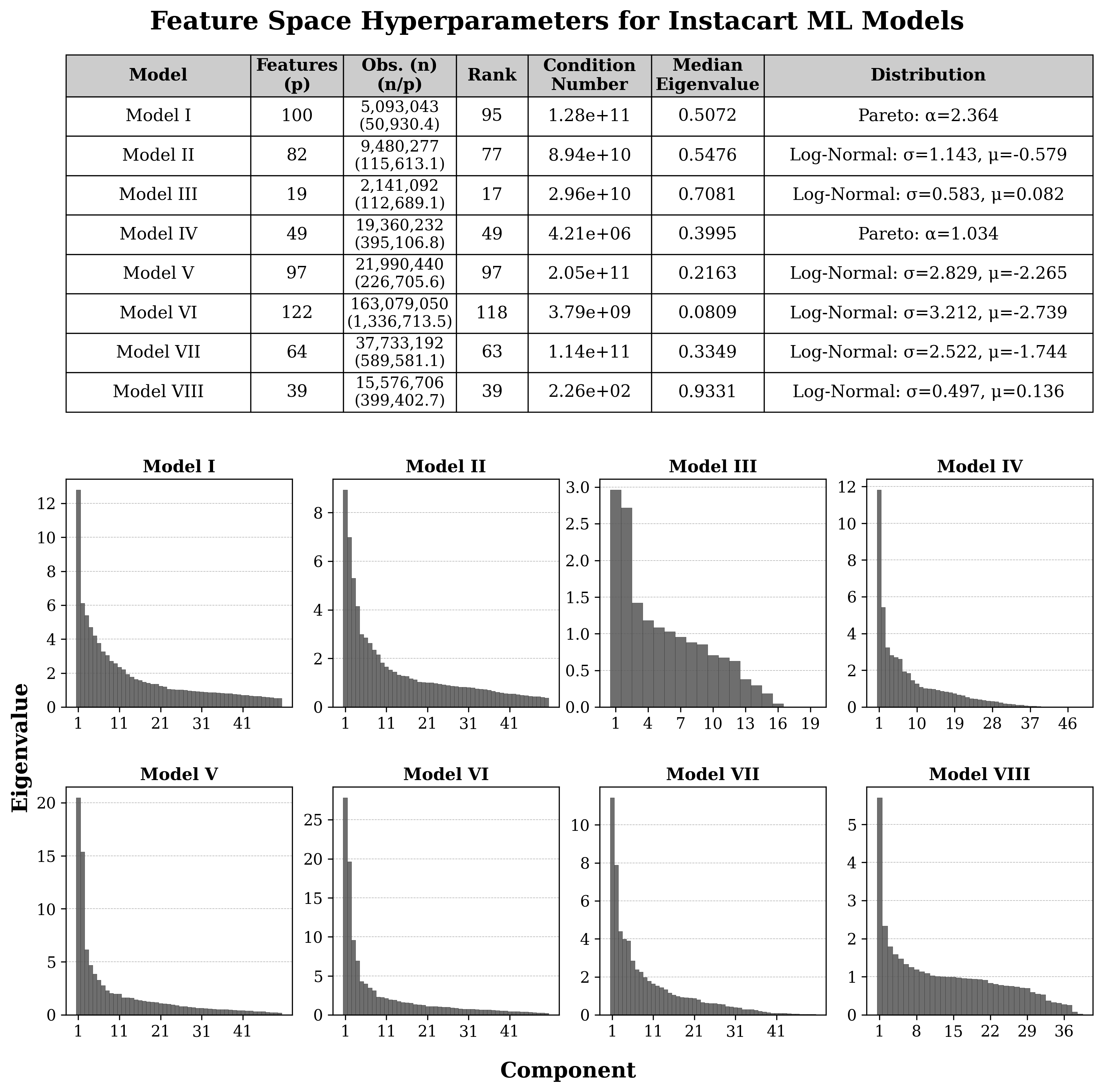}
\caption{Distributions of eigenvalues across 8 ML models sampled from production.}
\label{fig:hyperparameter_sampling}
\end{figure}

\clearpage 

\section[Elected alpha Values]{Elected $\alpha$ Values}

\begin{figure}[!htbp]
\centering
\includegraphics[width=0.86\textwidth]{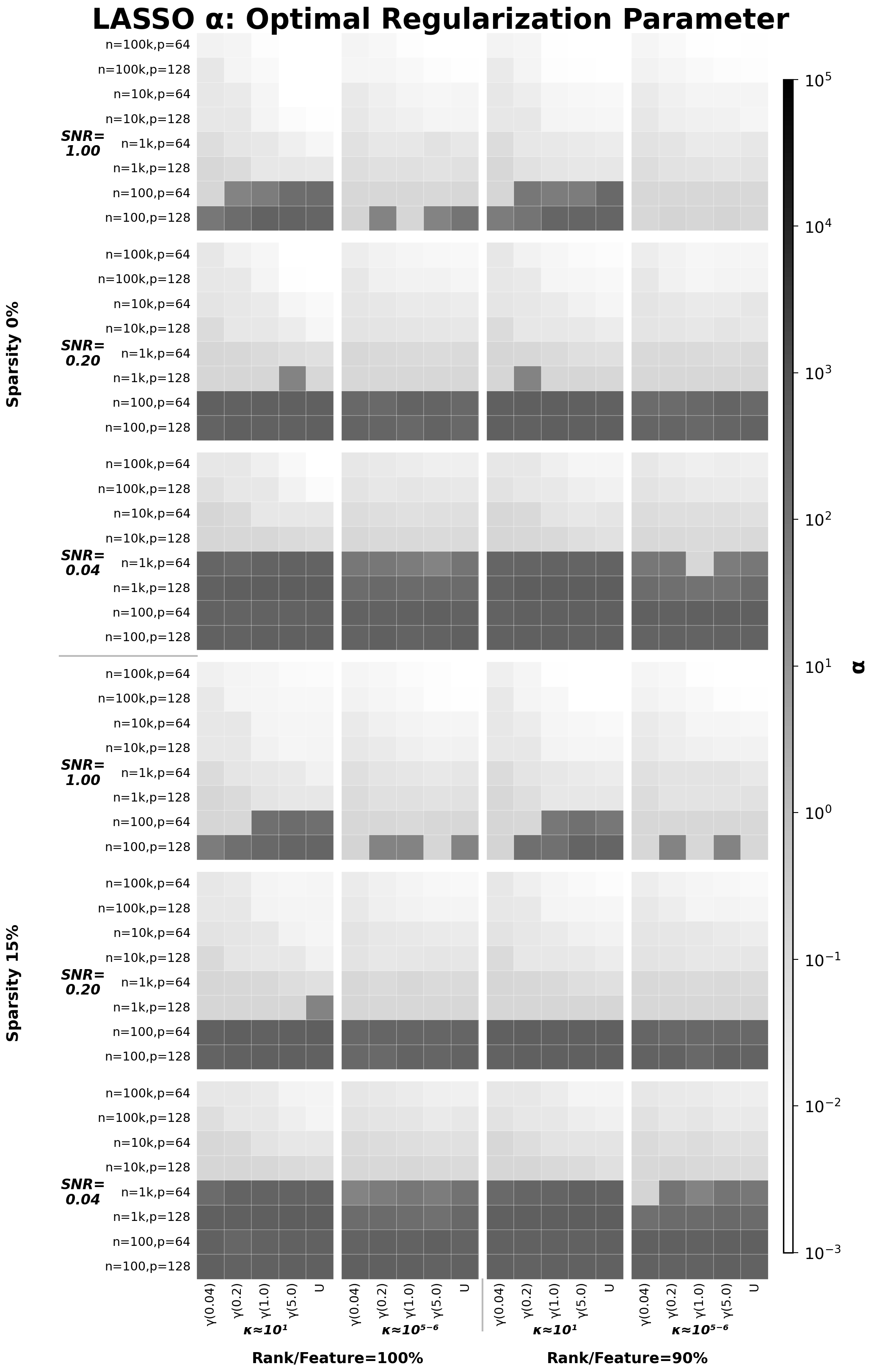}
\small \caption{The optimal Lasso regularization parameter $\alpha$ is primarily driven by SNR and sample size $n$; lower SNR and smaller datasets require significantly higher $\alpha$ values to prevent overfitting.}
\label{fig:alpha_opt_heatmap}
\end{figure}

\clearpage

\section{F1 Score Performance / Coefficient Recovery}

\begin{figure}[!htbp]
\centering
\small \includegraphics[width=0.86\textwidth]{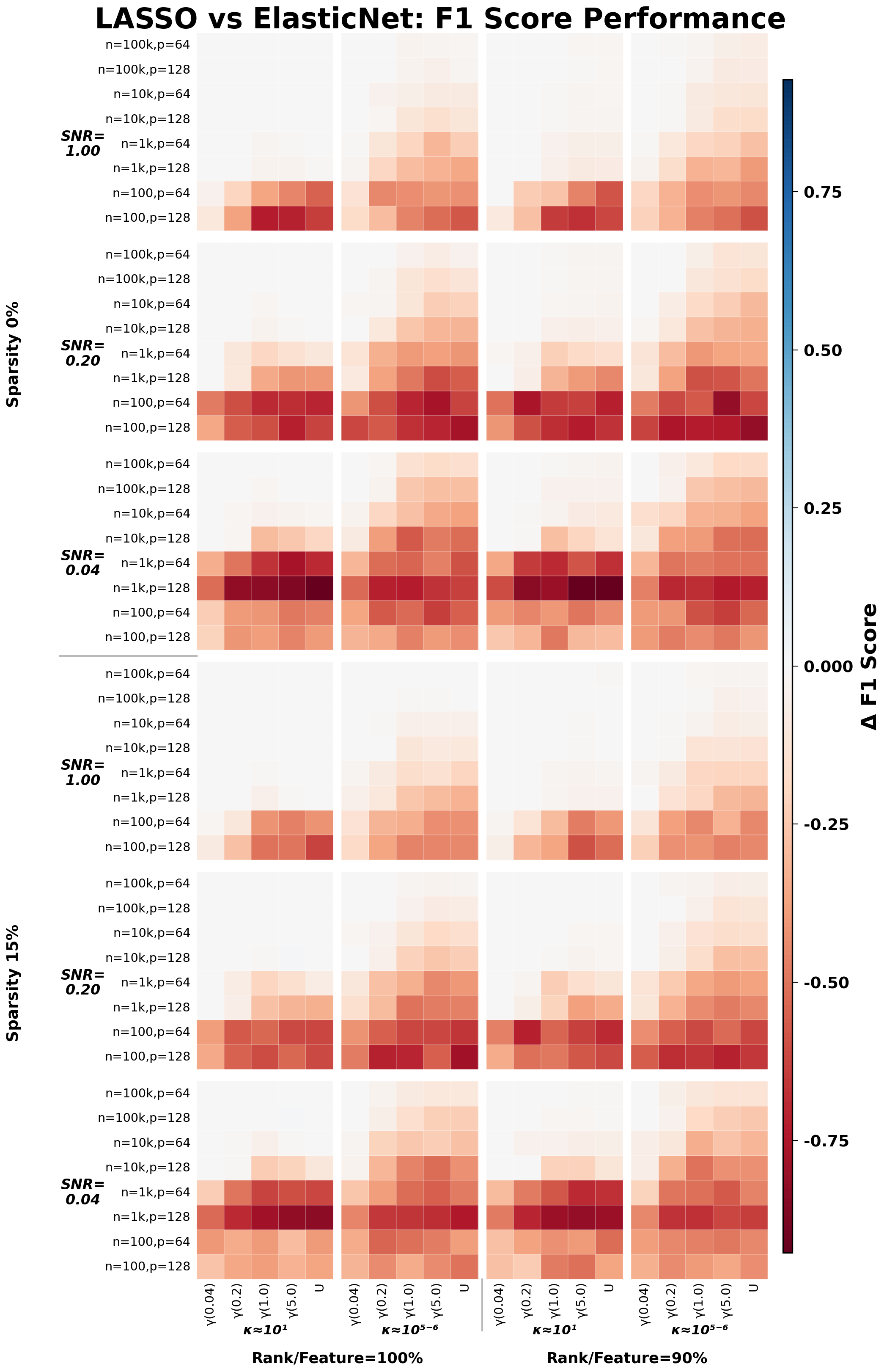}
\caption{ElasticNet consistently outperforms Lasso, especially as SNR decreases. This performance gap is most pronounced in small-$n$, large-$p$ scenarios.}
\end{figure}

\begin{figure}[!htbp]
\centering
\includegraphics[width=0.86\textwidth]{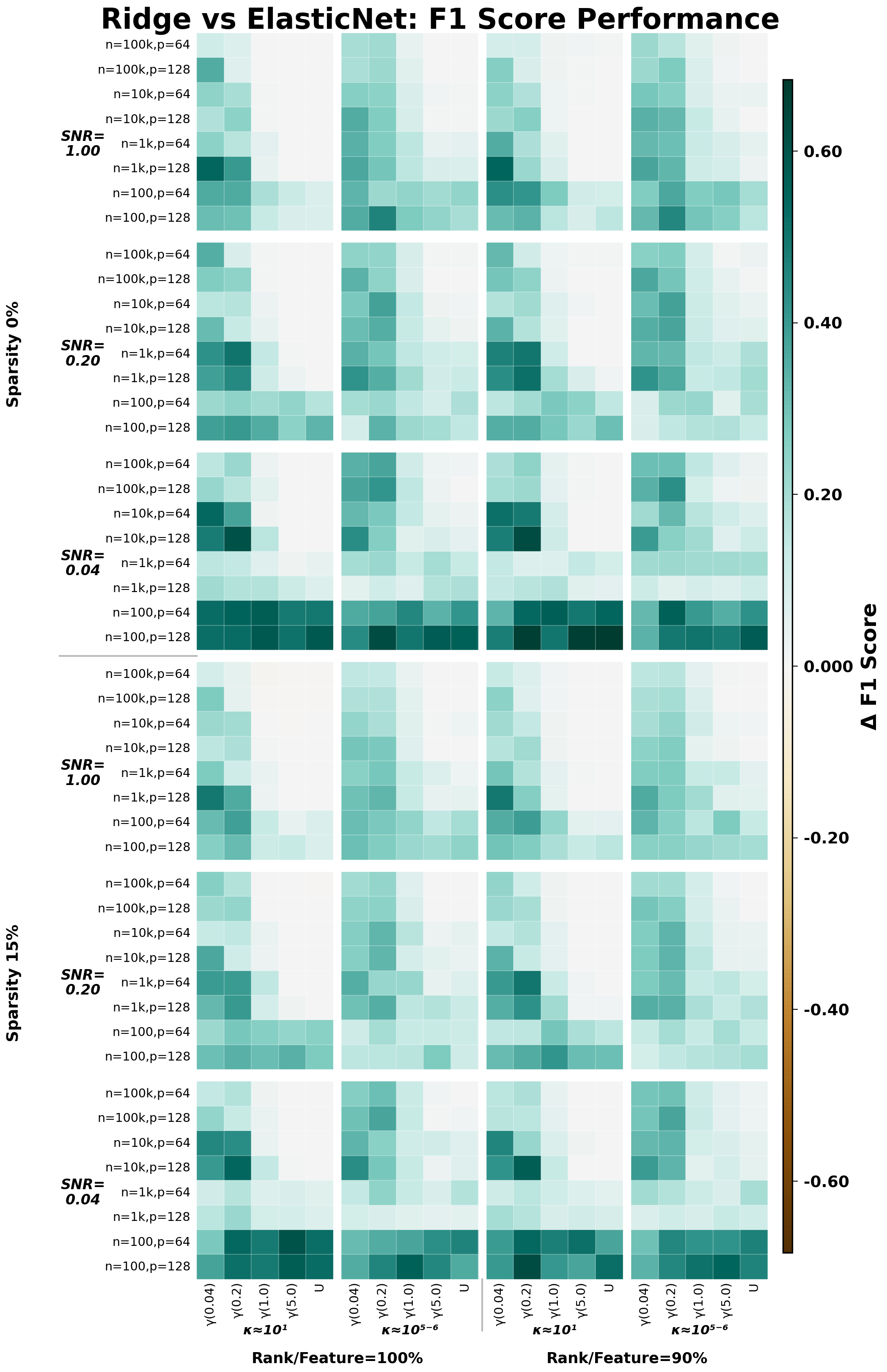}
\caption{Ridge demonstrates superior F1 performance over ElasticNet across the majority of simulation parameters, with the advantage becoming most prominent in low sample size ($n=100$) and low SNR ($SNR=0.04$) environments.}
\end{figure}

\begin{figure}[!htbp]
\centering
\includegraphics[width=0.86\textwidth]{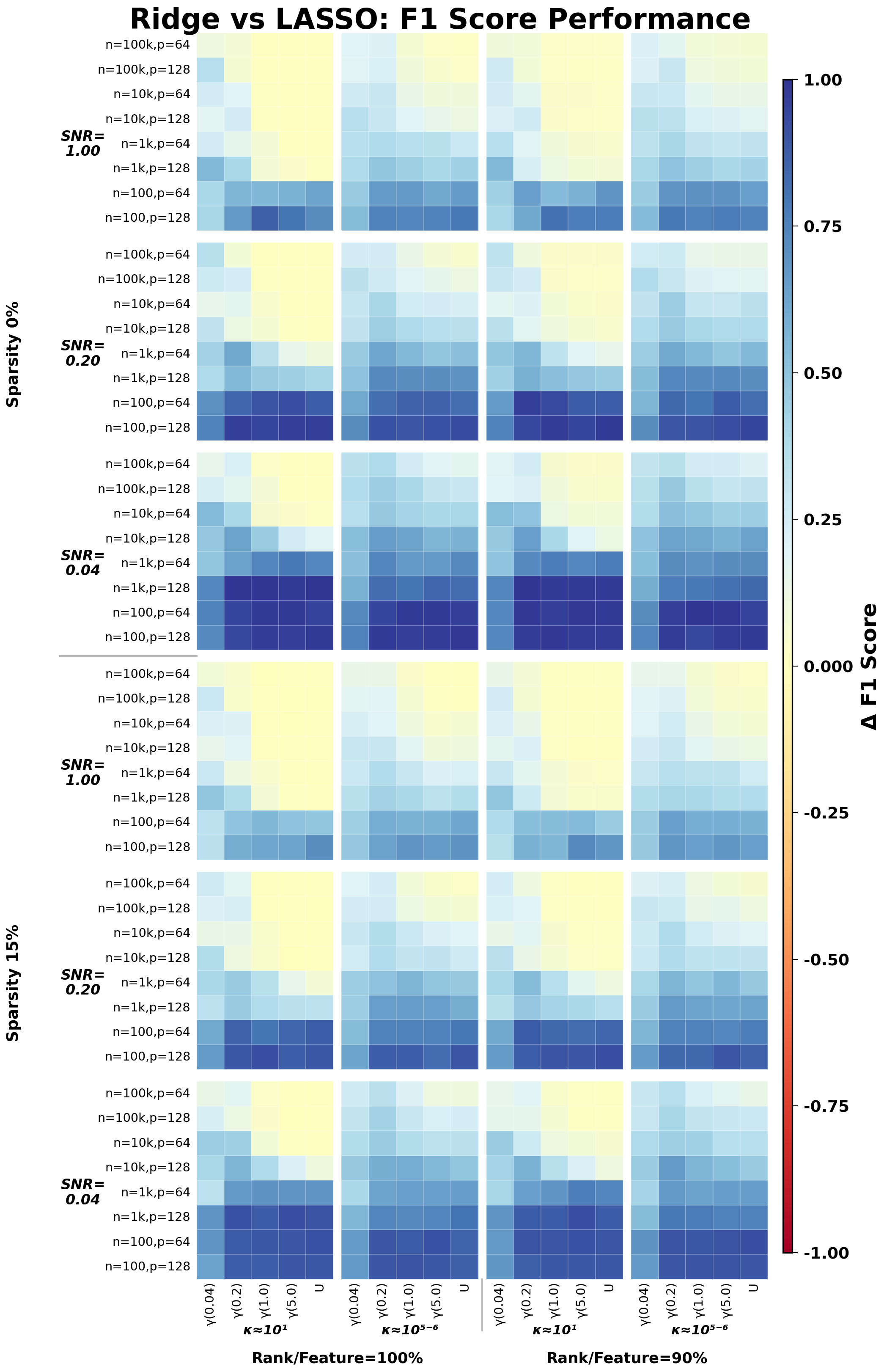}
\caption{Ridge consistently outperforms or matches Lasso across most tested conditions, with the strongest performance gains occurring at lower SNR and smaller sample sizes (e.g., $n=100$).}
\end{figure}

\clearpage

\section{L2 Coefficient Error}

\begin{figure}[!htbp]
\centering
\includegraphics[width=0.86\textwidth]{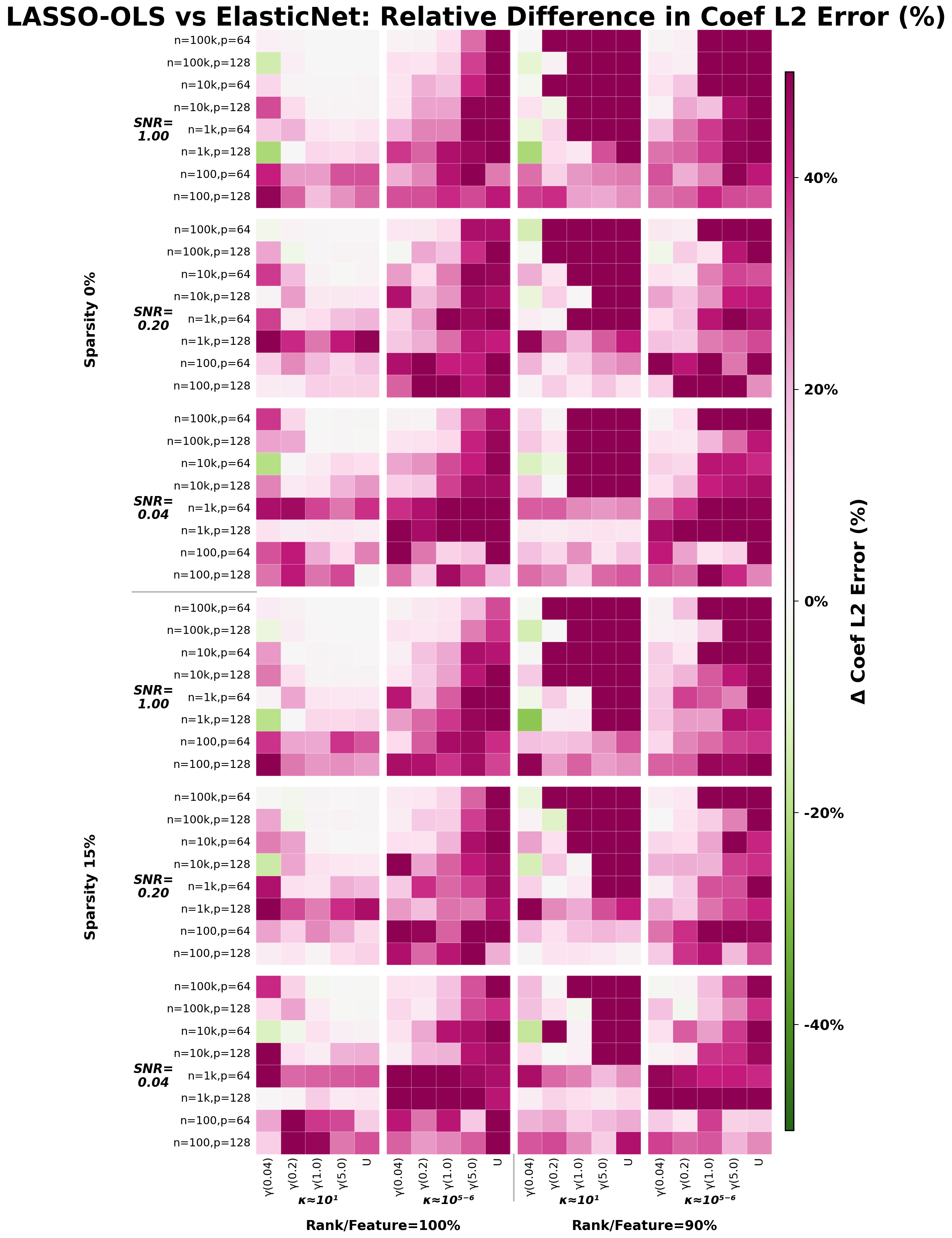}
\caption{ElasticNet consistently outperforms Post-Lasso OLS in coefficient accuracy across most simulated conditions, with its greatest performance advantages appearing in scenarios with high multicollinearity (low condition number $\kappa$) and lower SNR.}
\end{figure}

\begin{figure}[!htbp]
\centering
\includegraphics[width=0.86\textwidth]{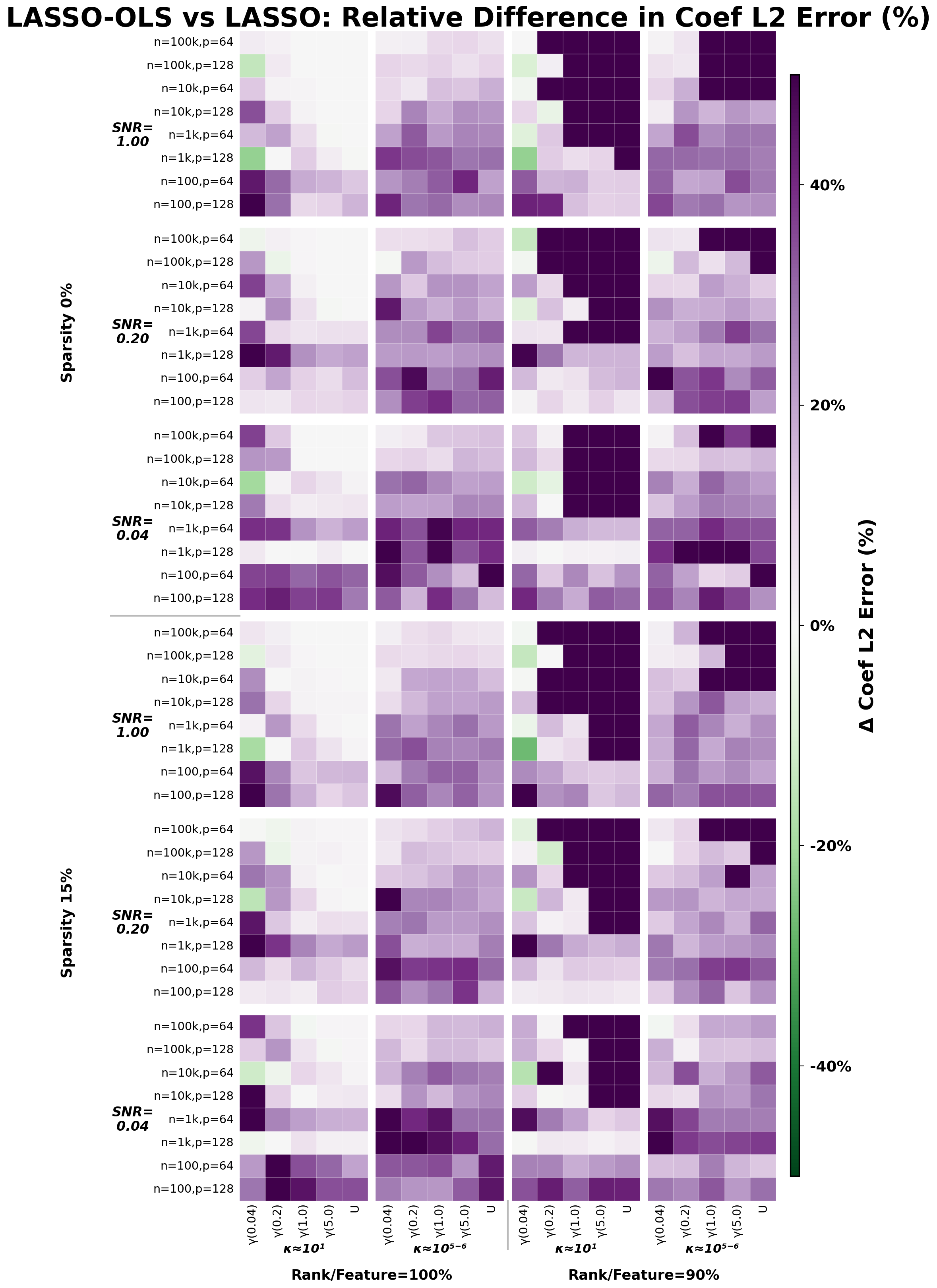}
\caption{Post-Lasso OLS consistently yields a higher coefficient $L_2$ error than standard Lasso, with this performance gap intensifying as SNR decreases and the condition number $\kappa$ increases.}
\end{figure}

\begin{figure}[!htbp]
\centering
\includegraphics[width=0.86\textwidth]{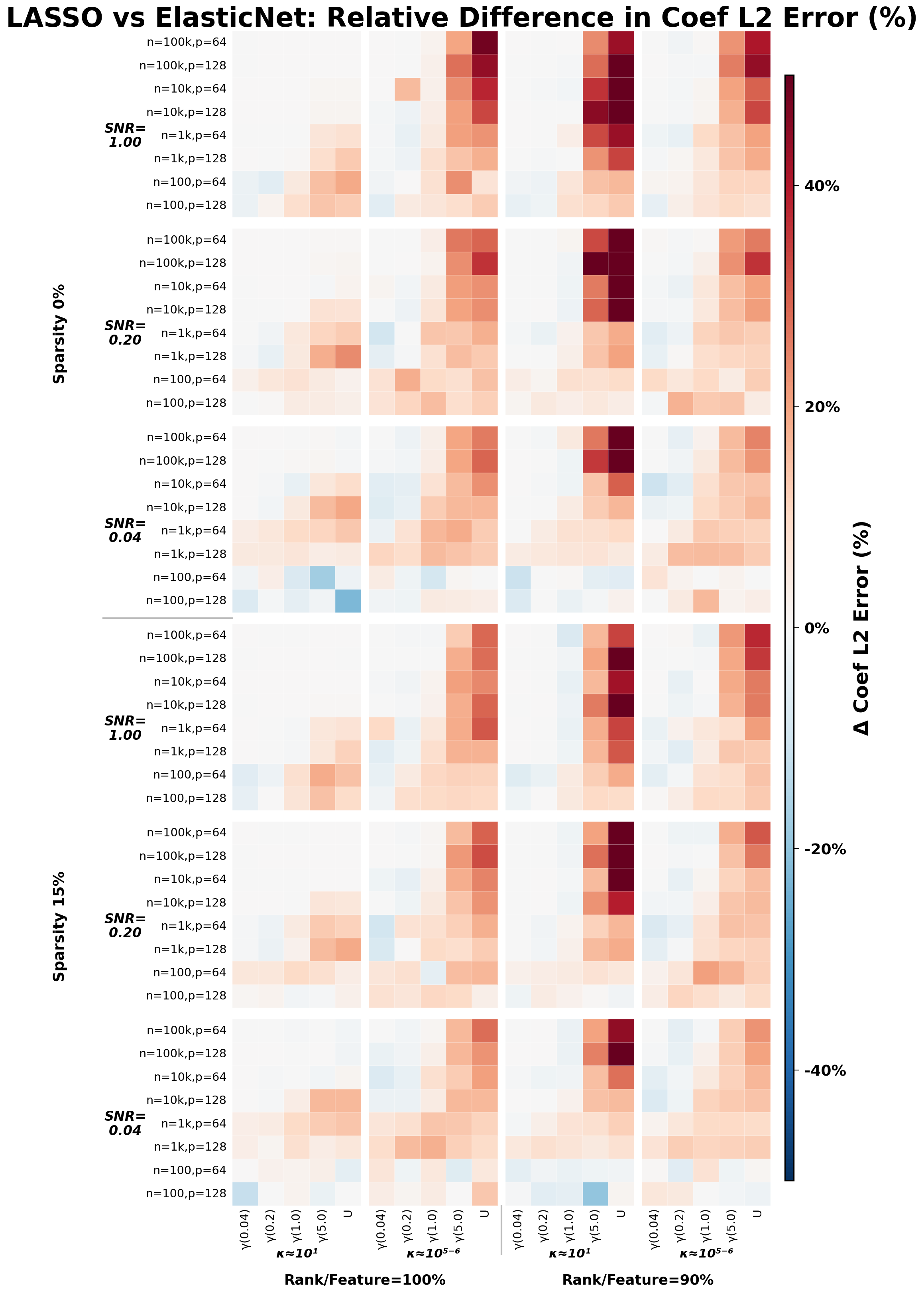}
\caption{ElasticNet generally achieves a lower coefficient $L_{2}$ error compared to Lasso as the feature correlation ($\gamma$) and condition number ($\kappa$) increase.}
\end{figure}

\begin{figure}[!htbp]
\centering
\includegraphics[width=0.86\textwidth]{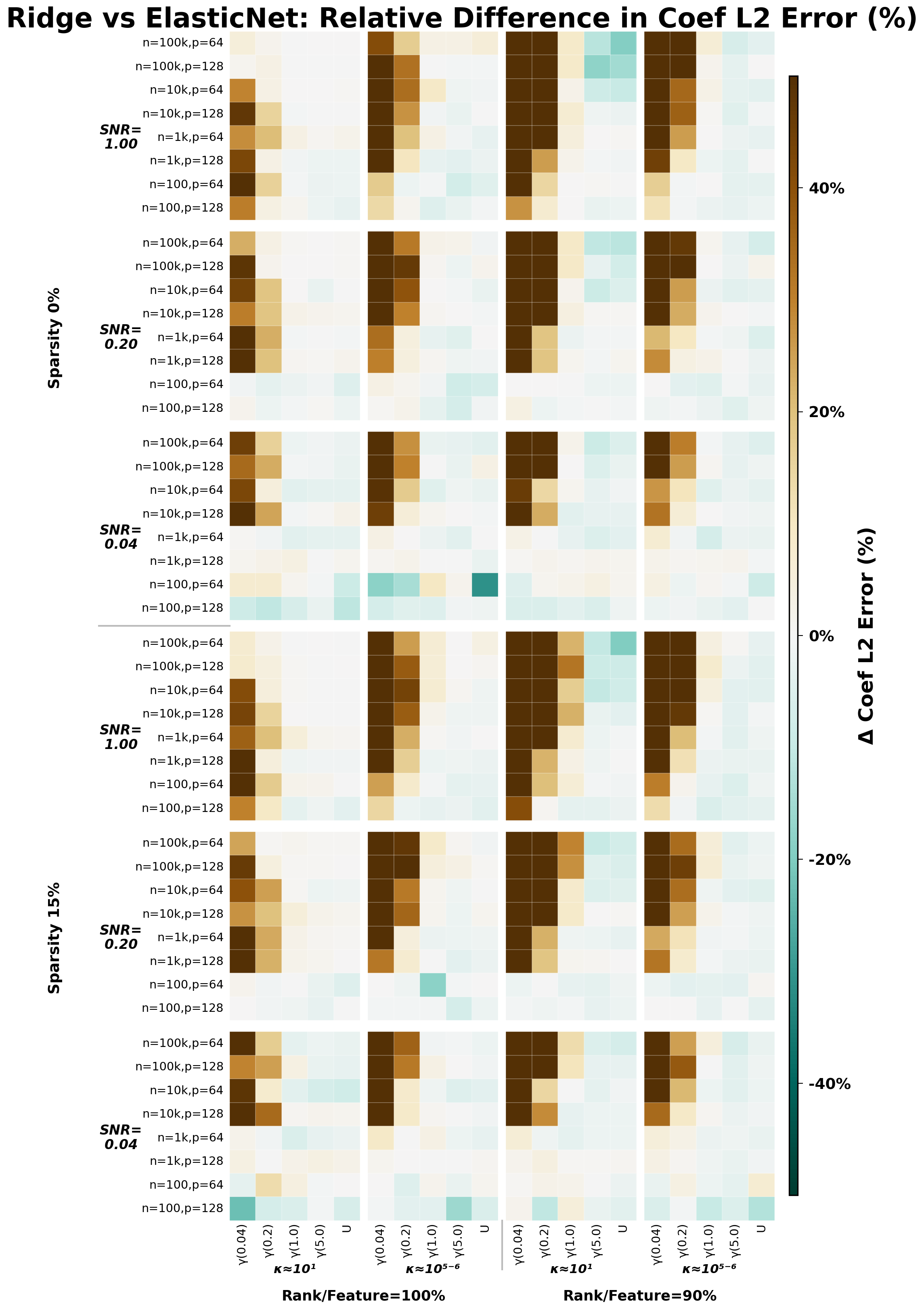}
\caption{ElasticNet significantly outperforms Ridge regression as $\gamma$ increases and SNR decreases, with the advantage being most pronounced in sparse (Sparsity 15\%) and high-dimensional ($\kappa \approx 10^{-6}$) settings.}
\end{figure}

\begin{figure}[!htbp]
\centering
\includegraphics[width=0.86\textwidth]{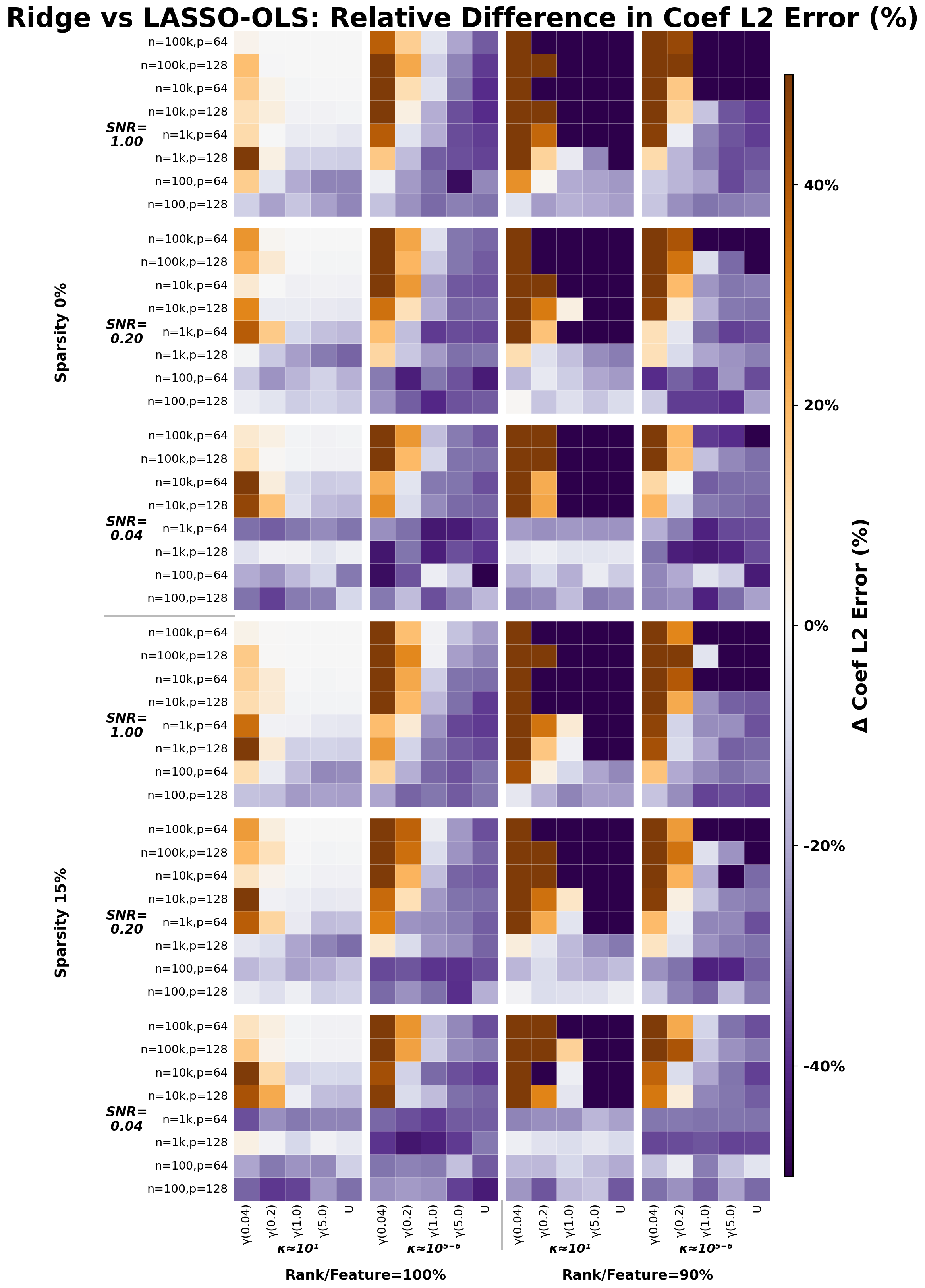}
\caption{Ridge consistently achieves lower coefficient $L_2$ error compared to Post-Lasso OLS as $\gamma$ increases, a trend that remains robust across varying sparsity levels and SNR.}
\end{figure}

\begin{figure}[!htbp]
\centering
\includegraphics[width=0.86\textwidth]{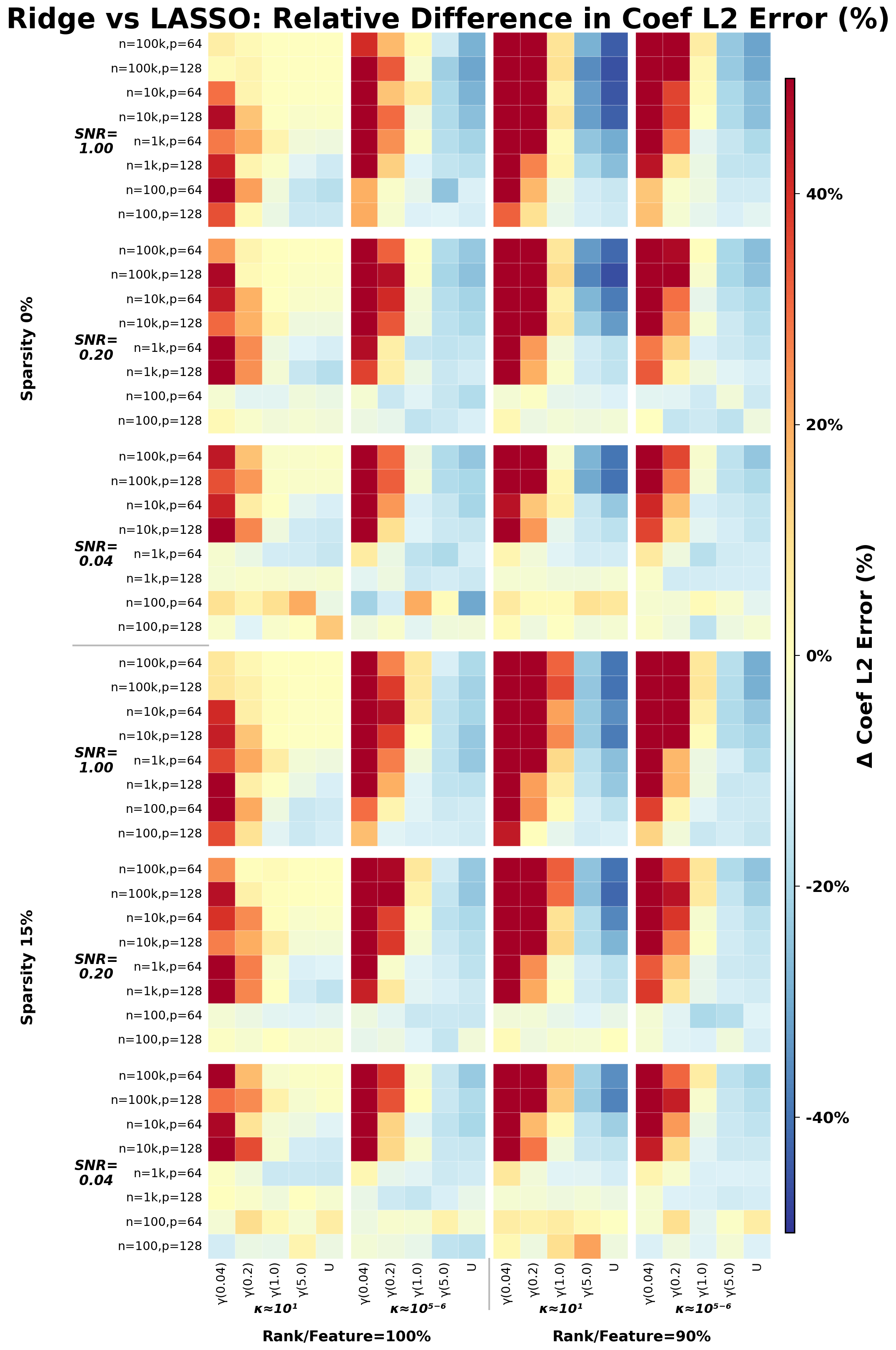}
\caption{While Lasso generally outperforms Ridge in high-sparsity scenarios, Ridge tends to achieve lower coefficient $L_2$ error as $\kappa$ increases and the feature rank-to-dimension ratio decreases.}
\end{figure}

\clearpage

\section{Root Mean Square Error}

\begin{figure}[!htbp]
\centering
\includegraphics[width=0.86\textwidth]{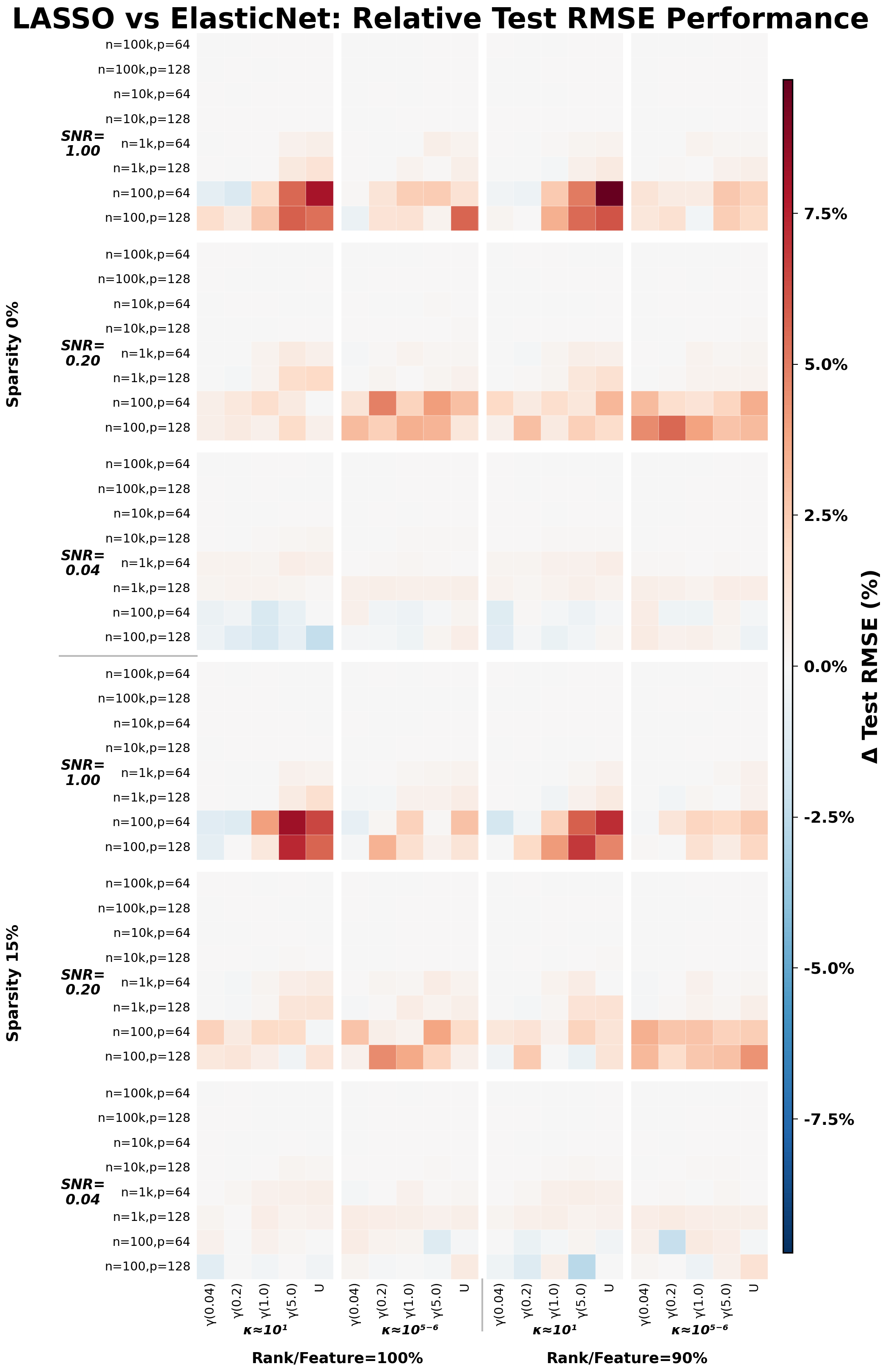}
\caption{ElasticNet generally outperforms Lasso in scenarios with low sample sizes ($n=100$) and high SNR, particularly as the correlation between features ($\gamma$) increases.}
\end{figure}

\begin{figure}[!htbp]
\centering
\includegraphics[width=0.86\textwidth]{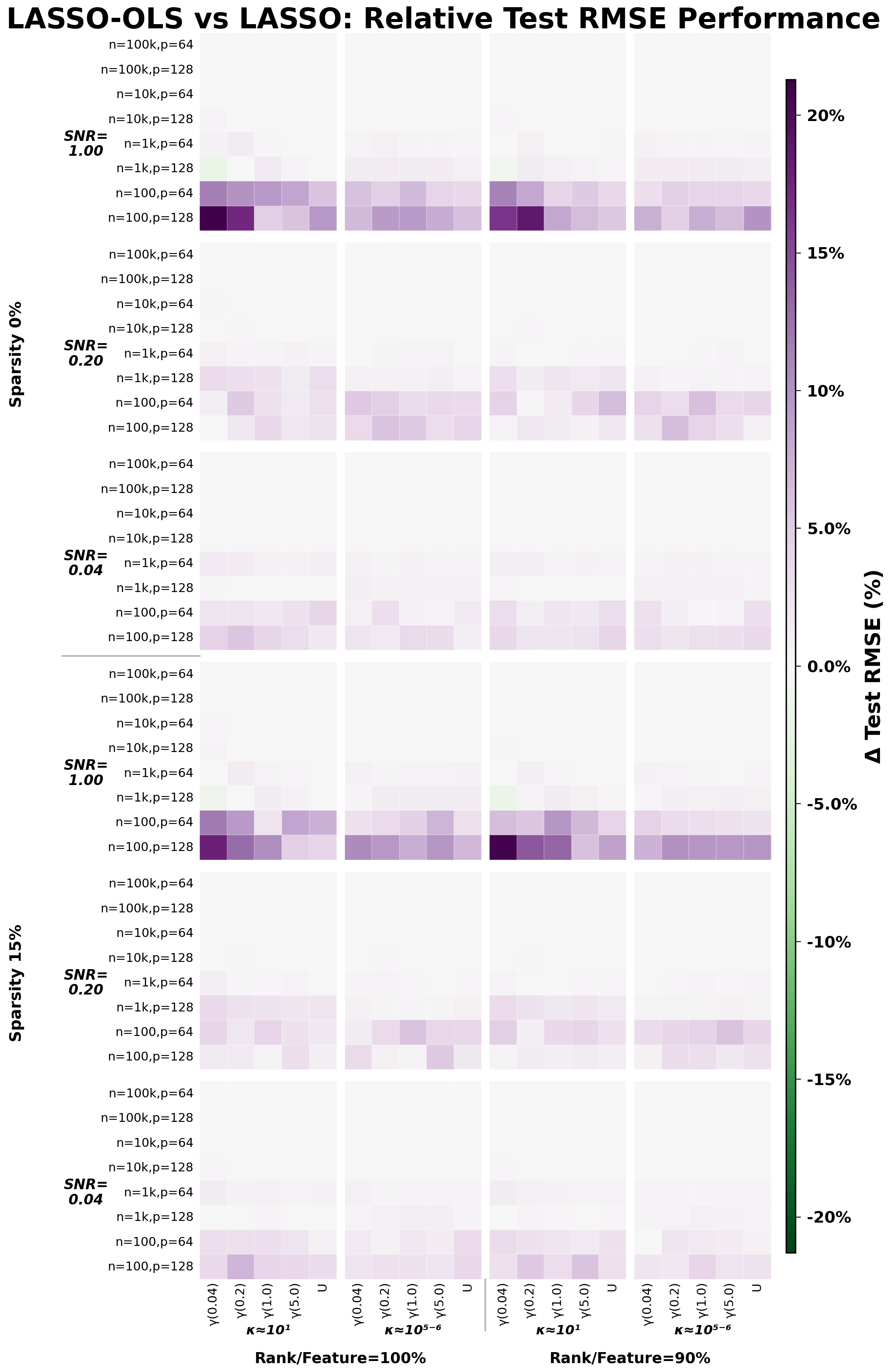}
\caption{Lasso generally outperforms Post-Lasso OLS (positive $\Delta$ Test RMSE) particularly in low-$n$ ($n=100$), high-noise ($SNR=1.00$) scenarios.}
\end{figure}

\begin{figure}[!htbp]
\centering
\includegraphics[width=0.86\textwidth]{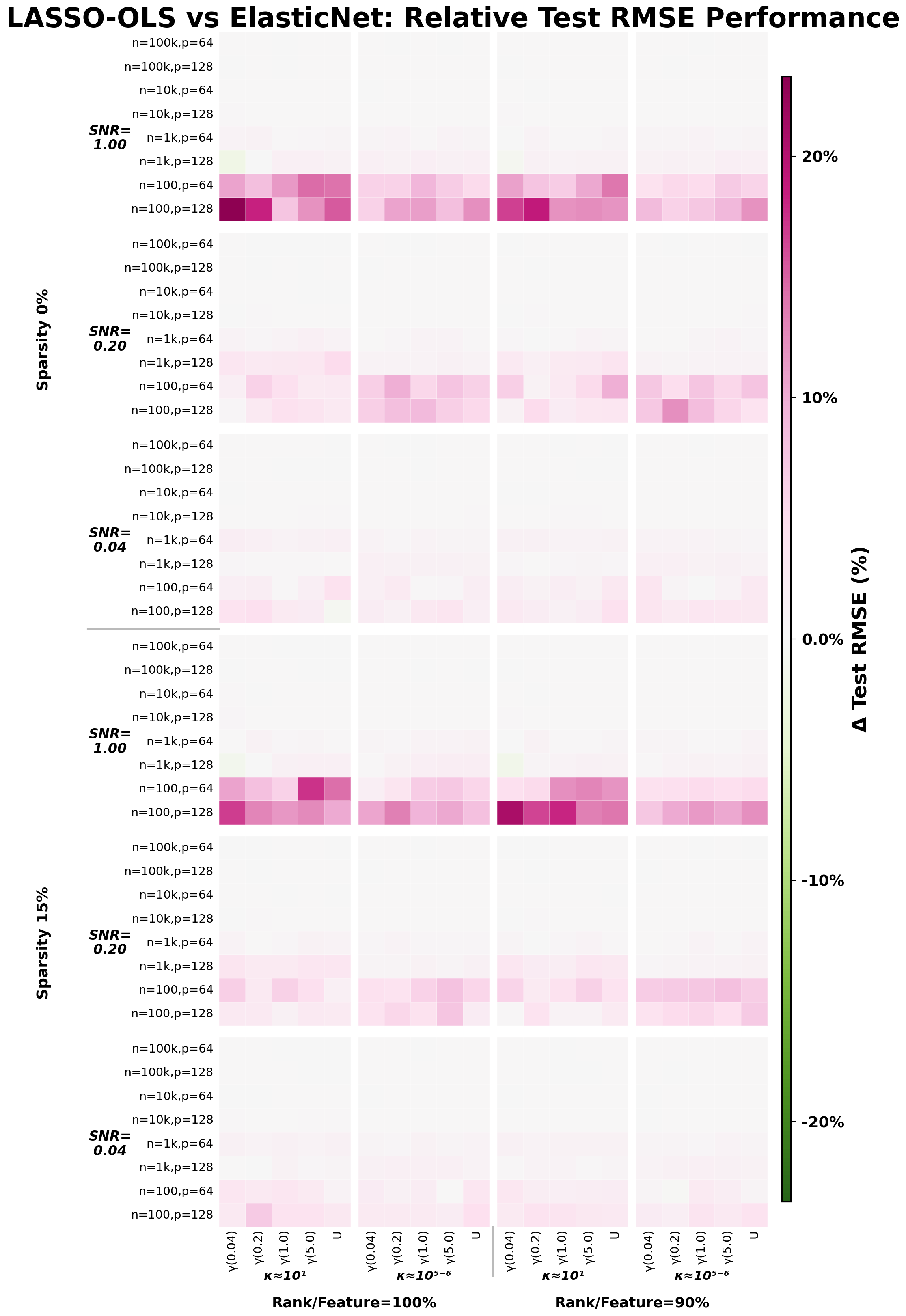}
\caption{ElasticNet generally outperforms Post-Lasso OLS across most sparsity and SNR levels, with the most significant relative reductions in test RMSE occurring in low-$n$ ($n=100$) and high-SNR ($SNR=1.00$) scenarios.}
\end{figure}

\begin{figure}[!htbp]
\centering
\includegraphics[width=0.86\textwidth]{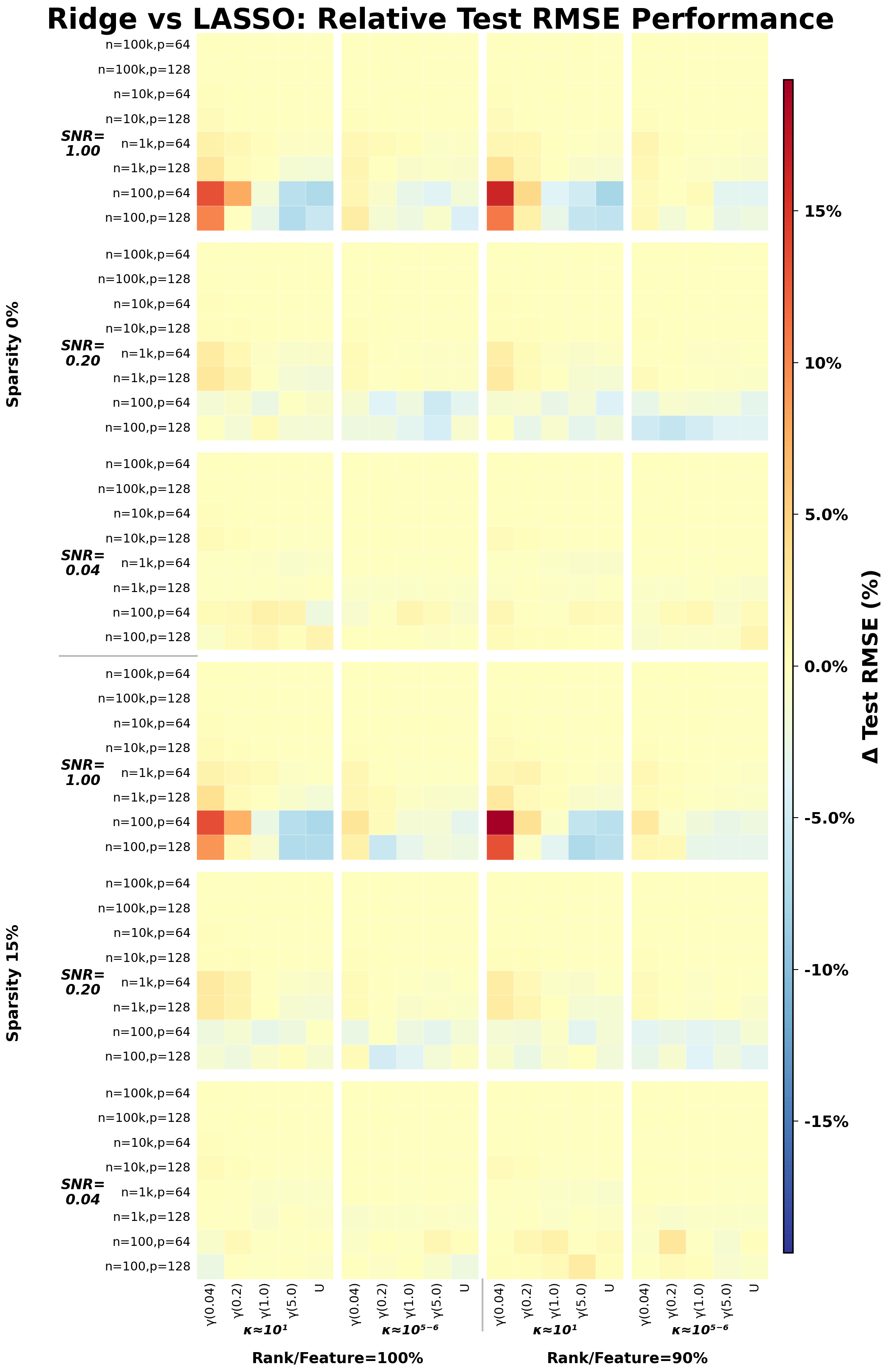}
\caption{Ridge generally outperforms Lasso (negative $\Delta$ Test RMSE) in low SNR and high-rank scenarios, while Lasso tends to excel in sparse, high-SNR conditions with small sample sizes ($n=100$).}
\end{figure}

\begin{figure}[!htbp]
\centering
\includegraphics[width=0.86\textwidth]{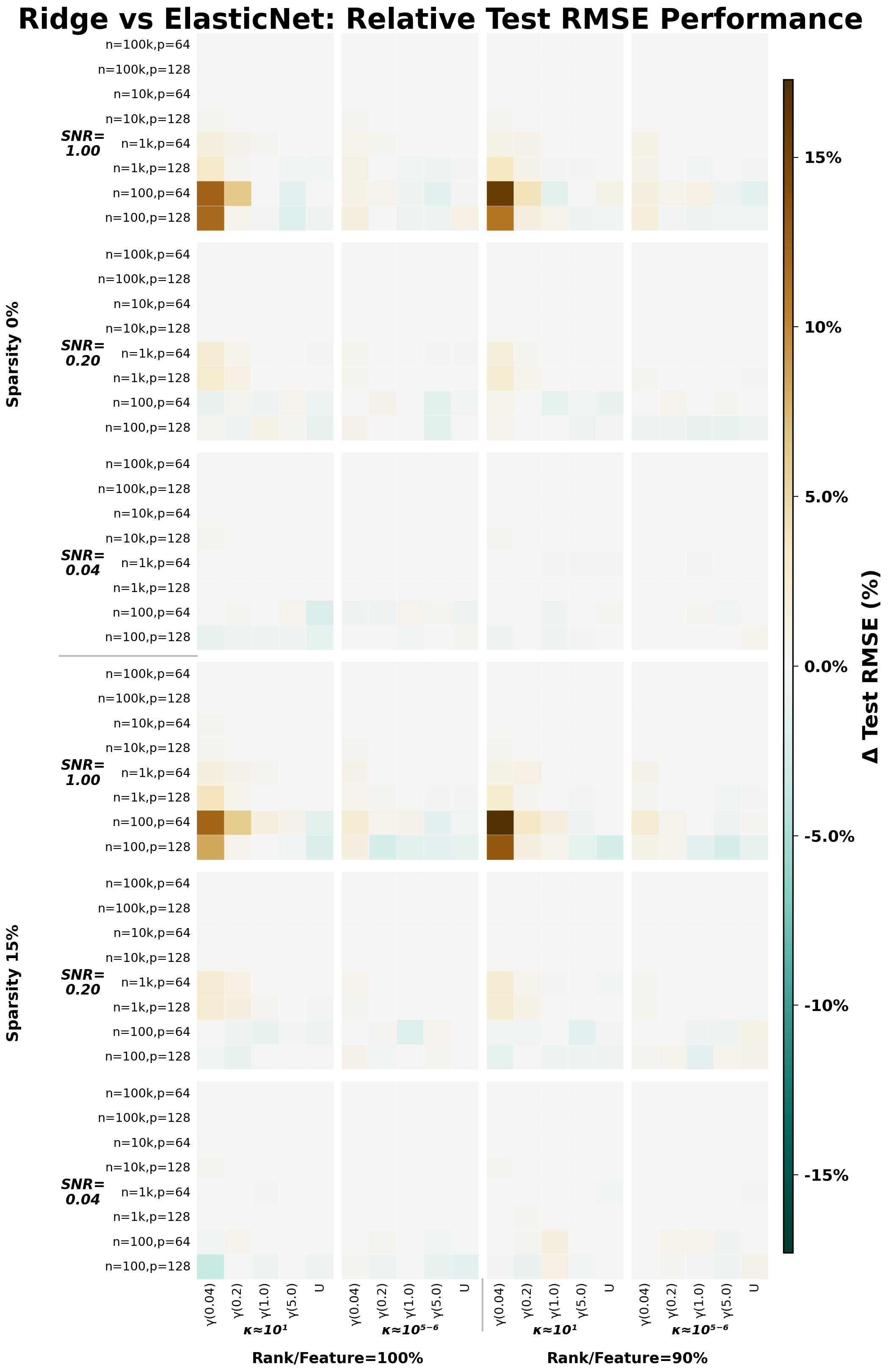}
\caption{While performance is similar across most conditions, ElasticNet generally outperforms Ridge (lower RMSE) in high-SNR ($SNR=1.00$) and low-$n$ ($n=100$) scenarios.}
\end{figure}

\begin{figure}[!htbp]
\centering
\includegraphics[width=0.86\textwidth]{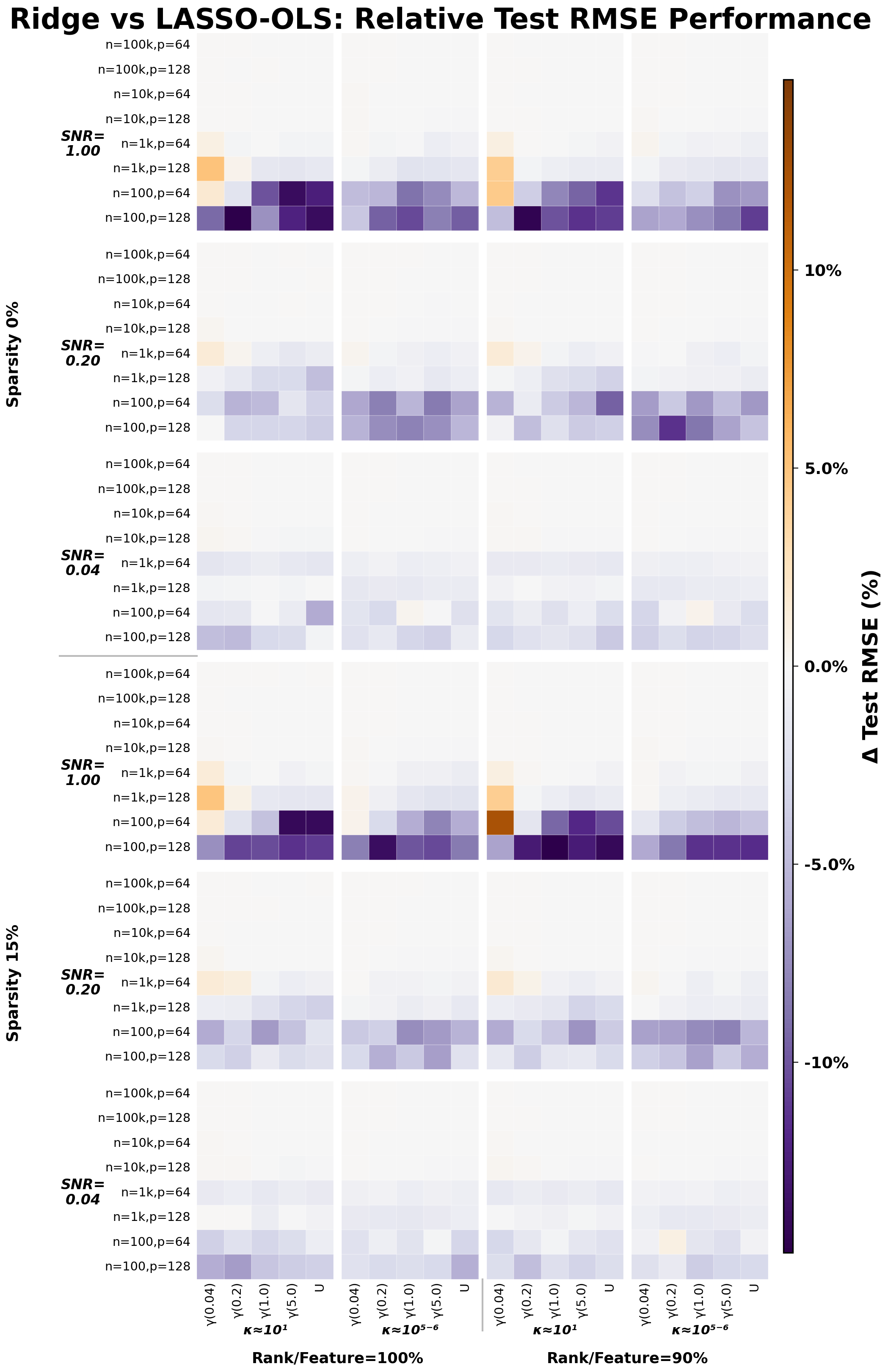}
\caption{Ridge consistently achieves a lower RMSE than Post-Lasso OLS in scenarios with low sample sizes ($n=100$) and high SNR, outperforming its counterpart as the feature-to-sample ratio increases.}
\end{figure}

\clearpage

\section{Compute Times}

\begin{figure}[!htbp]
\centering
\includegraphics[width=1.0\textwidth]{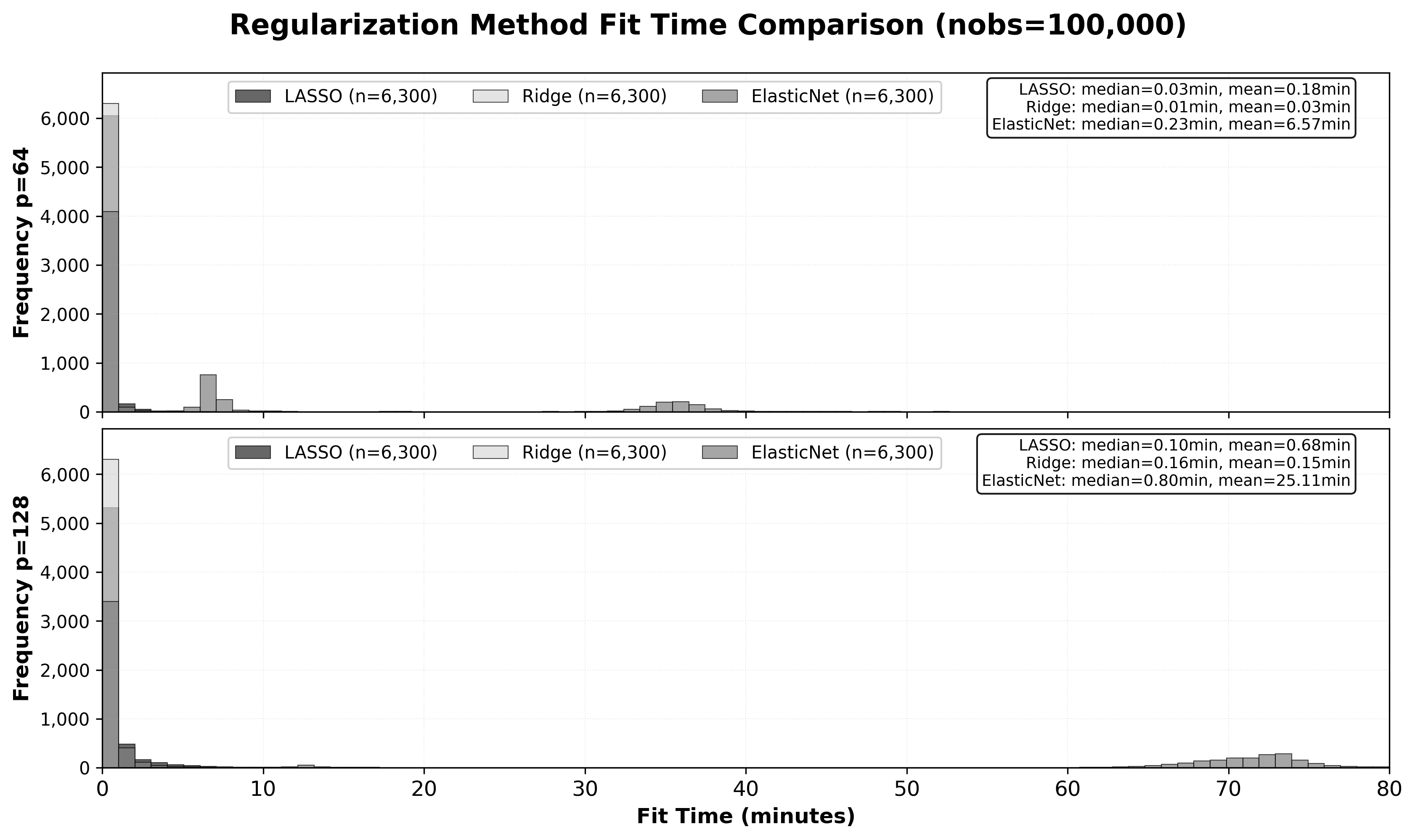}
\caption{Ridge is consistently the most computationally efficient method across both feature dimensions, while ElasticNet exhibits significantly higher mean fit times.}
\end{figure}

\end{appendices}

%% file: references.bib
@inproceedings{aggarwal2001distance,
  author    = {Aggarwal, Charu C. and Hinneburg, Alexander and Keim, Daniel A.},
  title     = {On the surprising behavior of distance metrics in high dimensional space},
  booktitle = {Database Theory — ICDT 2001},
  series    = {Lecture Notes in Computer Science},
  pages     = {420--434},
  year      = {2001},
  doi       = {10.1007/3-540-44503-X_27}
}

@article{berk2013postselection,
  author    = {Berk, Richard and Brown, Lawrence and Buja, Andreas and Zhang, Kai and Zhao, Linda},
  title     = {Valid post-selection inference},
  journal   = {The Annals of Statistics},
  volume    = {41},
  number    = {2},
  pages     = {802--837},
  year      = {2013},
  doi       = {10.1214/12-AOS1077}
}

@article{bertsimas2020sparse,
  author    = {Bertsimas, Dimitris and Pauphilet, Jean and Van Parys, Bart},
  title     = {Sparse Regression: Scalable Algorithms and Empirical Performance},
  journal   = {Statistical Science},
  volume    = {35},
  number    = {4},
  pages     = {555--578},
  year      = {2020},
  url       = {https://www.jstor.org/stable/26997931}
}

@article{bickel2009simultaneous,
  author    = {Bickel, Peter J. and Ritov, Ya'acov and Tsybakov, Alexandre B.},
  title     = {Simultaneous analysis of {Lasso} and {Dantzig} selector},
  journal   = {The Annals of Statistics},
  volume    = {37},
  number    = {4},
  pages     = {1705--1732},
  year      = {2009},
  doi       = {10.1214/08-AOS620}
}

@article{bertsimas2020highdim,
  author    = {Bertsimas, Dimitris and Van Parys, Bart},
  title     = {Sparse high-dimensional regression: Exact scalable algorithms and phase transitions},
  journal   = {Annals of Statistics},
  volume    = {48},
  number    = {1},
  pages     = {300--323},
  year      = {2020},
  doi       = {10.1214/18-AOS1804}
}

@article{bertsimas2020rejoinder,
  author    = {Bertsimas, Dimitris and Pauphilet, Jean and Van Parys, Bart},
  title     = {Rejoinder: Sparse regression: Scalable algorithms and empirical performance},
  journal   = {Statistical Science},
  volume    = {35},
  number    = {4},
  pages     = {623--624},
  year      = {2020},
  doi       = {10.1214/20-STS701REJ}
}

@article{bertsimas2021classification,
  author    = {Bertsimas, Dimitris and Pauphilet, Jean and Van Parys, Bart},
  title     = {Sparse classification: a scalable discrete optimization perspective},
  journal   = {Machine Learning},
  volume    = {110},
  number    = {11–12},
  pages     = {3177--3209},
  year      = {2021},
  doi       = {10.1007/s10994-021-06085-5}
}

@article{bertsimas2016subset,
  author    = {Bertsimas, Dimitris and King, Alex and Mazumder, Rahul},
  title     = {Best subset selection via a modern optimization lens},
  journal   = {Annals of Statistics},
  volume    = {44},
  number    = {2},
  pages     = {813--852},
  year      = {2016},
  doi       = {10.1214/15-AOS1388}
}

@article{breiman1992bootstrap,
  author    = {Breiman, Leo},
  title     = {The Little Bootstrap and Other Methods for Dimensionality Selection in Regression: {X}-Fixed Prediction Error},
  journal   = {Journal of the American Statistical Association},
  volume    = {87},
  number    = {419},
  pages     = {738--754},
  year      = {1992},
  url       = {https://www.jstor.org/stable/2290212}
}

@article{breiman1995garrote,
  author    = {Breiman, Leo},
  title     = {Better Subset Regression Using the Nonnegative Garrote},
  journal   = {Technometrics},
  volume    = {37},
  number    = {4},
  pages     = {373--384},
  year      = {1995},
  doi       = {10.1080/00401706.1995.10484371}
}

@article{breiman1996heuristics,
  author    = {Breiman, Leo},
  title     = {Heuristics of Instability and Stabilization in Model Selection},
  journal   = {The Annals of Statistics},
  volume    = {24},
  number    = {6},
  pages     = {2350--2383},
  year      = {1996},
  url       = {http://www.jstor.org/stable/2242688}
}

@book{buhlmann2011highdim,
  author    = {B{\"u}hlmann, Peter and van de Geer, Sara},
  title     = {Statistics for High-Dimensional Data: Methods, Theory and Applications},
  year      = {2011},
  publisher = {Springer},
  series    = {Springer Series in Statistics},
  doi       = {10.1007/978-3-642-20192-9},
  isbn      = {978-3-642-20191-2}
}

@article{candes2007dantzig,
  author    = {Candès, Emmanuel and Tao, Terence},
  title     = {The {Dantzig} selector: Statistical estimation when $p$ is much larger than $n$},
  journal   = {The Annals of Statistics},
  volume    = {35},
  number    = {6},
  pages     = {2313--2351},
  year      = {2007},
  doi       = {10.1214/009053606000001523}
}

@article{candes2018modelx,
  author    = {Candès, Emmanuel and Fan, Yingying and Janson, Lucas and Lv, Jinchi},
  title     = {Panning for gold: 'model-{X}' knockoffs for high dimensional controlled variable selection},
  journal   = {Journal of the Royal Statistical Society. Series B (Statistical Methodology)},
  volume    = {80},
  number    = {3},
  pages     = {551--577},
  year      = {2018},
  url       = {https://faculty.marshall.usc.edu/yingying-fan/publications/JRSSB-CFJL18.pdf}
}

@article{carvalho2010horseshoe,
  author    = {Carvalho, Carlos M. and Polson, Nicholas G. and Scott, James G.},
  title     = {The horseshoe estimator for sparse signals},
  journal   = {Biometrika},
  volume    = {97},
  number    = {2},
  pages     = {465--480},
  year      = {2010},
  doi       = {10.1093/biomet/asq017}
}

@article{chen2020robustness,
  author  = {Chen, Yuxin and Taeb, Ahmad and Bühlmann, Peter},
  title   = {A Look at Robustness and Stability of {$\ell_1$}-versus {$\ell_0$}-Regularization: Discussion of Papers by Bertsimas et al. and Hastie et al.},
  journal = {Statistical Science},
  volume  = {35},
  number  = {4},
  pages   = {614--622},
  year    = {2020},
  doi     = {10.1214/20-STS809}
}

@article{copas1983shrinkage,
  author    = {Copas, J. B.},
  title     = {Regression, Prediction and Shrinkage},
  journal   = {Journal of the Royal Statistical Society. Series B (Methodological)},
  volume    = {45},
  number    = {3},
  pages     = {311--354},
  year      = {1983},
  url       = {http://www.jstor.org/stable/2345402}
}

@book{draper1998regression,
  author    = {Draper, Norman R. and Smith, Harry},
  title     = {Applied Regression Analysis},
  edition   = {3},
  year      = {1998},
  publisher = {Wiley},
  doi       = {10.1002/9781118625590}
}

@article{efron2004lar,
  author    = {Efron, Bradley and Hastie, Trevor and Johnstone, Iain and Tibshirani, Robert},
  title     = {Least angle regression},
  journal   = {Annals of Statistics},
  volume    = {32},
  number    = {2},
  pages     = {407--499},
  year      = {2004},
  doi       = {10.1214/009053604000000067}
}

@incollection{efroymson1960regression,
  author    = {Efroymson, M. A.},
  title     = {Multiple regression analysis},
  booktitle = {Mathematical Methods for Digital Computers},
  pages     = {191--203},
  year      = {1960},
  publisher = {Wiley}
}

@article{fu1998bridge,
  author    = {Fu, Wenjiang J.},
  title     = {Penalized Regressions: The Bridge versus the Lasso},
  journal   = {Journal of Computational and Graphical Statistics},
  volume    = {7},
  number    = {3},
  pages     = {397--416},
  year      = {1998},
  doi       = {10.2307/1390712}
}

@article{fei2019spares,
  author    = {Fei, Zhe and Zhu, Ji and Banerjee, Moulinath and Li, Yi},
  title     = {Drawing inferences for high-dimensional linear models: A selection-assisted partial regression and smoothing approach},
  journal   = {Biometrics},
  volume    = {75},
  number    = {2},
  pages     = {551--561},
  year      = {2019},
  doi       = {10.1111/biom.13013}
}

@article{fan2010sis,
  author    = {Fan, Jianqing and Song, Rui},
  title     = {Sure independence screening in generalized linear models with NP-dimensionality},
  journal   = {Annals of Statistics},
  volume    = {38},
  number    = {6},
  pages     = {3567--3604},
  year      = {2010},
  doi       = {10.1214/10-AOS798}
}

@article{fan2001nonconcave,
  author    = {Fan, Jianqing and Li, Runze},
  title     = {Variable Selection via Nonconcave Penalized Likelihood and its Oracle Properties},
  journal   = {Journal of the American Statistical Association},
  volume    = {96},
  number    = {456},
  pages     = {1348--1360},
  year      = {2001},
  doi       = {10.1198/016214501753382273}
}

@inproceedings{gamarnik2017binary,
  author    = {Gamarnik, David and Zadik, Ilias},
  title     = {High Dimensional Regression with Binary Coefficients: Estimating Squared Error and a Phase Transition},
  booktitle = {Proceedings of the 2017 Conference on Learning Theory},
  series    = {Proceedings of Machine Learning Research},
  volume    = {65},
  pages     = {948--953},
  year      = {2017},
  url       = {https://proceedings.mlr.press/v65/david17a.html}
}

@book{hastie2009elements,
  author    = {Hastie, Trevor and Tibshirani, Robert and Friedman, Jerome},
  title     = {The Elements of Statistical Learning},
  year      = {2009},
  publisher = {Springer},
  doi       = {10.1007/b94608_1}
}

@article{hesterberg2008review,
  author    = {Hesterberg, Timothy and Choi, Nam Hee and Meier, Lukas and Fraley, Chris},
  title     = {Least angle and $\ell$1 penalized regression: A review},
  journal   = {Statistics Surveys},
  volume    = {2},
  pages     = {61--93},
  year      = {2008},
  doi       = {10.1214/08-SS035}
}

@article{hoerl1970ridgeapp,
  author    = {Hoerl, Arthur E. and Kennard, Robert W.},
  title     = {Ridge Regression: Applications to Nonorthogonal Problems},
  journal   = {Technometrics},
  volume    = {12},
  number    = {1},
  pages     = {69--82},
  year      = {1970},
  doi       = {10.2307/1267352}
}

@article{javanmard2014confidence,
  author  = {Javanmard, Adel and Montanari, Andrea},
  title   = {Confidence Intervals and Hypothesis Testing for High-Dimensional Regression},
  journal = {Journal of Machine Learning Research},
  volume  = {15},
  pages   = {2869--2909},
  year    = {2014},
  url     = {https://jmlr.org/papers/v15/javanmard14a.html}
}

@article{lederer2021lasso,
  author  = {Lederer, Johannes and Vogt, Michael},
  title   = {Estimating the {Lasso}'s Effective Noise},
  journal = {Journal of Machine Learning Research},
  volume  = {22},
  number  = {276},
  pages   = {1--32},
  year    = {2021},
  url     = {https://www.jmlr.org/papers/v22/20-539.html}
}

@article{leng2006lasso,
  author    = {Leng, Chenlei and Lin, Yi and Wahba, Grace},
  title     = {A Note on the Lasso and Related Procedures in Model Selection},
  journal   = {Statistica Sinica},
  volume    = {16},
  number    = {4},
  pages     = {1273--1284},
  year      = {2006},
  url       = {http://www.jstor.org/stable/24307787}
}

@article{lee2016postselection,
  author    = {Lee, Jason D. and Sun, Dennis L. and Sun, Yuekai and Taylor, Jonathan E.},
  title     = {Exact post-selection inference, with application to the lasso},
  journal   = {Annals of Statistics},
  volume    = {44},
  number    = {3},
  pages     = {907--927},
  year      = {2016},
  doi       = {10.1214/15-AOS1371}
}

@article{meinshausen2007discussion,
  author    = {Meinshausen, N. and Rocha, G. and Yu, B.},
  title     = {Discussion: A tale of three cousins: {Lasso}, {L2Boosting} and {Dantzig}},
  journal   = {The Annals of Statistics},
  volume    = {35},
  number    = {6},
  pages     = {2373--2384},
  year      = {2007},
  doi       = {10.1214/009053607000000523},
  note      = {Discussion of ``The Dantzig selector'' by Candes and Tao}
}

@article{meinshausen2010stability,
  author    = {Meinshausen, Nicolai and Bühlmann, Peter},
  title     = {Stability selection},
  journal   = {Journal of the Royal Statistical Society. Series B (Statistical Methodology)},
  volume    = {72},
  number    = {4},
  pages     = {417--473},
  year      = {2010},
  doi       = {10.1111/j.1467-9868.2010.00740.x},
  url       = {https://rss.onlinelibrary.wiley.com/doi/full/10.1111/j.1467-9868.2010.00740.x}
}

@book{miller2019subset,
  author    = {Miller, Alan J.},
  title     = {Subset Selection in Regression},
  year      = {2019},
  publisher = {Chapman and Hall/CRC}
}

@article{pedregosa2011scikit,
  author    = {Pedregosa, Fabian and Varoquaux, Gaël and Gramfort, Alexandre and Michel, Vincent and Thirion, Bertrand and Grisel, Olivier and Blondel, Mathieu and Prettenhofer, Peter and Weiss, Ron and Dubourg, Vincent and Vanderplas, Jake and Passos, Alexandre and Cournapeau, David and Brucher, Matthieu and Perrot, Matthieu and Duchesnay, Edouard},
  title     = {Scikit-learn: Machine learning in Python},
  journal   = {Journal of Machine Learning Research},
  volume    = {12},
  pages     = {2825--2830},
  year      = {2011}
}

@article{pilanci2015boolean,
  author    = {Pilanci, Mert and Wainwright, Martin J. and El Ghaoui, Laurent},
  title     = {Sparse learning via Boolean relaxations},
  journal   = {Mathematical Programming},
  volume    = {151},
  number    = {1},
  pages     = {63--87},
  year      = {2015},
  doi       = {10.1007/s10107-015-0894-1}
}

@article{rencher1980inflation,
  author    = {Rencher, Alvin C. and Pun, Fu C.},
  title     = {Inflation of R² in Best Subset Regression},
  journal   = {Technometrics},
  volume    = {22},
  number    = {1},
  pages     = {49--53},
  year      = {1980},
  doi       = {10.2307/1268382}
}

@article{simon2013sparsegroup,
  author    = {Simon, Noah and Friedman, Jerome and Hastie, Trevor and Tibshirani, Robert},
  title     = {A Sparse-Group Lasso},
  journal   = {Journal of Computational and Graphical Statistics},
  volume    = {22},
  number    = {2},
  pages     = {231--245},
  year      = {2013},
  doi       = {10.1080/10618600.2012.681250},
  url       = {https://www.tandfonline.com/doi/full/10.1080/10618600.2012.681250}
}

@article{su2017lassopath,
  author    = {Su, Weijie and Bogdan, Malgorzata and Candès, Emmanuel},
  title     = {False discoveries occur early on the Lasso path},
  journal   = {Annals of Statistics},
  volume    = {45},
  number    = {5},
  pages     = {2133--2150},
  year      = {2017},
  doi       = {10.1214/16-AOS1521}
}

@inproceedings{tang2025benign,
  author    = {Tang, Shange and Wu, Jiayun and Fan, Jianqing and Jin, Chi},
  title     = {Benign Overfitting in Out-of-Distribution Generalization of Linear Models},
  booktitle = {Proceedings of the International Conference on Learning Representations (ICLR)},
  year      = {2025}
}

@article{tibshirani1996lasso,
  author    = {Tibshirani, Robert},
  title     = {Regression Shrinkage and Selection via the Lasso},
  journal   = {Journal of the Royal Statistical Society. Series B (Methodological)},
  volume    = {58},
  number    = {1},
  pages     = {267--288},
  year      = {1996},
  url       = {http://www.jstor.org/stable/2346178}
}

@article{tibshirani2005fusedlasso,
  author    = {Tibshirani, Robert and Saunders, Michael and Rosset, Saharon and Zhu, Ji and Knight, Keith},
  title     = {Sparsity and smoothness via the fused lasso},
  journal   = {Journal of the Royal Statistical Society. Series B (Statistical Methodology)},
  volume    = {67},
  number    = {1},
  pages     = {91--108},
  year      = {2005},
  url       = {https://academic.oup.com/jrsssb/article-abstract/67/1/91/7110658},
  note      = {Accessed February 15, 2026}
}

@article{vandegeer2014hdtests,
  author  = {van de Geer, Sara and B{\"u}hlmann, Peter and Ritov, Ya'acov and Dezeure, Ruben},
  title   = {On Asymptotically Optimal Confidence Regions and Tests for High-Dimensional Models},
  journal = {The Annals of Statistics},
  volume  = {42},
  number  = {3},
  pages   = {1166--1202},
  year    = {2014},
  doi     = {10.1214/14-AOS1221}
}

@article{wainwright2009sparsity,
  author    = {Wainwright, Martin J.},
  title     = {Information-theoretic limits on sparsity recovery in the high-dimensional and noisy setting},
  journal   = {IEEE Transactions on Information Theory},
  volume    = {55},
  number    = {12},
  pages     = {5728--5741},
  year      = {2009},
  doi       = {10.1109/tit.2009.2032816}
}

@article{wang2010limits,
  author    = {Wang, Wei and Wainwright, Martin J. and Ramchandran, Kannan},
  title     = {Information-theoretic limits on sparse signal recovery: Dense versus sparse measurement matrices},
  journal   = {IEEE Transactions on Information Theory},
  volume    = {56},
  number    = {6},
  pages     = {2967--2979},
  year      = {2010},
  doi       = {10.1109/tit.2010.2046199}
}

@article{wilkinson1981forward,
  author    = {Wilkinson, Leland and Dallal, Gerard E.},
  title     = {Tests of Significance in Forward Selection Regression With an F-to-Enter Stopping Rule},
  journal   = {Technometrics},
  volume    = {23},
  number    = {4},
  pages     = {377--380},
  year      = {1981},
  doi       = {10.1080/00401706.1981.10487682}
}

@article{huang2012groupselection,
  author    = {Huang, Jian and Breheny, Patrick and Ma, Shuangge},
  title     = {A Selective Review of Group Selection in High-Dimensional Models},
  journal   = {Statistical Science},
  volume    = {27},
  number    = {4},
  year      = {2012},
  doi       = {10.1214/12-STS392},
  url       = {https://pmc.ncbi.nlm.nih.gov/articles/PMC3810358/},
  note      = {Accessed February 15, 2026}
}

@article{luo2026penalization,
  author    = {Luo, J. and Kong, Y. and Li, G.},
  title     = {From Penalization to Over-Parameterization: Achieving Implicit Regularization for High-Dimensional Linear Errors-in-Variables Models},
  journal   = {Journal of Business \& Economic Statistics},
  year      = {2026},
  pages     = {1--13},
  doi       = {10.1080/07350015.2025.2583457},
  note      = {Published online ahead of print}
}

@article{yuan2006grouped,
  author    = {Yuan, Ming and Lin, Yi},
  title     = {Model Selection and Estimation in Regression with Grouped Variables},
  journal   = {Journal of the Royal Statistical Society},
  volume    = {68},
  pages     = {49--67},
  year      = {2006},
  url       = {https://www.scirp.org/reference/referencespapers?referenceid=2514485},
  note      = {Accessed February 15, 2026}
}

@article{zhang2010mcp,
  author    = {Zhang, Cun-Hui},
  title     = {Nearly unbiased variable selection under minimax concave penalty},
  journal   = {Annals of Statistics},
  volume    = {38},
  number    = {2},
  pages     = {894--942},
  year      = {2010},
  doi       = {10.1214/09-aos729}
}

@article{zhang2014confidence,
  author  = {Zhang, Cun-Hui and Zhang, Stephanie S.},
  title   = {Confidence intervals for low dimensional parameters in high dimensional linear models},
  journal = {Journal of the Royal Statistical Society: Series B (Statistical Methodology)},
  year    = {2014},
  volume  = {76},
  number  = {1},
  pages   = {217--242},
}

@article{zou2006adaptive,
  author    = {Zou, Hui},
  title     = {The Adaptive Lasso and Its Oracle Properties},
  journal   = {Journal of the American Statistical Association},
  volume    = {101},
  number    = {476},
  pages     = {1418--1429},
  year      = {2006},
  doi       = {10.1198/016214506000000735}
}

@article{zou2005elasticnetaddendum,
  author    = {Zou, Hui and Hastie, Trevor},
  title     = {Addendum: Regularization and variable selection via the elastic net},
  journal   = {Journal of the Royal Statistical Society. Series B (Statistical Methodology)},
  volume    = {67},
  number    = {5},
  pages     = {768--768},
  year      = {2005},
  doi       = {10.1111/j.1467-9868.2005.00527.x}
}

@article{frank1993statistical,
  author  = {Frank, I. E. and Friedman, J. H.},
  title   = {A statistical view of some chemometrics regression tools (with discussion)},
  journal = {Technometrics},
  volume  = {35},
  pages   = {109--148},
  year    = {1993}
}

@online{lemaitre2023ridge,
  author    = {Lemaitre, Guillaume},
  title     = {Why is the Ridge regression loss not normalized by the number of samples?},
  year      = {2023},
  date      = {2023-05-22},
  url       = {https://github.com/scikit-learn/scikit-learn/discussions/23407},
  note      = {Comment by \texttt{glemaitre} on GitHub Discussion},
  organization = {scikit-learn},
}

@article{park2008bayesianlasso,
  author    = {Trevor Park and George Casella},
  title     = {The Bayesian Lasso},
  journal   = {Journal of the American Statistical Association},
  year      = {2008},
  volume    = {103},
  number    = {482},
  pages     = {681--686},
  url       = {https://people.eecs.berkeley.edu/~jordan/courses/260-spring09/other-readings/park-casella.pdf},
  doi       = {10.1198/016214508000000337}
}

@book{cohen1988statistical,                                       
  author    = {Cohen, Jacob},                                   
  title     = {Statistical Power Analysis for the Behavioral Sciences},           edition   = {2nd},                                                              publisher = {Lawrence Erlbaum Associates},               
  address   = {Hillsdale, NJ},               
  year      = {1988},                                                    
  isbn      = {0-8058-0283-5}
}

@article{freijeiro_gonzalez_critical_lasso_2022,
  title        = {A Critical Review of {{LASSO}} and Its Derivatives for Variable Selection Under Dependence Among Covariates},
  author       = {Freijeiro‐Gonz{\'a}lez, Laura and Febrero‐Bande, Manuel and Gonz{\'a}lez‐Manteiga, Wenceslao},
  journal      = {International Statistical Review},
  year         = {2022},
  volume       = {90},
  number       = {1},
  pages        = {118--145},
  doi          = {10.1111/insr.12469},
}

@article{belloni2013least,
  title = {Least squares after model selection in high-dimensional sparse models},
  author = {Belloni, Alexandre and Chernozhukov, Victor},
  journal = {Bernoulli},
  volume = {19},
  number = {2},
  pages = {521--547},
  year = {2013},
  doi = {10.3150/11-BEJ410},
  publisher = {Bernoulli Society for Mathematical Statistics and Probability}
}

@article{fitzgerald_sice_practical_insights_double_lasso_2023,
  title   = {Double {LASSO}: Replication and Practical Insights},
  author  = {Fitzgerald Sice, Jack and Lattimore, Finnian and Robinson, Tim and Zhu, Anna},
  journal = {Journal of Applied Econometrics},
  year    = {2026},
  doi     = {10.1002/jae.70041},
  note    = {Published online February 15, 2026},
}
